\documentclass[nohyperref]{article}

\usepackage{microtype}
\usepackage{graphicx}
\usepackage{subfigure}
\usepackage{booktabs} 

\usepackage{hyperref}

 \usepackage[accepted]{icml2022}

\usepackage{amsmath}
\usepackage{amssymb}
\usepackage{mathtools}
\usepackage{amsthm}

\usepackage[capitalize,noabbrev]{cleveref}

\theoremstyle{plain}

\theoremstyle{definition}

\theoremstyle{remark}

\usepackage[textsize=tiny]{todonotes}

\icmltitlerunning{Accelerated Federated Learning with Decoupled Adaptive Optimization}

\usepackage{amssymb,amsmath,amsfonts,epsfig,graphics,enumerate,float,subfigure}
\usepackage{cases}
\usepackage{color}
\usepackage{upgreek}
\usepackage{xspace}
\usepackage{float}
\usepackage{latexsym}
\usepackage{url}
\usepackage{algorithm}
\usepackage{algorithmic}
\usepackage{graphicx}
\usepackage{subfigure}
\usepackage{bigdelim}
\usepackage{multicol}
\usepackage{multirow}
\usepackage{rotating}
\usepackage{caption}
\usepackage{extarrows}

\newcommand{\method}{FedDA}

\begin{document}

\twocolumn[
\icmltitle{Accelerated Federated Learning with Decoupled Adaptive Optimization}



\icmlsetsymbol{equal}{*}

\begin{icmlauthorlist}
\icmlauthor{Jiayin Jin$^{*}$}{AU}
\icmlauthor{Jiaxiang Ren$^{*}$}{AU}
\icmlauthor{Yang Zhou}{AU}
\icmlauthor{Lingjuan Lyu}{Sony}
\icmlauthor{Ji Liu}{BD}
\icmlauthor{Dejing Dou}{BD,UO}
\end{icmlauthorlist}

\icmlaffiliation{AU}{Auburn University, USA}
\icmlaffiliation{Sony}{Sony AI, Japan}
\icmlaffiliation{BD}{Baidu Research, China}
\icmlaffiliation{UO}{University of Oregon, USA}

\icmlcorrespondingauthor{Yang Zhou}{yangzhou@auburn.edu}
\icmlcorrespondingauthor{Lingjuan Lyu}{lingjuan.lv@sony.com}

\icmlkeywords{Machine Learning, ICML}

\vskip 0.3in
]



\printAffiliationsAndNotice{\icmlEqualContribution} 

\begin{abstract}
The federated learning (FL) framework enables edge clients to collaboratively learn a shared inference model while keeping privacy of training data on clients. Recently, many heuristics efforts have been made to generalize centralized adaptive optimization methods, such as SGDM, Adam, AdaGrad, etc., to federated settings for improving convergence and accuracy. However, there is still a paucity of theoretical principles on where to and how to design and utilize adaptive optimization methods in federated settings. This work aims to develop novel adaptive optimization methods for FL from the perspective of dynamics of ordinary differential equations (ODEs). First, an analytic framework is established to build a connection between federated optimization methods and decompositions of ODEs of corresponding centralized optimizers. Second, based on this analytic framework, a momentum decoupling adaptive optimization method, {\sc \method}, is developed to fully utilize the global momentum on each local iteration and accelerate the training convergence. Last but not least, full batch gradients are utilized to mimic centralized optimization in the end of the training process to ensure the convergence and overcome the possible inconsistency caused by adaptive optimization methods.
\end{abstract}

\vspace{-0.15cm}
\section{Introduction}\label{sec.introduction}
\vspace{-0.1cm}
Recent advances in federated learning (FL) present a promising machine learning (ML) paradigm that enables collaborative training of shared ML models over multiple distributed devices without the need for data sharing, while mitigating the data isolation as well as protecting the data privacy~\cite{KMYR16,KMRR16,MMRA16,MMRH17,KMAB21}. In a FL model, end clients (e.g., Android phones) use their local data to train a local ML model, while keeping the local data decentralized. The end clients send the parameter updates of local models (e.g., Android phone updates) rather than raw data to the central server (e.g., Android cloud). The server produces a shared global ML model by aggregating the local updates.

One of critical challenges in the FL paradigm is expensive communication cost between the clients and server~\cite{KMYR16,CKMT18,LKZL19,HMS20,HAA20}. Traditional federated optimization methods, such as FedAvg and its variants~\cite{KMYR16,KMRR16,MMRH17,KMAB21}, use local client updates, where the clients perform multiple epochs of stochastic gradient descent (SGD) on their local datasets to update their models before communicating to the server. This can dramatically  reduce the communication required to train a FL model.

Despite achieving remarkable performance, FL techniques often suffer from two key challenging issues: (1) Client drift. Too many SGD epochs on the same clients may result in overfitting to their local datasets, such that the local models are far away from globally optimal models and the FL training achieves slower convergence~\cite{KKMR20,WPS20,RCZG21,KJKM21}; and (2) lack of adaptivity. Standard SGD optimization methods used in most FL models may be unsuitable for federated settings and result in high communication costs~\cite{ZKVK19,RCZG21}.

In centralized ML models, adaptive optimization methods, such as SGD with Momentum (SGDM)~\cite{RHW86,Qian99,SMDH13}, Adaptive Moment Estimation (Adam)~\cite{KiBa14}, Adaptive Gradient (AdaGrad)~\cite{McSt10,DHS11}, etc., have achieved superior success in speeding up the training convergence. Adaptive optimization methods are designed to control possible deviations of mini-batch gradients in centralized models. Several recent studies try to adapt centralized optimization algorithms to the federated learning settings for achieving faster convergence and higher test accuracy~\cite{XKGL19,RCZG21,KJKM21,WXGC21}.

The above adaptive optimization methods can be broadly classified into two categories: (1) Server adaptive methods. FedOpt is a generalization of FedAvg that allows the clients and the server to choose different optimization methods~\cite{RCZG21}. Although FedOPT is a generic framework that can use adaptive optimizers as client or server optimizers, the core section (Section 3) of the FedOPT paper only considers the setting that ClientOPT is SGD, which fails to make full use of the strength of the adaptive methods for improving the convergence~\cite{WXGC21}; (2) Client adaptive techniques. Local AdaAlter proposes a novel SGD variant based on AdaGrad, and adopt the concept of local SGD to reduce the communication~\cite{XKGL19}. Mime uses a combination of control-variates and server-level optimizer state (e.g. momentum) at every client-update step to ensure that each local update mimics that of the centralized method run on i.i.d. data~\cite{KJKM21}. However, Local AdaAlter and Mime use the same optimizer states (momentums and pre-conditioners) on all clients, which is similar to server adaptive methods in FedOpt and does not exploit local adaptivity~\cite{WXGC21}.
FedLocal is the first real client adaptive approach that employs adaptive optimization methods for local updates at clients~\cite{WXGC21}. It restarts the update of client optimizer states (i.e., momentums and pre-conditioners) at the beginning of each round. This restart operation fails to inherit and aggregate the states from previous rounds and may loss the strength of adaptive optimization methods in centralized models, which aggregate the states from previous rounds to speed up the convergence. For the standard FL models, applying adaptive optimization methods to local updates may come with the cost of a non-vanishing solution bias: instead of using current gradient to update local parameters in the standard SGD, the adaptive methods utilize the aggregation of client optimizer states (i.e., momentum aggregates previous and past gradients) in multiple client-update steps to update local parameters. In this case, the heterogeneity property in federated settings results in large deviations among uploaded local gradients as well as  makes the converge point far away from the global minimizer and slow down the convergence, compared with the standard SGD. FedLocal proposes correction techniques to overcome this bias issue and to complement the local adaptive methods for FL. However, the correction techniques that are essentially heuristic post-processing methods without knowledge of previous client optimizer states on the server are hard to solve this problem fundamentally for stochastic optimizers, such as SGDM, Adam, AdaGrad, etc. 
Therefore, the above techniques often follow manually-crafted heuristics to generalize centralized adaptive optimization methods to the federated settings. There is still a paucity of theoretical principles on where to and how to design and utilize adaptive optimization methods in federated settings.

This work aims to develop novel adaptive optimization methods for FL from the perspective of the decomposition of ODEs of centralized optimizers. We establish an analytic framework to connect the federated optimization methods with the decompositions of ODEs of corresponding centralized optimizers. The centralized gradient descent (CGD) is utilized as a warm-up example to briefly illustrate our underlying idea. The CGD reads $W(t+1) = W(t) - \sum_{i=1}^M \frac{N^i}{N}L^i(W(t))*\eta$, where $W(t)$ is the global parameter, $M$ is the number of clients, $N^i$ is the number of training samples on $i^{\text{th}}$ device, $N = \sum_i^M N^i$ is the total number of training samples, $L^i$ is the loss function on $i^{\text{th}}$ client. It is straightforward to check that $\sum_{i=1}^M \frac{N^i}{N}L^i(W(t))$ is the total loss of centralized training. Therefore, the CGD is the numerical solution of an ODE system $\frac{d}{d\tau}W(\tau) = - \sum_{i=1}^M \frac{N^i}{N}L^i(W(\tau))$. One way to decompose the ODE system is as follows: $\bar W(\tau) = \sum_{i=1}^M \frac{N^i}{N}W^i(\tau)$,  where $W^i(\tau)$ solves $\frac{d}{d\tau}W^i(\tau) = - L^i(W^i(\tau))$, $i = 1,\cdots M$. Thus, $\bar W(\tau)$ is an approximate solution to the above ODE system. The numerical solution of the decomposed system is $W^i(t+1) = W^i(t) - \nabla L^i(W^i(t))*\eta$, and then $\bar W(t+1) = \sum_{i=1}^M \frac{N^i}{N}W^i(t)$ is an approximate numerical solution to the ODE system of CGD. This scheme can be extended to the SGD case by replacing full batch gradients with mini-batch gradients, which is exactly the update rule of FedAvg~\cite{MMRH17}. This example illustrates how to derive federated optimization methods based on the decomposition of ODEs of centralized optimizers. Moreover, this example also demonstrates the rationality of FedAvg, since FedAvg is an approximation of the above ODE system of CGD and the convergence of the CGD guarantees the convergence of FedAvg if the approximation error is well controlled. In the same spirit, to design adaptive optimization methods for FL, it is crucial to discover suitable decompositions of ODEs of centralized adaptive optimizers. 

Based on the above analytic framework, we develop a momentum decoupling adaptive optimization method for FL. By decoupling the global momentum from local updates, the equation of global momentum becomes a linear equation. Consequently, we can decompose the equation of global momentum exactly and distribute the update of global momentum to local devices. The aggregation of the portions of the global momentum on local devices is exactly equal to the global momentum without any deviation. Particularly, in our \method\ method: (1) the global momentums are updated with local gradients in each local iteration; (2) all global momentums updated through local iterations will attend global training to update global model. This is an analogy to centralized training, which fully utilizes the global momentum. Notice that momentums can provide fast convergence and high accuracy for centralized training. The global momentum in our \method\ method makes the best effort to mimic the role of momentum in centralized training, which can accelerate the convergence of FL training. We theoretically demonstrate that (1) local momentum deviates from the centralized one at exponential rate; (2) global momentum in FedDA deviates from the centralized one at algebraic rate.

Adaptive optimization methods for FL are often faced with convergence inconsistency in training~\cite{WXGC21}. We propose to utilize full batch gradients of clients to mimic centralized optimization in the end of the training process to ensure the convergence. Based on our proposed framework of decomposing ODEs, if local devices only iterate once with full batch gradients in a training round, then our \method\ method is in agreement with centralized training in that round. Take the advantage of this, by reducing the iteration number to $1$ in the end of training, our \method\ method is able to mimic centralized training, which ensures the convergence and overcome the inconsistency.

Empirical evaluation on real federated tasks and datasets demonstrates the superior performance of our momentum decoupling adaptive optimization model against several state-of-the-art regular federated learning and federated optimization approaches. In addition, more experiments, implementation details, and hyperparameter selection and setting are presented in Appendices~\ref{sec.AdditionalExperiments}-\ref{sec.ExperimentDetails}. 

\vspace{-0.1cm}
\section{Preliminaries and Notations}\label{sec.problem}
\vspace{-0.1cm}
Federated learning aims to solve the following optimization problem.

\vspace{-0.35cm}
\begin{equation} \label{eq:Loss}
\begin{split}
&\min _{W \in \mathbb{R}^{d}} \mathcal{L}(W)=\sum_{i=1}^M \frac{N_i}{N} L^{i}(W) \\
&\text{where} \ L^{i}(W)=\frac{1}{N_{i}} \sum_{k \in \mathcal{P}_{i}} l_{k}(W)
\end{split}
\end{equation}
\vspace{-0.5cm}

where $l_{k}(W) = l(x_{k}, y_{k}; W)$ denotes the loss of the prediction on example $(x_{k}, y_{k})$ made with model parameters $W$.  
There are $M$ clients over which the data is partitioned, with $\mathcal{P}_{i}$ the set of indexes of data points on client $S^i$, with $N_{i}=|\mathcal{P}_{i}|$. $W^i$, $L^i$, $G^i$, $g^i$, and $N^i$ denote the local model parameter, loss function, full batch gradient, mini-batch gradient, number of training data on client $S^i$ respectively. $N$ is the total number of training data, i.e., $N = N^1+\cdots+N^M$ and $\eta$ represents the learning rate. In each round, there are $K$ clients participating in the training ($K \le M$).

This work aims to derive the theoretical principle on where to and how to design and utilize adaptive optimization methods in federated settings. Based on the theoretical principle, it tries to propose an effective and efficient framework to generalize adaptive optimization methods in centralized settings to FL with fast convergence and high accuracy.

\vspace{-0.1cm}
\vspace{-0.1cm}
\section{Decomposition of ODEs and FL}
In this section, we establish the connection between the decomposition of ODEs of centralized optimizers and FL. As a warm up, we start with the most simple centralized optimizer Gradient Descent. In particular, we demonstrate that relation between the decomposition of the ODE for GD and FedAvg, and explain why FedAvg works from ODE theory. 

Centralized training and FL share the same goal, that is to minimize the total loss $\mathcal{L}$.
 
\begin{equation}
\mathcal{L}(W) = \sum_{i=1}^M \frac{N^i}{N}L^i(W).	
\end{equation}

The most classical centralized optimization method is Gradient Descent (GD):
\begin{equation}\label{GD}
W(t+1) = W(t) - \nabla \mathcal{L}(W(t))*\eta.
\end{equation}
A more popular optimization is Stochastic Gradient Descent (SGD). It updates the model with the gradient of the loss of a mini-batch at each step, which accelerates the training process comparing to GD. Denote the loss of mini-batch at step $t$ by $\mathcal{L}^t$, then the training process of SGD can be expressed by
\begin{equation}\label{SGD}
W(t+1) = W(t) - \nabla \mathcal{L}^t(W(t))*\eta.
\end{equation}
From the point of view of ODEs, the training process of GD in Eq.\eqref{GD} is the numerical solution of the autonomous ODE
\begin{equation}\label{ODE-GD}
\frac{d}{d\tau}W(\tau) = -\nabla \mathcal{L}\big(W(\tau)\big) \le 0.
\end{equation}
The training process of SGD in Eq.\eqref{SGD} is as the numerical solution of the non-autonomous ODE
\begin{equation}\label{ODE-SGD}
\frac{d}{d\tau}W(\tau) = -\nabla \mathcal{L}^\tau\big(W(\tau)\big).
\end{equation}

In fact, Eq. \eqref{ODE-GD} is a gradient system, i.e., the vector field of the equation $-\nabla \mathcal{L}\big(W(\tau)\big)$ is in the gradient form. As a gradient system, it is typical that the loss $\mathcal{L}$ descends along the solutions until it reaches local minimum. This is because 
\begin{equation}\frac{d}{d\tau} \mathcal{L}\big(W(\tau)\big) = -|\nabla \mathcal{L}\big(W(\tau)\big)|^2.
\end{equation}

The decentralization of the training process onto local devices in FL is  an analogy to the decomposition of Eq.\eqref{ODE-GD} into a system of ODEs. Theoretically, a precise decomposition of Eq.\eqref{ODE-GD} is $W(\tau) = \sum_{i=1}^M \frac{N^i}{N}W^i(\tau)$, where $W^i(\tau)$ satisfies
\begin{equation}
\frac{d}{d\tau}W^i(\tau) = -\nabla L^i\big(W(\tau)\big), \ i = 1,\cdots, M.
\end{equation}
Numerical solutions to the above decomposed ODE system is
\begin{subequations}
\begin{equation}\label{DGD-1}
W^i(t+1) = W^i(t) -\nabla L^i\big(W(t)\big)*\eta, \ i = 1,\cdots, M.
\end{equation}
\begin{equation}\label{DGD-2}
W(t+1) = \sum_{i=1}^M \frac{N^i}{N}W^i(t).
\end{equation}
\end{subequations}
Eq.\eqref{DGD-1} is the local update rule and Eq.\eqref{DGD-2} is the global aggregation rule. These update rules guarantees the training process is equivalent to centralized GD. When local devices are updated with mini-batch gradients, i.e., 
\begin{equation}\label{DSGD-1}
W^i(t+1) = W^i(t) -g^i\big(W(t)\big)*\eta,
\end{equation} 
then this framework agrees with centralized SGD. Moreover, if only part of the devices participate the training (either full batch or mini-batch) each round and aggregate after one iteration as above, then it is also indistinguishable to centralized SGD. These optimization methods are essentially identical to centralized optimization though they are in FL framework, which theoretically are able to produce a global model as good as centralized one. However, in Eq.\eqref{DGD-1} $\nabla L^i$ must be evaluate at $W(t)$, i.e, global parameters $W(t)$ are necessary for the local update $W^i(t+1)$ at any step $t$. Therefore, local models have to aggregate after each iteration, which will cause expensive communication costs and poor efficiency. A redemption is to trade some accuracy for efficiency. To reduce communication cost,  more local iterations have to be operated each round. Accordingly, Eq. \eqref{DGD-1} is modified to following approximate decomposition of Eq.\eqref{ODE-GD} so that more local iterations are admited. 

\vspace{-0.5cm}
\begin{equation}\label{DGD-3}
\begin{split}
&\frac{d}{d\tau}W^i(\tau) = -\nabla L^i\big(W^i(\tau)\big), \\
&W^i(0) = W(0), i = 1,\cdots, M. 
\end{split}
\end{equation}
Let $T$ denote local iteration number. The corresponding numerical solution of Eq.\eqref{DGD-3} is
\begin{equation}\label{NDGD-3}
W^i(t+1) = W^i(t) - \nabla L^i(W^i(t))*\eta, t = 0,1,\cdots, T-1,
\end{equation}
which is exactly GD for $S^i$. To further accelerate local training, one may perform SGD for local devices, i.e.,
\begin{equation}\label{NDGD-4}
W^i(t+1) = W^i(t) - g^i(W^i(t))*\eta, t = 0,1,\cdots, T-1.
\end{equation}
The cost of this method is that the aggregation of $W^i$ in Eq.\eqref{DGD-3} is not the solution to Eq.\eqref{ODE-GD}. Let $\bar W(\tau) = \sum_{i=1}^M \frac{N^i}{N}W^i(\tau)$ be the aggregation of the solution to Eq.\eqref{DGD-3}. It is straightforward to check that   
\begin{equation}
\frac{d}{d\tau}\bar W(\tau) = -\sum_{i=1}^M \frac{N^i}{N}\nabla L^i(W^i(\tau)). 
\end{equation}
Comparing  the above equation with Eq.\eqref{ODE-GD}, one has 
\begin{equation}\label{diff-wbw}
\begin{split}
\frac{d}{d\tau}\big(W(\tau) - \bar W(\tau) \big) = &-\sum_{i=1}^M \frac{N^i}{N}\big(\nabla L^i(W(\tau)) \\
&- \nabla L^i(W^i(\tau)\big). 
\end{split}
\end{equation}
It is clear $W(0) = \bar W(0)$. Integrating Eq.\eqref{diff-wbw} from $0$ to $t\eta$, it arrives 
\begin{equation} \label{Estimation1}
\begin{split}
& \quad \ W(t\eta) - \bar W(t\eta) \\
&= -\int_0^{t\eta} \sum_{i=1}^M \frac{N^i}{N}\big(\nabla L^i(W(\tau)) - \nabla L^i(W^i(\tau)\big)d\tau.
\end{split}
\end{equation}
By a very rough estimate, we have that for $t=1,\cdots,T$
\begin{equation} \label{Estimation2}
|W(t\eta) - \bar W(t\eta)|\le 2\underset{i}{\sup}\|\nabla L^i\|_{L^\infty} t\eta,
\end{equation}
which yields that $\bar W(\tau)$ is a good approximation of $W(\tau)$ for $\tau \le T\eta$ when $T\eta$ is small. Therefore, FL framework has to aggregate local models periodically and reset all local and global parameters to be identical. By doing so, the aggregations of local models are always stay close to the centralized one. Since the centralized GD and SGD converge to local minimum of the total loss function, so does the one for FL. This method that performs SGD or GD on local devices and aggregates periodically is exactly the FedAvg. The above arguments also yields the underlying mechanism of FedAvg, which agrees with the essential idea of the proof of convergence of FedAvg.

Notice that the purpose of Eqs.(\ref{Estimation1}) and (\ref{Estimation2}) is to illustrate the idea of how to connect decomposition of ODEs to FL optimizations. We only used rough estimates instead of sharp ones to illustrate the underlying ideas more straightforwardly. Our \method\ method has never used the estimate in Eqs.(\ref{Estimation1}) and (\ref{Estimation2}) for attending any computations.

The superb performance of adaptive optimization methods, such as SGDM, Adam, AdaGrad, have been demonstrated for centralized training.  Naturally, successful extensions of these adaptive methods to FL are mostly desired. 

\section{Momentum Decoupling Adaptive Optimization: FedDA+SGDM}\label{sec.FedDA+SGDM}
Recall the centralized SGDM.
\begin{equation}\label{SGDm}
\begin{split}
&m(t+1)  = \beta * m(t) + (1 - \beta) *g(W(t)),\\
&W(t+1) = W(t) - \eta* m(t+1),
\end{split}
\end{equation}
where $m$ is momentum, $W$ is the parameter vector, $g$ is the mini-batch gradient and $\eta$ is the learning rate . 
The corresponding ODE is a slow-fast system
\begin{equation}\label{SGDm-ODE}
\begin{split}
&\eta\frac{d}{d\tau}m(\tau) = -(1-\beta)m(\tau) + (1-\beta)g(W(\tau)), \\
&\frac{d}{d\tau}W(\tau) = - m(\tau).
\end{split}
\end{equation}

$m$ is a fast variable cause its derivative $\frac{d}{d\tau}m(\tau)$ is $O(1/\eta)$ and $\eta$ is small. Therefore the rate of change of $m$ is much faster than that of $W$. It is easy to check Eq.\eqref{SGDm} is the numerical solution of Eq.\eqref{SGDm-ODE} using Euler's scheme. 

Similar to GD case, the numerical solution of a precise decomposition of SGDM is
\begin{equation}
\begin{split}
&m^i(t+1)  = \beta * m^i(t) + (1 - \beta) *g^i(W(t)),\\
&W^i(t+1) = W^i(t) - \eta* m^i(t+1),\\
&m(t+1) = \sum_{i=1}^M \frac{N^i}{N}m^i(t),\\ 
&W(t+1) = \sum_{i=1}^M \frac{N^i}{N}W^i(t). 
\end{split}
\end{equation}
Again, the drawback is the local iteration number must be $1$. To allow more local iterations, the most naive approach is to perform SGDM on local devices directly
\begin{equation}\label{Naive}
\begin{split}
&m^i(t+1)  = \beta * m^i(t) + (1 - \beta) * g^i(W^i(t)),\\
&W^i(t+1) = W^i(t) - \eta* m^i(t+1).\\
\end{split}
\end{equation}
Then aggregate the above local updates periodically. This method is indeed an approximation of centralized SGDM for the same reason as in FedAvg case, and therefore this naive method may converge. However, the performance of this naive method is not satisfying in experiments. The reasons are twofold. First, the momentum is used to buff the deviation of gradients. Decomposing the global momentum $m$ onto local devices  as in Eq.\eqref{Naive} will incapacitate it because $\sum_{i=1}^M \frac{N^i}{N}m^i(t)$ is deviated from the momentum of centralized optimizer. Second, since the aggregated momentum is not accurate enough, it may also harm the local training if it is inherited by the local devices. This is why the restart local adaptive method works better~\cite{WXGC21}. Therefore, though the Eq.\eqref{Naive} could be an approximation of centralized SGDM, it may need a rather small learning rate to ensure the quality of approximation, which results in unsatisfactory training performance. Local training generates local gradients deviating from the global ones, therefore a global momentum is favorable to control the deviations of the gradients uploaded by local devices. FedOpt updates global momentum one time after the server receiving local gradients generated by several local iterations each round. Though it achieves great performance, it does not fully take the advantage of global momentum. In FedOpt the global momentum buffs the total gradients of multiple iterations. However, for centralized SGDM, the momentum buffs the gradient in each iteration. To let the global momentum fully plays its role and accelerate the training of the global model, we propose a decoupling adaptive technique (FedDA) to approximately mimic centralized SGDM. In our FedDA method, though the global momentum does not participate local training directly, it buffers local gradients generated in each local iteration. 
We propose the following decoupled decomposition of Eq.\eqref{SGDm-ODE}.

\begin{equation}\label{SGDm-Decouple}
\begin{split}
&\frac{d}{d\tau}W^i(\tau) = - g^i(W^i(\tau)),\\
&\eta\frac{d}{d\tau}m(\tau) = -(1-\beta)m(\tau) + (1-\beta) \sum_{i=1}^M \frac{N^i}{N} g^i(W^i(\tau))\\
&\frac{d}{d\tau}W(\tau) = - \alpha * m(\tau),
\end{split}
\end{equation}
where $W^i$ is the parameters of local devices, $m$ is the global momentum, $W$ is the parameters of global model and $\alpha$ is a parameter to concede that local and global employ different learning rates. Note that in Eq.\eqref{SGDm-Decouple}, the local update of $W^i$ is independent of the global momentum $m$, this is why we name this method decoupling method. The crucial point of our decoupling method is that since the the equations of $W^i(\tau)$ are independent of the global momentum $m$, the equation of $m$ is totally linear and can be precisely decomposed as $m(\tau) = \sum_{i=1}^M \frac{N^i}{N} m^i(\tau) $, where $m^i\tau$ solves
\[\eta\frac{d}{d\tau}m^i(\tau) = -(1-\beta)m^i(\tau) + (1-\beta) \nabla g^i(W^i(\tau)).\] 
Therefore, the system in Eq.\eqref{SGDm-Decouple} is exactly equivalent to 
\begin{equation}\label{SGDm-Decouple2}
\begin{split}
&\frac{d}{d\tau}W^i(\tau) = - g^i(W^i(\tau)),\\
&\eta\frac{d}{d\tau}m^i(\tau) = -(1-\beta)m^i(\tau) + (1-\beta)g^i(W^i(\tau)),\\
&\frac{d}{d\tau}W(\tau) = - \alpha * m(\tau).
\end{split}
\end{equation}
That is we can calculate the global momentum $m$ by precisely decomposing it to local devices and the aggregation of the portion of the global momentum on local devices $m^i$ will be exactly the global momentum. The numerical solution to Eq.\eqref{SGDm-Decouple2} is
\begin{align}
&W^i(t+1) = W^i(t) - g^i(W^i(t))*\eta, \label{n1}\\
&m^i(t+1) = \beta*m^i(t) + (1-\beta)*g^i(W^i(t))\eta,\\
&m(t+1) = \sum_{i=1}^M \frac{N^i}{N}m^i(t), \label{n2}\\
&W(t+1) = W(t) - m(t+1)*\alpha*\eta \label{n3}. 
\end{align}

\begin{table*}[t]
\caption{Statistics of the Datasets}
\begin{center}
\begin{tabular}{l|cccc}
\hline {\bf Dataset} & {\bf \#Training Clients} & {\bf \#Test Clients} & {\bf \#Training Samples} & {\bf \#Test Samples} \\
\hline {\bf CIFAR-100} & 500 & 100 & 50,000 & 10,000 \\
\text {\bf EMNIST-62} & 3,400 & 3,400 & 671,585 & 77,483 \\
\text {\bf Stack Overflow} & 342,477 & 204,088 & 135,818,730 & 16,586,035 \\
\hline
\end{tabular}
\label{tbl.Sex}
\end{center}
\end{table*}

This numerical solution demonstrates our underlying ideas better. Eq.\eqref{n1} yields that local update does not depend on the global momentum $m$. However, the global momentum buffs local gradients in each local iteration as shown in Eq.\eqref{n2}.  Based on the numerical solution Eqs.\eqref{n1}-\eqref{n3}, the algorithm of our FedDA+SGDM method is as follows. 

At round $E$, pick participating clients $S^1,\cdots, S^k$. For each $S^i$, $i = 1,\cdots, K$, initialize $P^i=0$, $W^i(0) = W(E)$, $m^i(0) = m(E)$. For $t = 0, T-1$,
\begin{equation}
\begin{split}
&W^i(t+1) = W^i(t) -  g^i(W^i(t))*\eta,\\
&m^i(t+1) = \beta m^i(t) + (1-\beta) g^i(W^i(t)), \\
& P^i = P^i + m^i(t+1).
\end{split}
\end{equation}
The global update rule is:
\begin{equation} \label{Update}
\begin{split}
&P = \text{aggregation of }\; P^i,\\
&m(E+1) = \text{aggregation of}\; m^i(T),\\
&W (E+1)= W(E) - P*\alpha*\eta.
\end{split}
\end{equation}
Notice that $P$ is the summation of global momentums updated during local iterations, and global parameter $W$ is updated by $P$. This mimics the centralized SGDM.
We theoretically demonstrate that (1) local momentum deviates from the centralized one at exponential rate $O(e^{\lambda t})$; and (2) global momentum in FedDA deviates from the centralized one at algebraic rate $O(t^2)$. Please refer to Appendix~\ref{sec.Advantage} for detailed proof.

{\bf Full batch gradient stabilization.} Adaptive optimization methods for FL are often faced with inconsistency of convergence in training ~\cite{WXGC21}. When the training is almost finished, i.e, $W$ is close to the local minimum point of loss function, the gradient has to be more accurate to ensure the convergence to local minimum point. Not sufficiently accurate gradients will lead to unstable convergence and inconsistency.  Based on our framework, if local devices only iterate once with full batch gradients in a training round, then our FedDA method is in agreement with centralized training in that round. Therefore, by reducing the iteration number to $1$ in the end of training, our FedDA method mimics centralized training, which ensures the convergence and overcomes the inconsistency of training. Last but not least, since this full batch method is utilized only in the end of the training when the parameter $W$ is fairly close to local minimum point, the training converges and stabilized rather quickly. Therefore, it is a bargain to employ this full batch method in the end of the training to guarantee the convergence of the training to local minimum point.

Our proposed momentum decoupling adaptive optimization method has potential be extended to other federated learning algorithms, e.g., FedProx, etc. Due to space limit, we only use FedDA+SGDM as an example to show how to apply the FedDA optimization to FedProx. At round $E$, pick participating clients $S^1,\cdots, S^k$. For each $S^i$, $i = 1,\cdots, K$, initialize $P^i=0$, $W^i(0) = W(E)$, $m^i(0) = m(E)$. For $t = 0, \cdots, T-1$, $W^i(t+1) = W^i(t) -  \big(g^i(W^i(t))+\mu*(W^i(t)-W(E)\big)*\eta, m^i(t+1) = \beta*m^i(t) + (1-\beta)*\big(g^i(W^i(t)) + \mu*(W^i(t)-W(E)\big), P^i = P^i + m^i(t+1)$. The global update rule is the same as the one in Eq.(\ref{Update}). By following the similar strategy, other two versions of FedDA (FedDA+ADAM and FedDA+AdaGrad) can be added to FedProx and other regular FL algorithms, which demonstrates the applicability and generality of FedDA.

\vspace{-0.2cm}
\section{Experiments}\label{sec.experiment}
\vspace{-0.25cm}
\begin{figure*}[!]
\centering
\begin{minipage}{0.496\linewidth}
\begin{figure}[H]
\mbox{
\hspace{-0.6cm}
\subfigure[SGDM]{\epsfig{figure=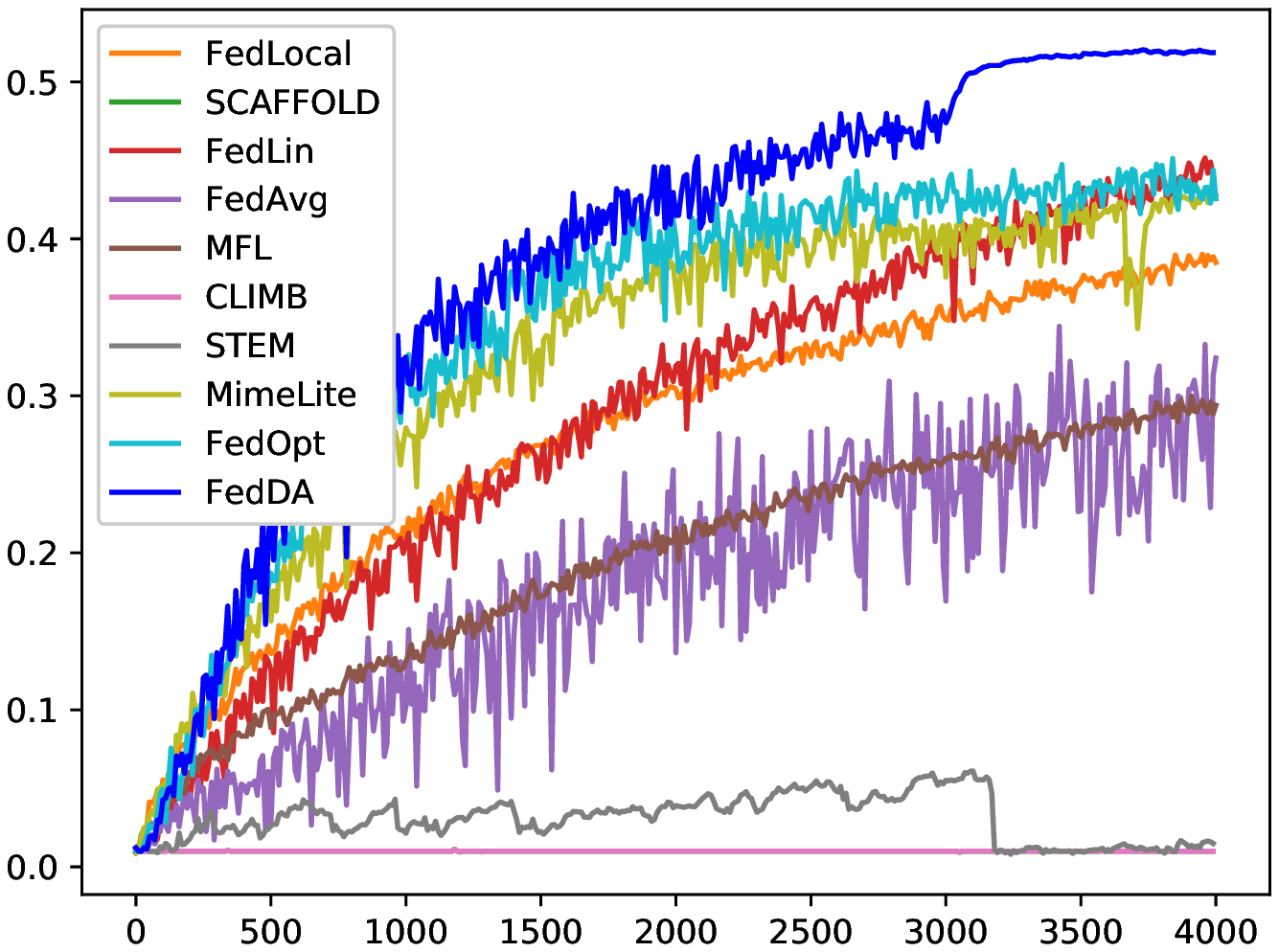, height=1.35in, width=0.38\linewidth}} \hspace{-0.475cm}
\subfigure[Adam]{\epsfig{figure=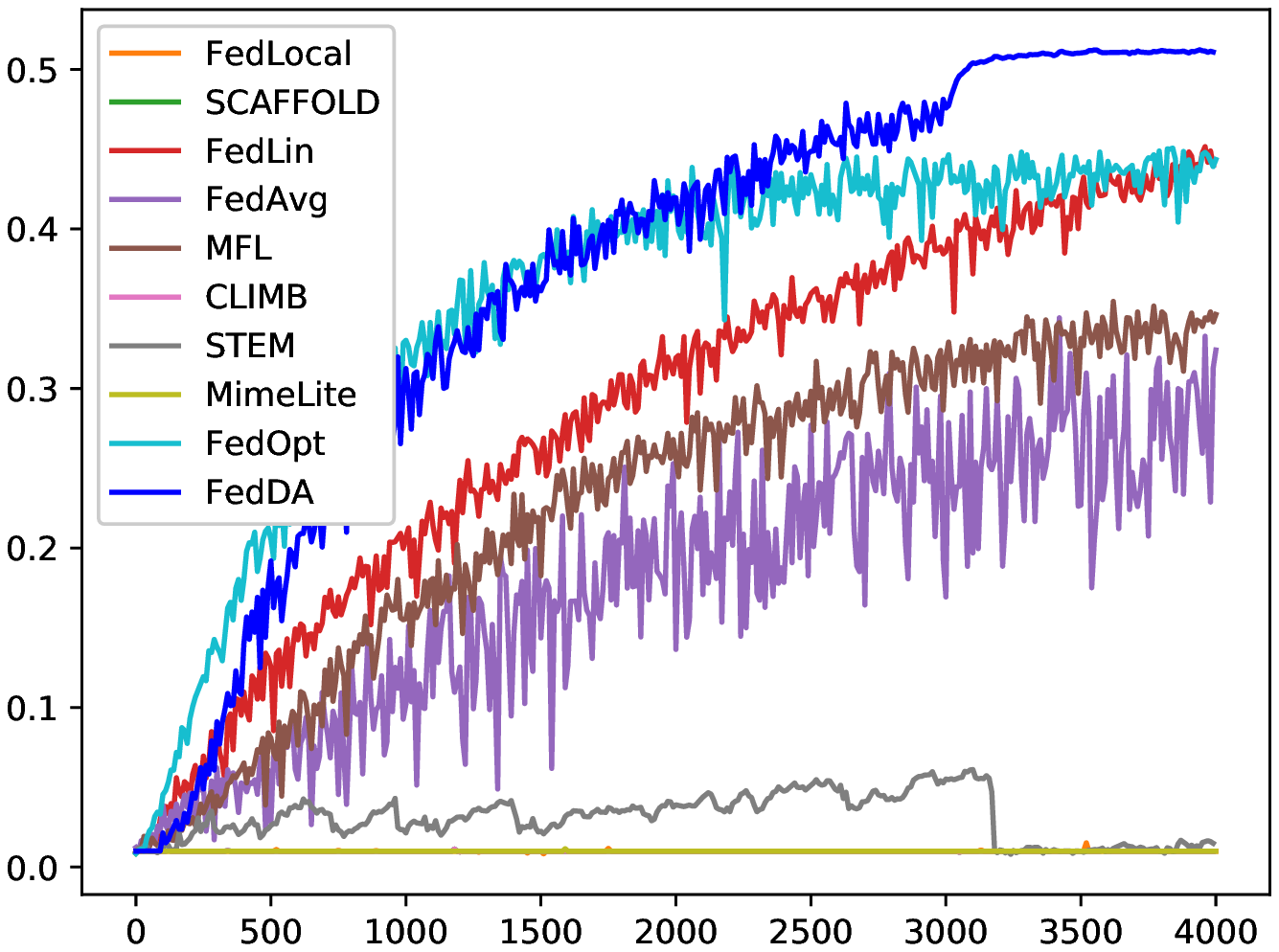, height=1.35in, width=0.38\linewidth}} \hspace{-0.475cm}
\subfigure[AdaGrad]{\epsfig{figure=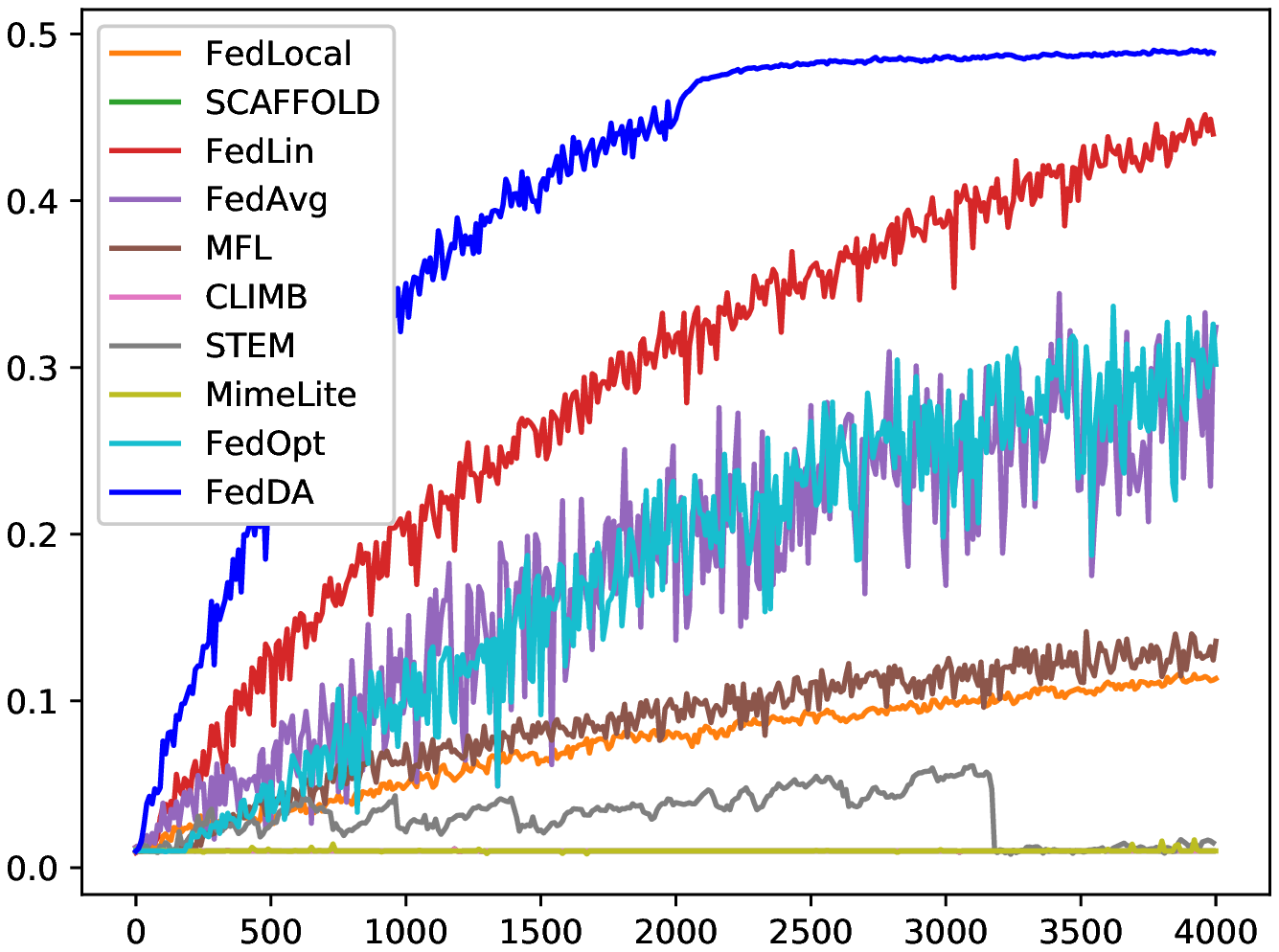, height=1.35in, width=0.38\linewidth}}}
\caption{Convergence on CIFAR-100 with Three Optimizers}
\label{fig.CIFARConvergence}
\end{figure}
\end{minipage}
\begin{minipage}{0.496\linewidth}
\begin{figure}[H]
\mbox{
\hspace{-0.3cm}
\subfigure[SGDM]{\epsfig{figure=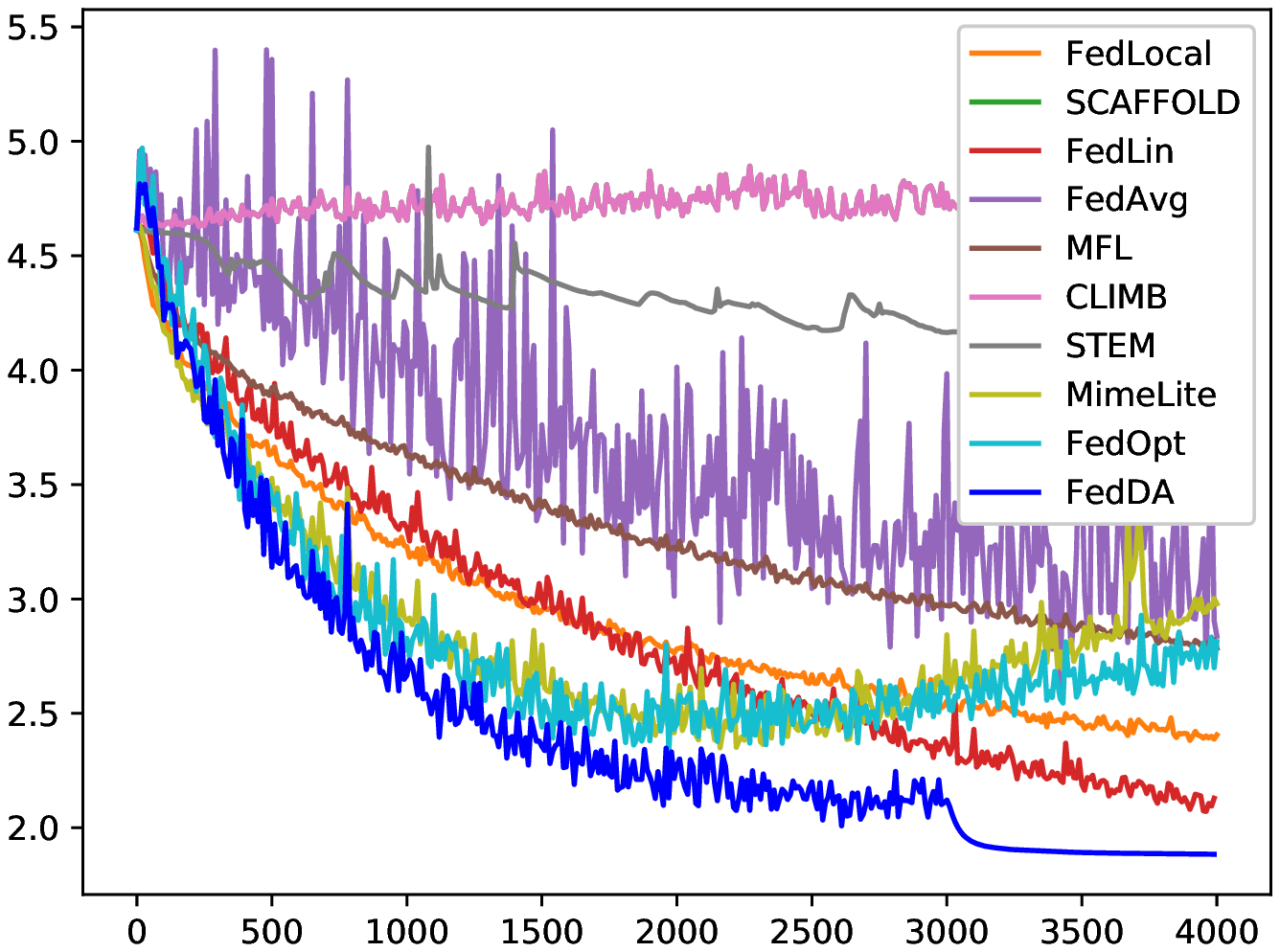, height=1.35in, width=0.38\linewidth}} \hspace{-0.475cm}
\subfigure[Adam]{\epsfig{figure=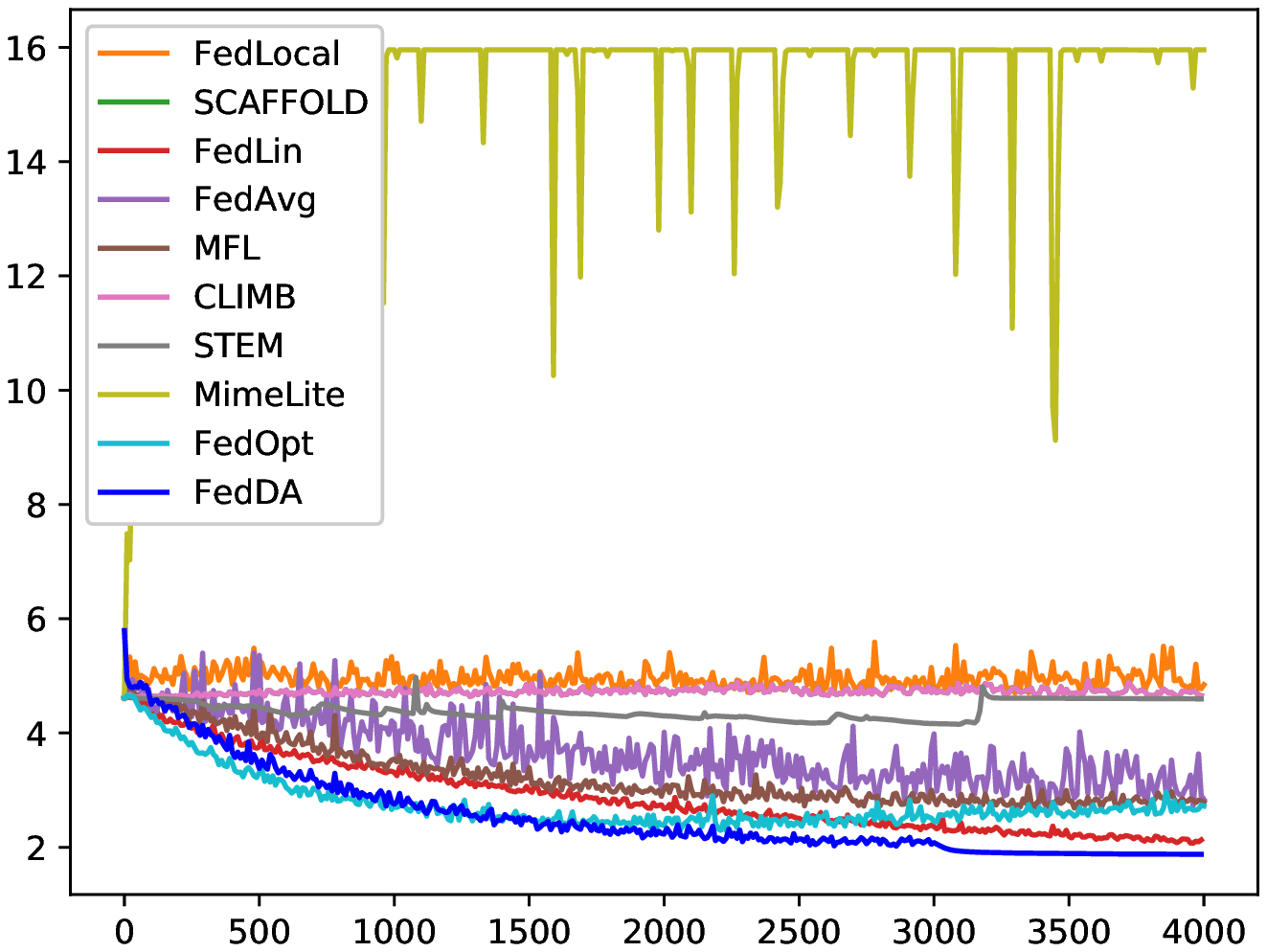, height=1.35in, width=0.38\linewidth}} \hspace{-0.475cm}
\subfigure[AdaGrad]{\epsfig{figure=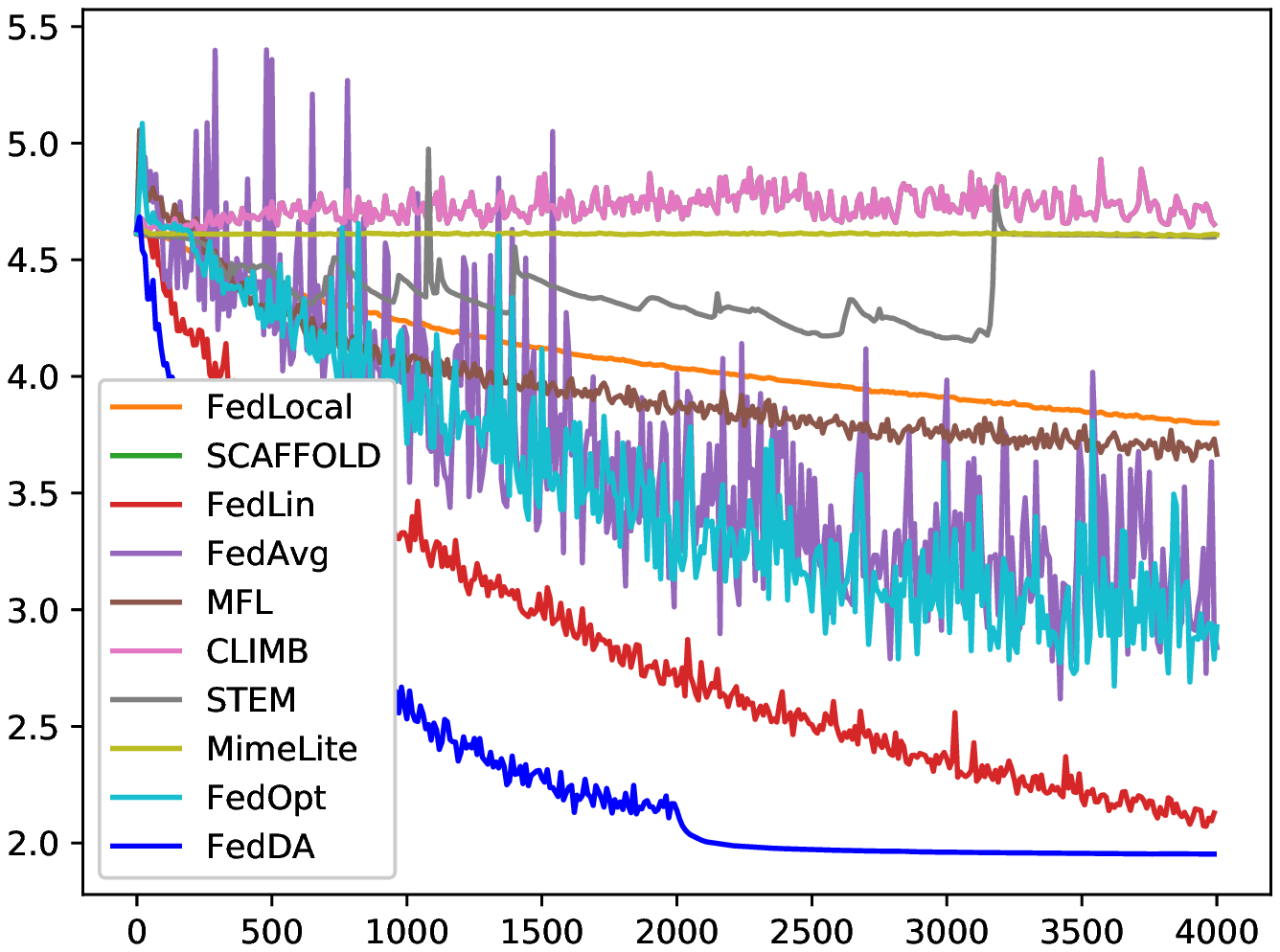, height=1.35in, width=0.38\linewidth}}}
\caption{Loss on CIFAR-100 with Three Optimizers}
\label{fig.CIFARLoss}
\end{figure}
\end{minipage}
\end{figure*}

\begin{figure*}[!]
\centering
\begin{minipage}{0.496\linewidth}
\begin{figure}[H]
\mbox{
\hspace{-0.6cm}
\subfigure[SGDM]{\epsfig{figure=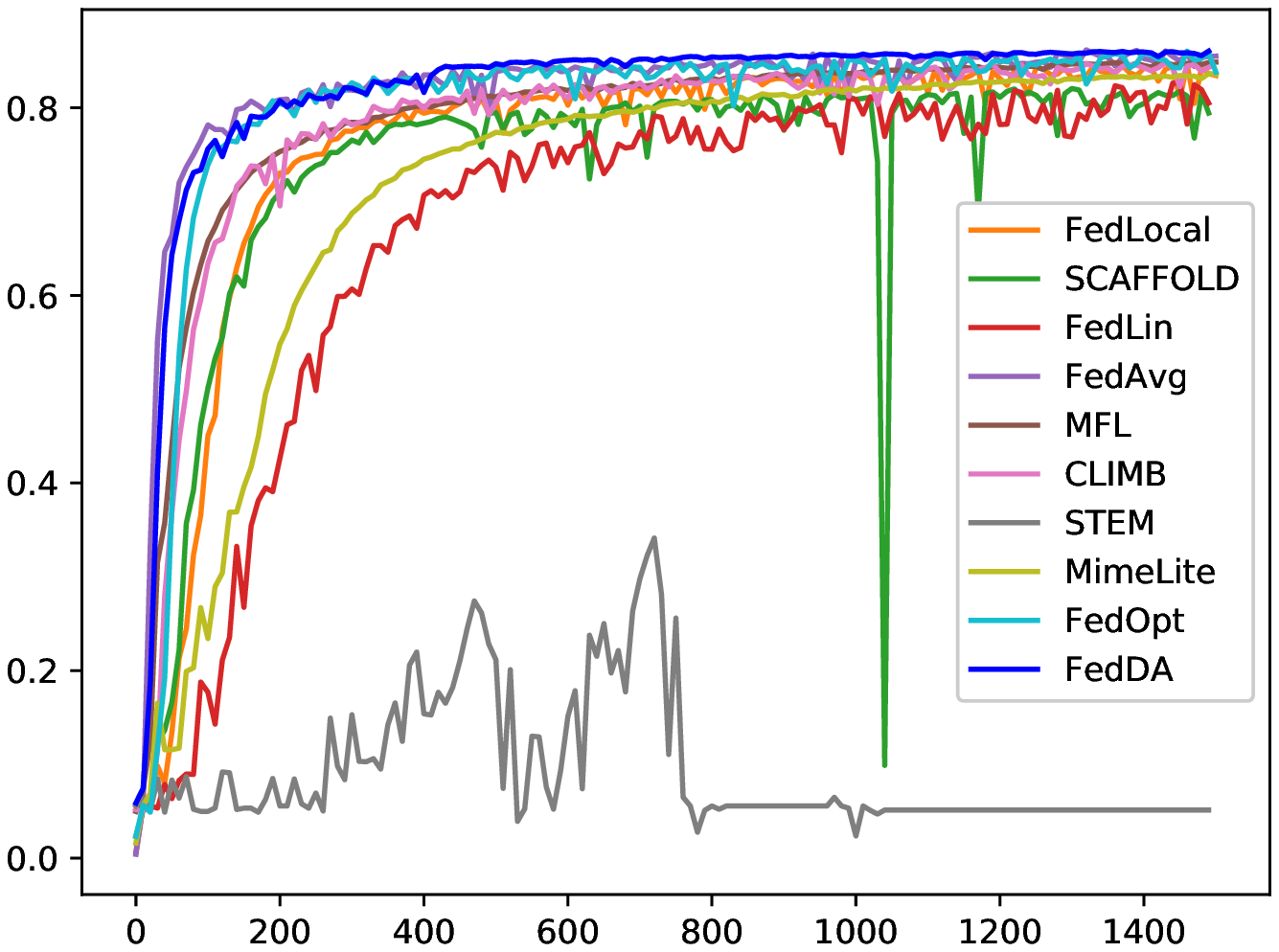, height=1.35in, width=0.38\linewidth}} \hspace{-0.475cm}
\subfigure[Adam]{\epsfig{figure=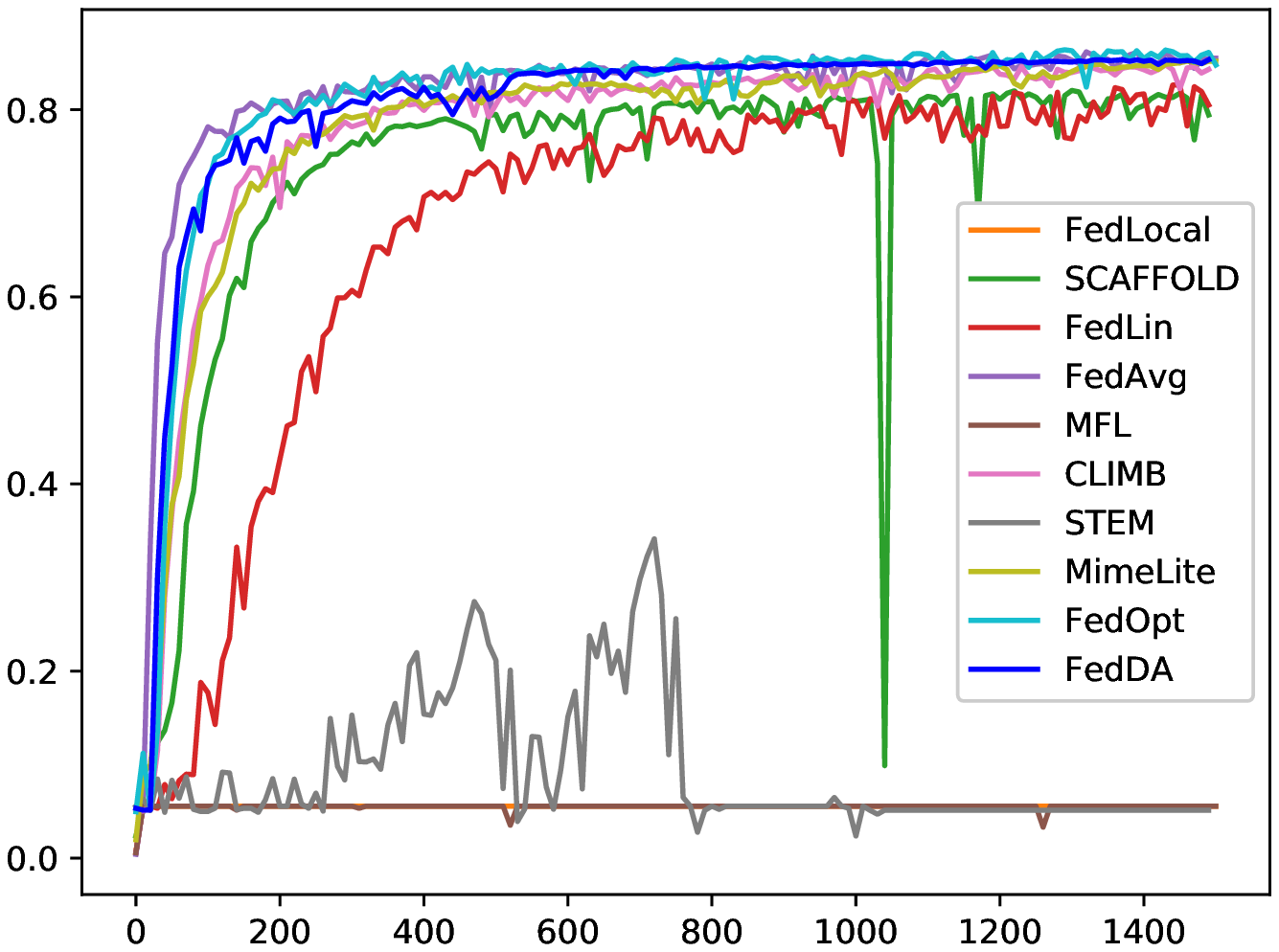, height=1.35in, width=0.38\linewidth}} \hspace{-0.475cm}
\subfigure[AdaGrad]{\epsfig{figure=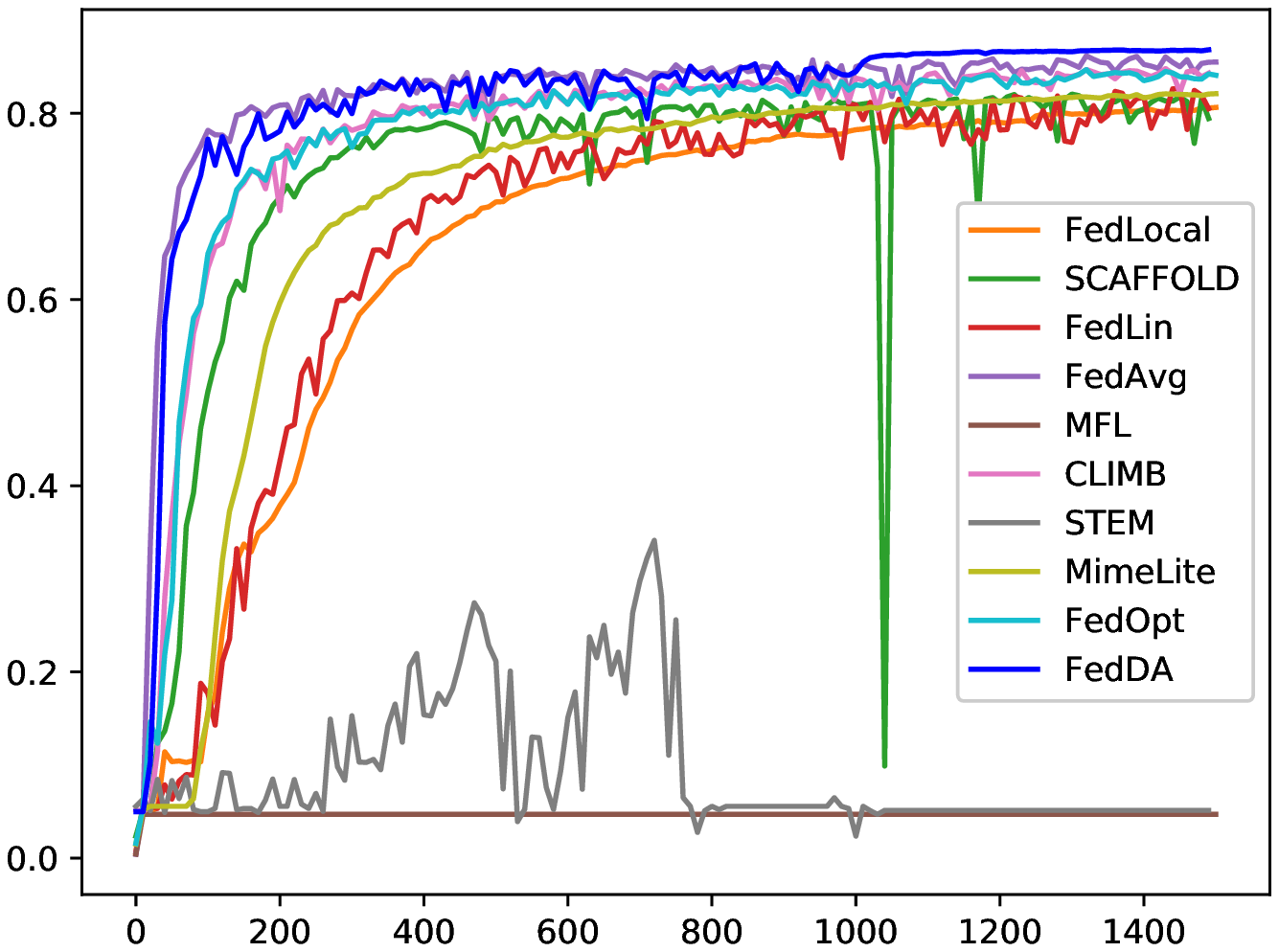, height=1.35in, width=0.38\linewidth}}}
\caption{Convergence on EMNIST with Three Optimizers}
\label{fig.EMNISTConvergence}
\end{figure}
\end{minipage}
\begin{minipage}{0.496\linewidth}
\begin{figure}[H]
\mbox{
\hspace{-0.3cm}
\subfigure[SGDM]{\epsfig{figure=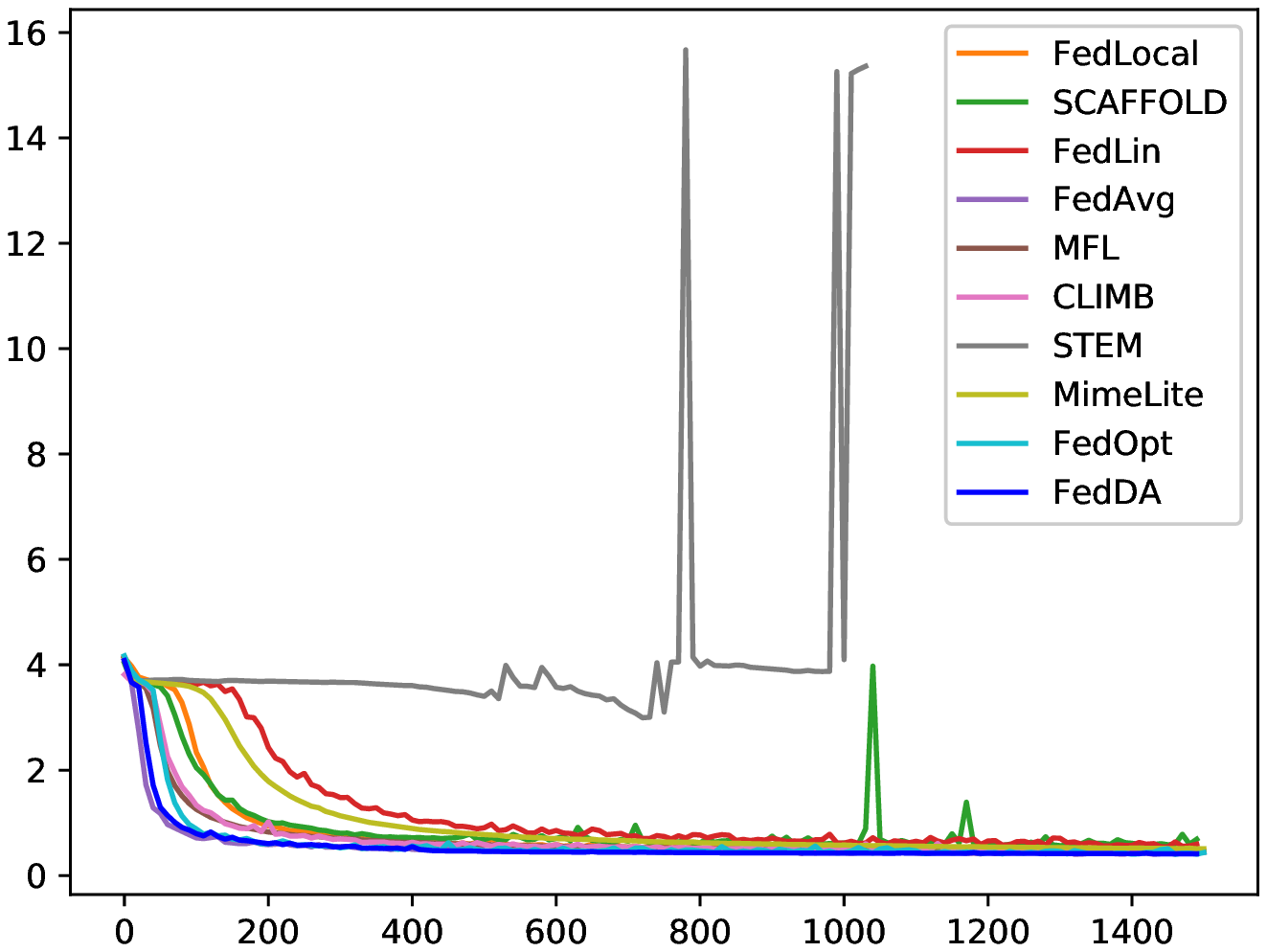, height=1.35in, width=0.38\linewidth}} \hspace{-0.475cm}
\subfigure[Adam]{\epsfig{figure=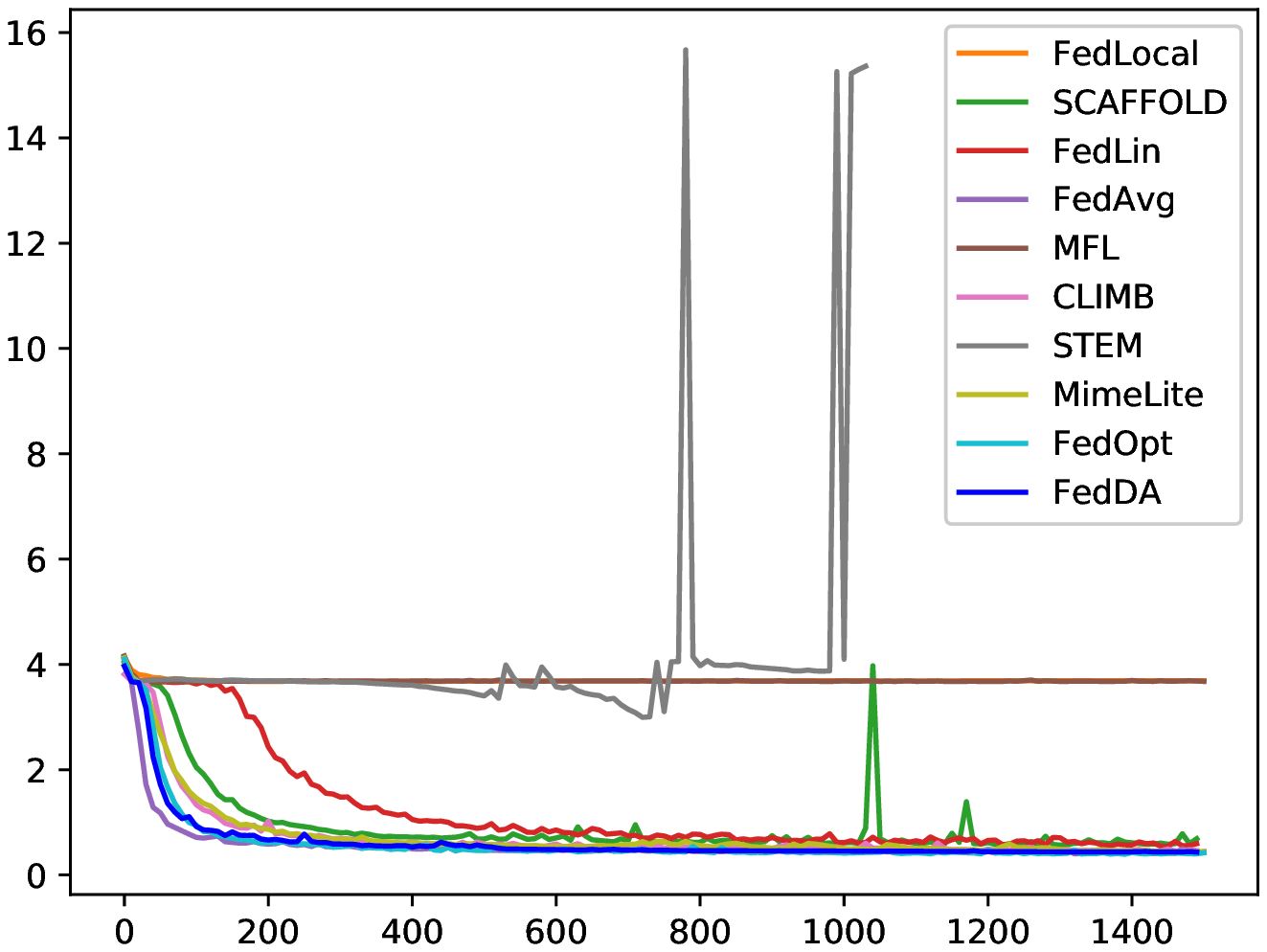, height=1.35in, width=0.38\linewidth}} \hspace{-0.475cm}
\subfigure[AdaGrad]{\epsfig{figure=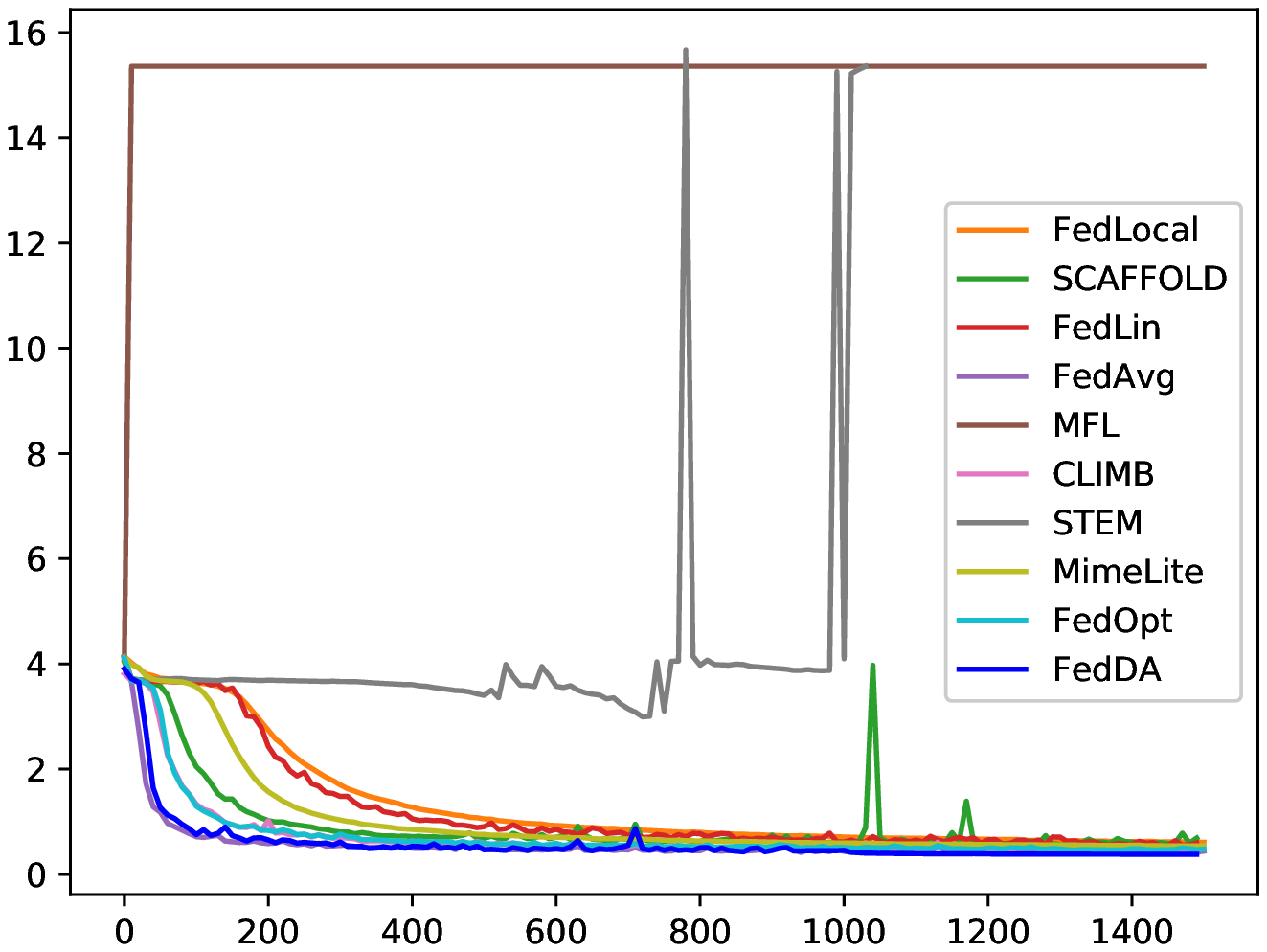, height=1.35in, width=0.38\linewidth}}}
\caption{Loss on EMNIST with Three Optimizers}
\label{fig.EMNISTLoss}
\end{figure}
\end{minipage}
\end{figure*}

In this section, we have evaluated the performance of our \method\ model and other comparison methods in serval representative federated datasets and learning tasks to date. We show that \method\ with decoupling and full batch gradient techniques is able to achieve faster convergence and higher test accuracy in cross-device settings against several state-of-the-art federated optimization methods. The experiments exactly follow the same settings described by a recent federated optimization method, FedOpt~\cite{RCZG21}.

{\bf Datasets.} We focus on three popular computer vision and natural language processing tasks over three representative benchmark datasets respectively: (1) image classification over CIFAR-100~\cite{Kriz09}. We train ResNet-18 by replacing batch norm with group norm~\cite{HPMG20}; (2) image classification over EMNIST~\cite{HPMG20}. We train a CNN for character recognition; and (3) text classification over Stack Overflow~\cite{SO19}. We perform tag prediction using logistic regression on bag-of-words vectors. We select the 10,000 most frequently used words, the 500 most frequent tags and a one-versus-rest classification strategy, by following the same strategy in FedOpt~\cite{RCZG21}. The detailed descriptions of the federated datasets and learning tasks are presented in Appendix~\ref{sec.ExperimentDetails}.

{\bf Baselines.} We compare the \method\ model with nine state-of-the-art federated learning models, including five regular federated learning and four federated optimization approaches.
{\bf FedAvg} is a classical as well as practical method for the federated learning of deep networks based on iterative model averaging~\cite{MMRH17}.
{\bf SCAFFOLD} is a algorithm which uses control variates to correct for the client-drift in its local updates~\cite{KKMR20}.
{\bf FedLin} is an algorithmic framework to tackle the key challenges of objective heterogeneity, systems heterogeneity, and imprecise communication in FL~\cite{MJPH21}.
{\bf STEM} is a stochastic two-sided momentum algorithm, that utilizes certain momentum-assisted stochastic gradient directions for both the client and  server updates~\cite{MJPH21}.
{\bf CLIMB} is an agnostic constrained learning formulation to tackle the class imbalance problem in FL without requiring further information beyond the standard FL objective~\cite{Anon22d}.
{\bf MFL} performs momentum gradient descent in local update step of FL system to solve the distributed machine learning problem~\cite{LCCZ20}.
{\bf FedOpt} is a flexible algorithmic framework that allows the clients and the server to choose different optimization methods more general than stochastic gradient descent (SGD) in FedAvg~\cite{RCZG21}.
{\bf MimeLite} uses a combination of control-variates and server-level optimizer state at every client-update step to ensure that each local update mimics that of the centralized method run on i.i.d. data~\cite{KJKM21}.
{\bf Local Adaptivity (FedLocal)} proposes techniques that enable the use of adaptive optimization methods for local updates at clients in federated learning~\cite{WXGC21}.
To our best knowledge, this work is the first to certify the group fairness of classifiers with theoretical input-agnostic guarantees, while there is no need to know the shift between training and deployment datasets with respect to sensitive attributes. All nine baselines used in our experiments either do not use momentum, or use momentum without momentum aggregation, or use momentum with aggregation but restart momentum at each FL round (FedLocal). Our \method\ method keeps the momentum aggregation in the entire FL process, which results in faster convergence but larger oscillation.

{\bf Evaluation metrics.} We use two popular measures in federated learning and plot the measure curves with increasing training rounds to verify the convergence of different methods: {\bf Accuracy} and {\bf Loss Function Values (Loss)}~\cite{KKMR20,MJPH21,LCCZ20,RCZG21,KJKM21,WXGC21}. A larger Accuracy or a smaller Loss score indicates a better federated learning result. We run 1,500 rounds of training on the EMNIST and Stack Overflow datasets, and 4,000 rounds over the CIFAR-100 dataset. In addition, we use final Accuracy to evaluate the quality of the federated learning algorithms.

\begin{figure*}[!]
\centering
\begin{minipage}{0.55\linewidth}
\begin{table}[H]
\caption{Final Accuracy on CIFAR-100}
\begin{center}
\begin{tabular}{l|ccc}
\hline {\bf Optimizer} & {\bf SGDM} & {\bf Adam} & {\bf AdaGrad} \\
\hline FedLocal & 0.384  & 0.009  & 0.113 \\
SCAFFOLD & 0.010  & 0.010  & 0.010 \\
FedLin & 0.440  & 0.440  & 0.440 \\
FedAvg & 0.324  & 0.324  & 0.324 \\
MFL & 0.293  & 0.346  & 0.135 \\
CLIMB & 0.010  & 0.010  & 0.010 \\
STEM & 0.014  & 0.014  & 0.014 \\
MimeLite & 0.427  & 0.009  & 0.009 \\
FedOpt & 0.425  & 0.443  & 0.301 \\
\hline FedDA & {\bf 0.518}  & {\bf 0.510}  & {\bf 0.488} \\
\hline
\end{tabular}
\label{tbl.FinalAccuracyCIFAR}
\end{center}
\end{table}
\end{minipage}
\begin{minipage}{0.44\linewidth}
\begin{table}[H]
\vspace{-0.3cm}
\caption{Final Accuracy on EMNIST}
\begin{center}
\begin{tabular}{l|ccc}
\hline {\bf Optimizer} & {\bf SGDM} & {\bf Adam} & {\bf AdaGrad} \\
\hline FedLocal & 0.834  & 0.055  & 0.806 \\
SCAFFOLD & 0.794  & 0.794  & 0.794 \\
FedLin & 0.805  & 0.805  & 0.805 \\
FedAvg & 0.850  & 0.850  & 0.850 \\
MFL & 0.848  & 0.055  & 0.047 \\
CLIMB & 0.843  & 0.843  & 0.843 \\
STEM & 0.051  & 0.051  & 0.051 \\
MimeLite & 0.835  & 0.851  & 0.821 \\
FedOpt & 0.838  & 0.847  & 0.840 \\
\hline FedDA & {\bf 0.860}  & {\bf 0.853}  & {\bf 0.868} \\
\hline
\end{tabular}
\label{tbl.FinalAccuracyEMNIST}
\end{center}
\end{table}
\end{minipage}
\end{figure*}

{\bf Convergence on CIFAR-100 and EMNIST.} Figures \ref{fig.CIFARConvergence} and \ref{fig.EMNISTConvergence} exhibit the $Accuracy$ curves of ten federated learning models for image classification over CIFAR-100 and character recognition on EMNIST respectively.
It is obvious that the performance curves by federated learning algorithms initially keep increasing with training rounds and  remains relatively stable when the curves are beyond convergence points, i.e., turning points from a sharp $Accuracy$ increase to a flat curve. This phenomenon indicates that most federated learning algorithms are able to converge to the invariant solutions after enough training rounds. However, among five regular federated learning and five federated optimization approaches, our \method\ federated optimization method can significantly speedup the convergence on two datasets in most experiments, showing the superior performance of \method\ in federated settings. Compared to the learning results by other federated learning models, based on training rounds at convergence points, \method, on average, achieves 34.3\% and 22.6\% convergence improvement on two datasets respectively.

\begin{table}[t]
\caption{Final Mean Squared Error on EMNIST for Autoencoder}
\begin{center}
\begin{tabular}{l|ccc}
\hline {\bf Optimizer} & {\bf SGDM} & {\bf Adam} & {\bf AdaGrad} \\
\hline FedLocal & 0.0169	 & 0.0289 & 0.0168 \\
FedAvg & 0.0171 & 0.0171 & 0.0171 \\
MFL & 0.0168	 & 0.0290 & 0.0291 \\
MimeLite & 0.0183 & 0.0307 & 0.0287 \\
FedOpt & 0.0175 & 0.0173 & 0.0145 \\
\hline FedDA & {\bf 0.0167}  & {\bf 0.0166}  & {\bf 0.0132} \\
\hline
\end{tabular}
\label{tbl.FinalAccuracyEMNIST1}
\end{center}
\end{table}

{\bf Loss on CIFAR-100 and EMNIST.} Figures \ref{fig.CIFARLoss} and \ref{fig.EMNISTLoss} present the $Loss$ curves achieved by ten federated learning models on two datasets respectively.
We have observed obvious that the reverse trends, in comparison with the $Accuracy$ curves. In most experiments, our \method\ federated optimization method is able to achieve the fastest convergence, especially, \method\ can converge within less than 200 training rounds and then always keep stable on the EMNIST dataset. A reasonable explanation is that \method\ fully utilizes the global momentum on each local iteration as well as employ full batch gradients to mimic centralized optimization in the end of the training process for accelerating the training convergence.

{\bf Final $Accuracy$ on CIFAR-100 and EMNIST.} Tables \ref{tbl.FinalAccuracyCIFAR}-\ref{tbl.FinalAccuracyEMNIST1} show the quality of ten federated learning algorithms over CIFAR-100 and EMNIST respectively. 
Notice that the performance achieved by five regular federated learning algorithms, including FedAvg, SCAFFOLD, FedLin, STEM, and CLIMB, keep unchanged when using different optimizers, such as SGDM, Adam, and AdaGrad, while five federated optimization approaches, including MFL, FedOpt, Mime, FedLocal, and our \method\ model obtain different performance.
We have observed that our \method\ federated optimization solution outperforms the competitor methods in most experiments. \method\ achieves the highest $Accuracy$ values ($>$ 0.488 over CIFAR-100 and $>$ 0.853 on EMNIST respectively), which are better than other nine baseline methods in all tests.
A reasonable explanation is that the combination of decoupling and full batch gradient techniques is able to achieve faster convergence as well as higher test accuracy in cross-device settings. In addition, the promising performance of \method\ over both datasets implies that \method\ has great potential as a general federated optimization solution to learning tasks over federated datasets, which is desirable in practice.

\setlength{\textfloatsep}{0.3pt}
\begin{figure*}[!]
\centering
\begin{minipage}{0.496\linewidth}
\begin{figure}[H]
\mbox{
\hspace{-0.2cm}
\subfigure[CIFAR-100]{\epsfig{figure=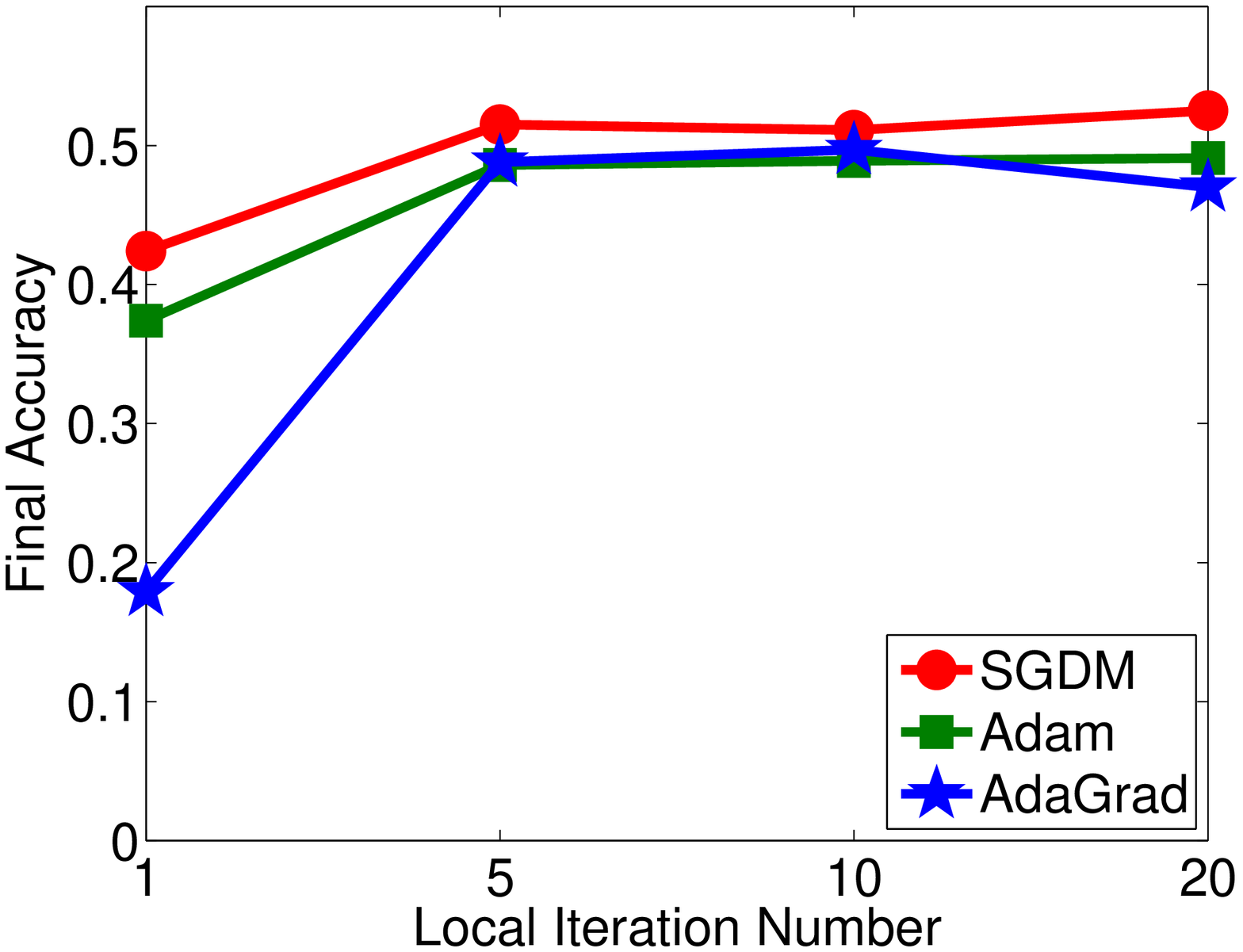, height=1.35in, width=0.34\linewidth}} \hspace{-0.225cm}
\subfigure[EMNIST]{\epsfig{figure=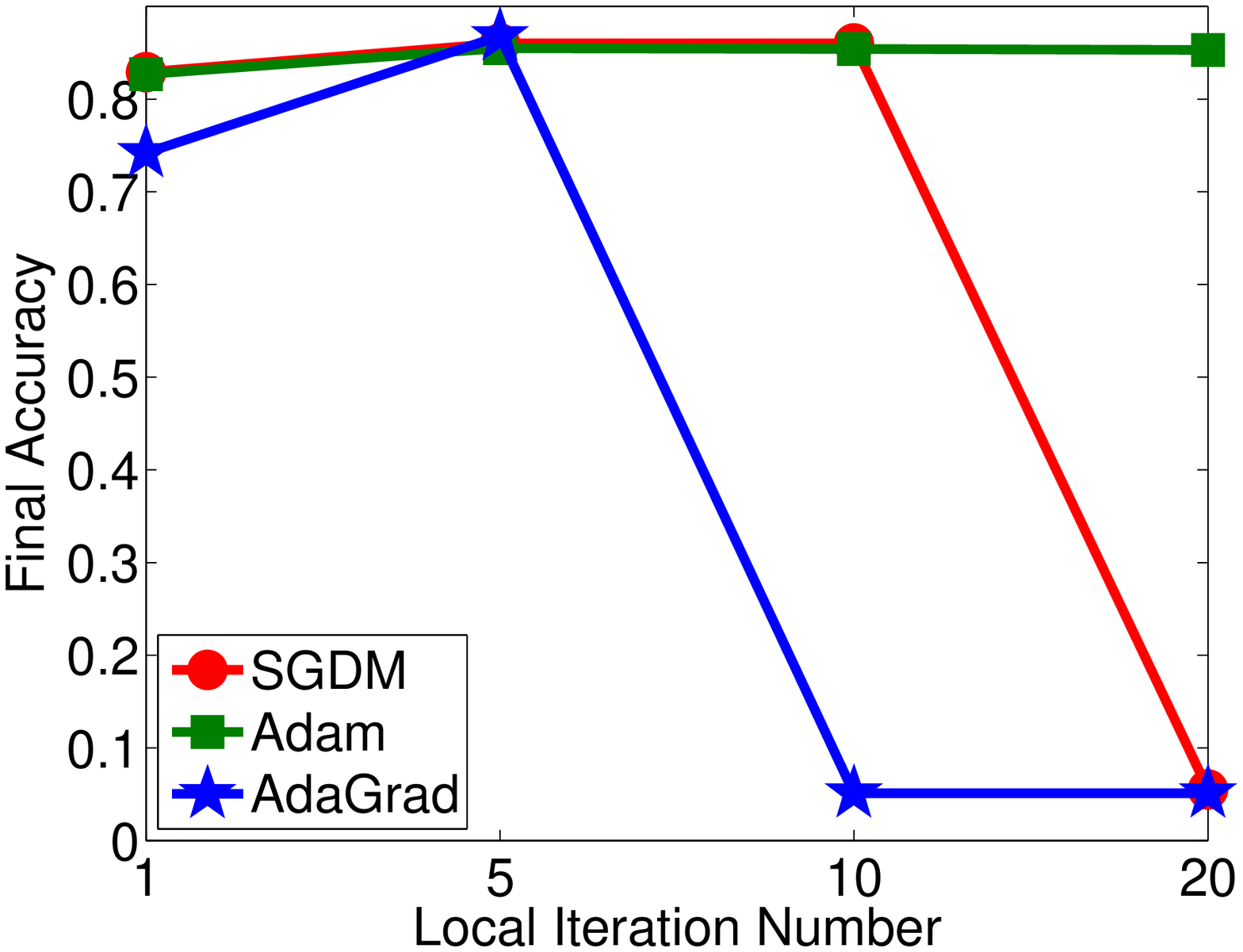, height=1.35in, width=0.34\linewidth}} \hspace{-0.225cm}
\subfigure[Stack Overflow]{\epsfig{figure=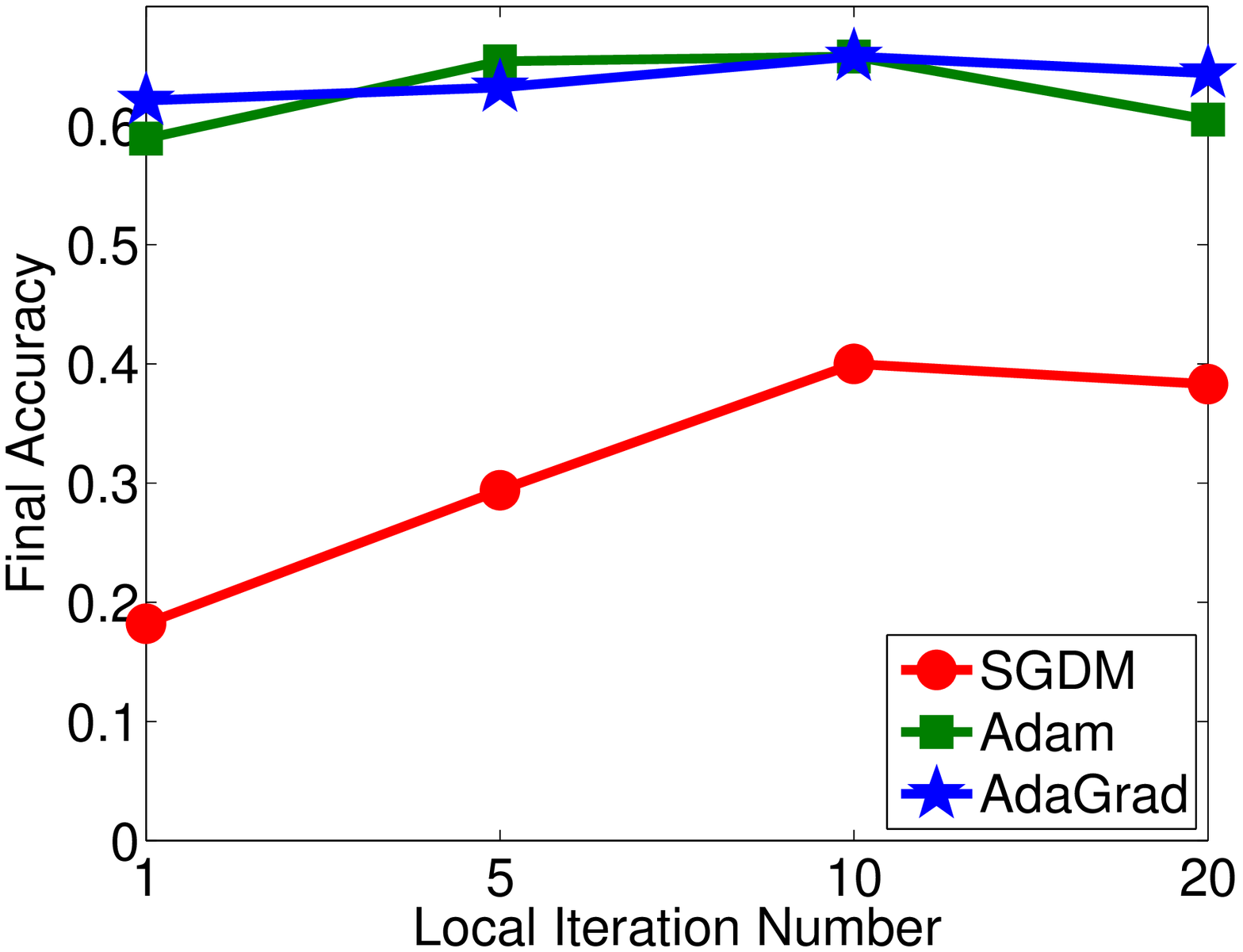, height=1.35in, width=0.34\linewidth}}}
\caption{Final Accuracy with Varying \#Local Iterations}
\label{fig.Local}
\end{figure}
\end{minipage}
\begin{minipage}{0.496\linewidth}
\begin{figure}[H]
\mbox{
\subfigure[CIFAR-100]{\epsfig{figure=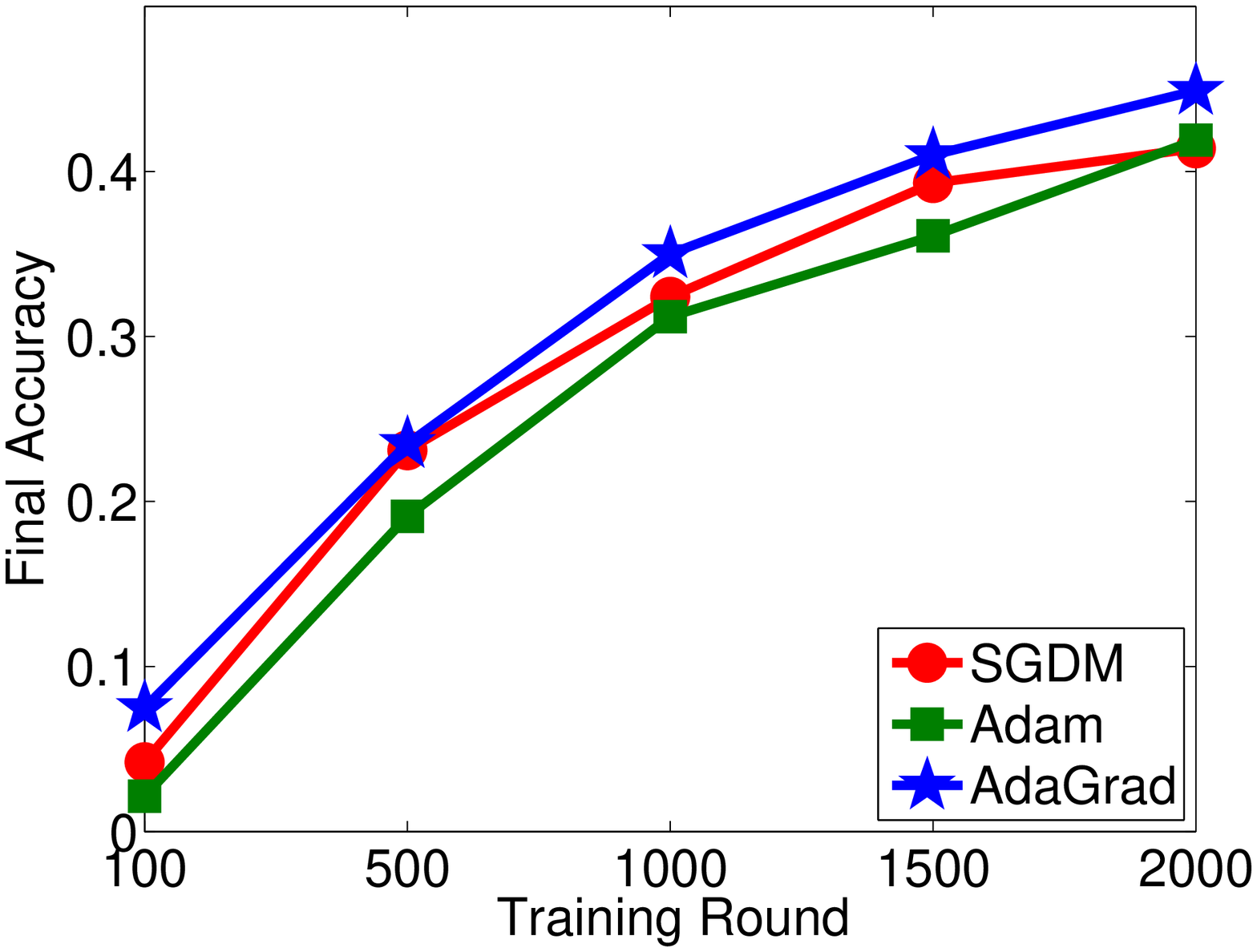, height=1.35in, width=0.34\linewidth}} \hspace{-0.225cm}
\subfigure[EMNIST]{\epsfig{figure=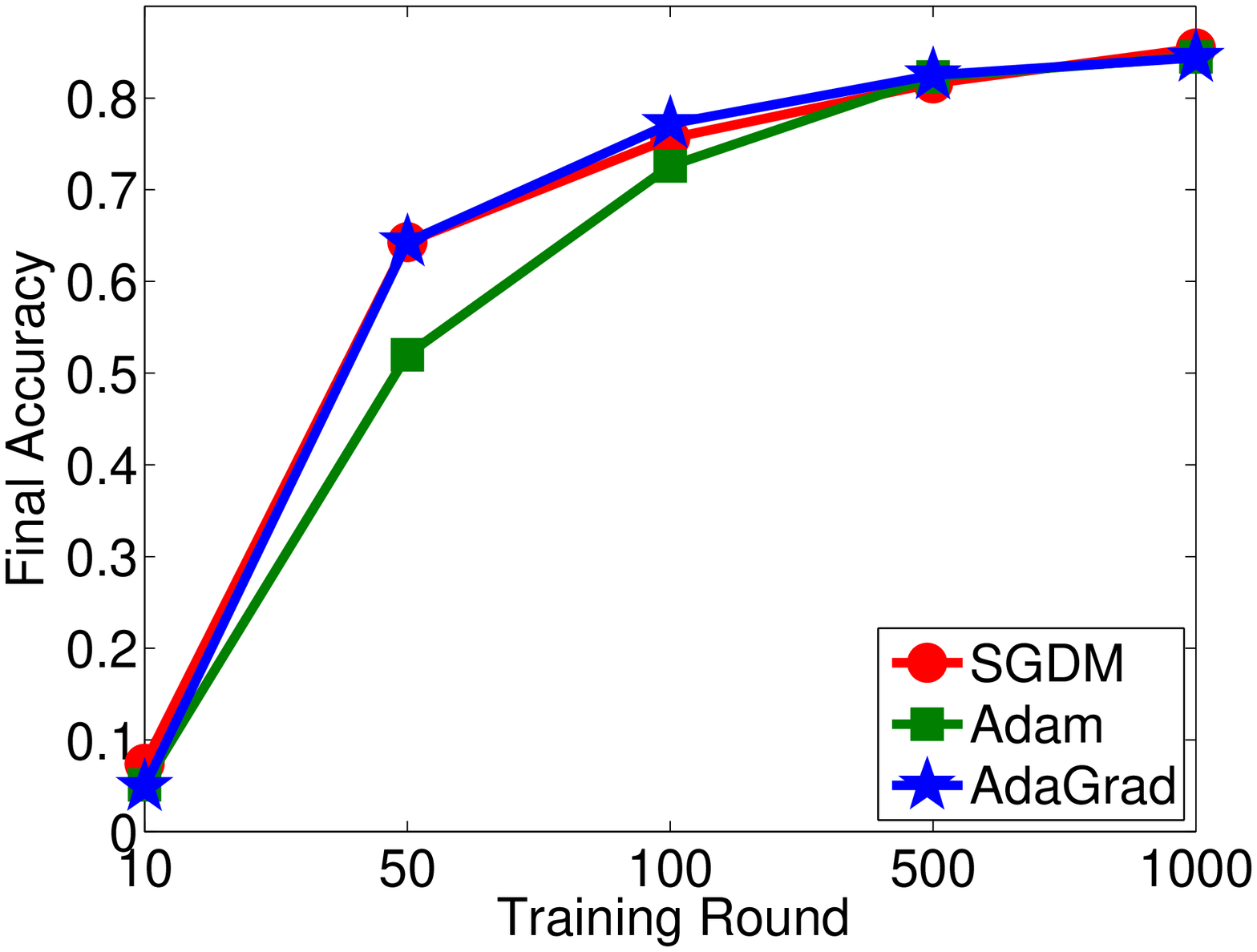, height=1.35in, width=0.34\linewidth}} \hspace{-0.225cm}
\subfigure[Stack Overflow]{\epsfig{figure=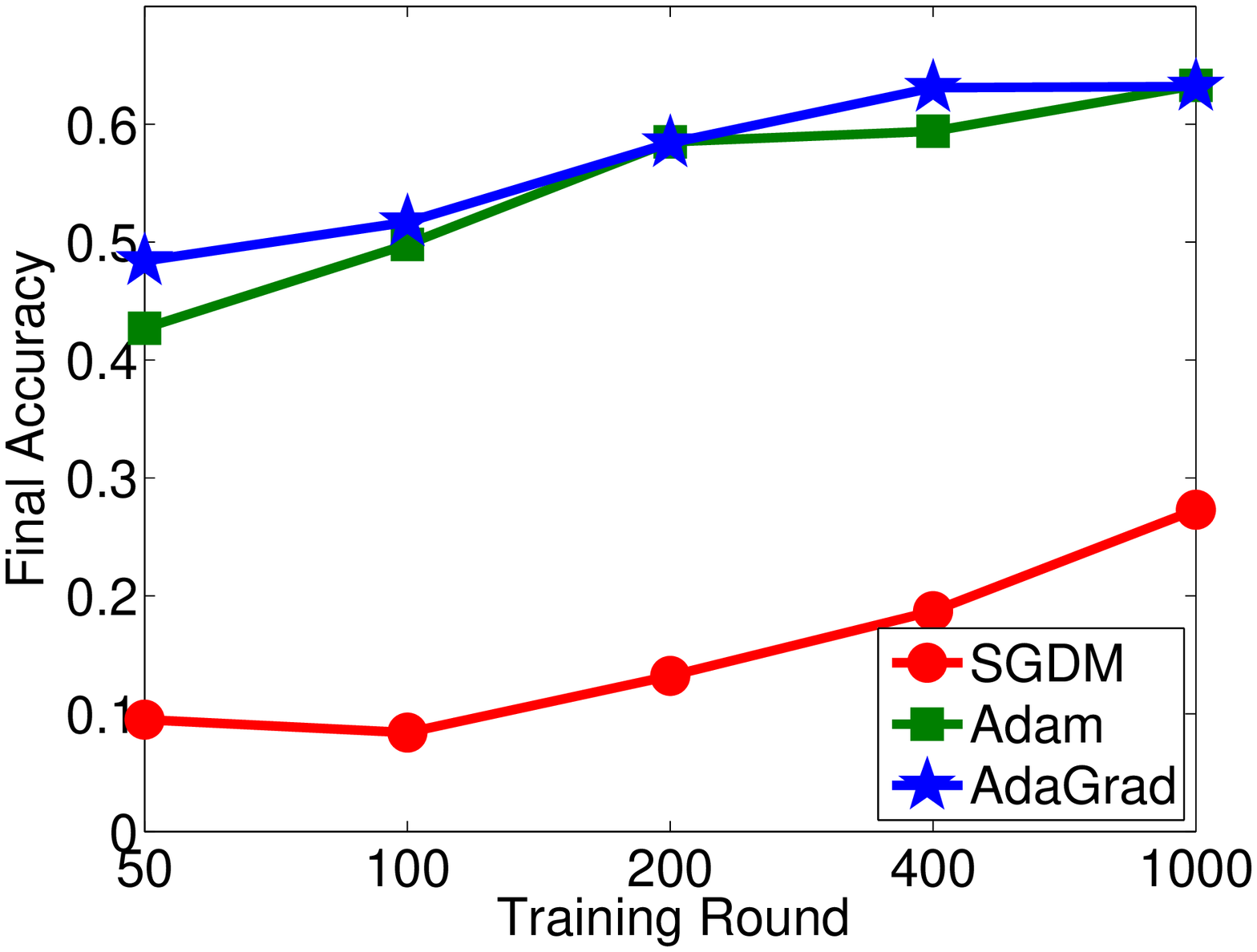, height=1.35in, width=0.34\linewidth}}}
\caption{Final Accuracy with Varying Training Rounds}
\label{fig.Training}
\end{figure}
\end{minipage}
\end{figure*}

{\bf Impact of local iteration numbers.} Figure \ref{fig.Local}  shows the impact of the numbers of local iterations in our \method\ model with the adaptive optimizers over the datasets of CIFAR-100, EMNIST, and Stack Overflow respectively. The performance curves initially raise when the local iteration number increases and then keep relatively stable or even drop if the local iteration number keeps increasing. This demonstrates that there must exist a suitable local iteration number for the FL training. A too large number may make the  clients overfit to their local datasets, such that the local models are far away from globally optimal models and the FL training achieves slower convergence. On the other hand, a too small number may result in slow convergence of local training and also increase the difficulty of convergence of global training. Thus, it is important to choose the appropriate numbers for well balancing the local training and global training. Notice that the final accuracy of AdaGrad and SGDM is closed to 0 on the EMNIST dataset when the local iteration number is larger than 10. A reasonable explanation is EMNIST is a simple dataset and a large local iteration number makes local models converge to their local minimum, which may be distant from the minimum of global model. This leads to lower accuracy of global model.

{\bf Impact of training round numbers.} Figures \ref{fig.Training} (a)-(c) present the performance achieved by our \method\ method with varying the numbers of training rounds from 100 to 2,000, from 10 to 1,000, and from 50 to 1,000 on three datasets. It is obvious that the performance curves with each optimizer keep increasing with the increased number of training rounds. This phenomenon indicates that the accuracy in the federated settings are sensitive to training rounds. This is because the special data and system heterogeneity issues in the FL increase the difficulty in converging in a short time and the FL models need more training rounds to obtain the desired learning results. However, as shown in the above experiments of convergence in Figures \ref{fig.CIFARConvergence}-\ref{fig.EMNISTLoss}, our \method\ method presents superior convergence performance, compared with other FL algorithms, including both regular federated learning and federated optimization approaches.

\vspace{-0.1cm}
\section{Conclusions}\label{sec.conclusions}
\vspace{-0.15cm}

This work presents a theoretical principle on where to and how to design and utilize adaptive optimization methods in federated settings. First, we establish a connection between federated optimization methods and decompositions of ODEs of corresponding centralized optimizers. Second, we develop a momentum decoupling adaptive optimization method to well balance fast convergence and high accuracy in FL. Finally, we utilize full batch gradient to mimic centralized optimization in the end of the training process.
\vspace{-0.2cm}

%
%


\bibliographystyle{icml2022}

\begin{thebibliography}{112}
\providecommand{\natexlab}[1]{#1}
\providecommand{\url}[1]{\texttt{#1}}
\expandafter\ifx\csname urlstyle\endcsname\relax
  \providecommand{\doi}[1]{doi: #1}\else
  \providecommand{\doi}{doi: \begingroup \urlstyle{rm}\Url}\fi

\bibitem[Acar et~al.(2021)Acar, Zhao, Matas, Mattina, Whatmough, and
  Saligrama]{acar2021federated}
Acar, D. A.~E., Zhao, Y., Matas, R., Mattina, M., Whatmough, P., and Saligrama,
  V.
\newblock Federated learning based on dynamic regularization.
\newblock In \emph{Int. Conf. on Learning Representations ({ICLR})}, pp.\
  1--36, 2021.

\bibitem[Afonin \& Karimireddy(2021)Afonin and Karimireddy]{afonin2021towards}
Afonin, A. and Karimireddy, S.~P.
\newblock Towards model agnostic federated learning using knowledge
  distillation.
\newblock \emph{arXiv preprint arXiv:2110.15210}, 2021.

\bibitem[Anonymous(2022{\natexlab{a}})]{anonymous2022acceleration}
Anonymous.
\newblock Acceleration of federated learning with alleviated forgetting in
  local training.
\newblock In \emph{Submitted to Int. Conf. on Learning Representations
  ({ICLR})}, pp.\  1--19, 2022{\natexlab{a}}.
\newblock under review.

\bibitem[Anonymous(2022{\natexlab{b}})]{anonymous2022an}
Anonymous.
\newblock An agnostic approach to federated learning with class imbalance.
\newblock In \emph{Submitted to Int. Conf. on Learning Representations
  ({ICLR})}, pp.\  1--12, 2022{\natexlab{b}}.
\newblock under review.

\bibitem[Anonymous(2022{\natexlab{c}})]{anonymous2022aquila}
Anonymous.
\newblock {AQUILA}: Communication efficient federated learning with adaptive
  quantization of lazily-aggregated gradients.
\newblock In \emph{Submitted to Int. Conf. on Learning Representations
  ({ICLR})}, pp.\  1--23, 2022{\natexlab{c}}.
\newblock under review.

\bibitem[Anonymous(2022{\natexlab{d}})]{anonymous2022hybrid}
Anonymous.
\newblock Hybrid local {SGD} for federated learning with heterogeneous
  communications.
\newblock In \emph{Submitted to Int. Conf. on Learning Representations
  ({ICLR})}, pp.\  1--41, 2022{\natexlab{d}}.
\newblock under review.

\bibitem[Anonymous(2022{\natexlab{e}})]{anonymous2022improving}
Anonymous.
\newblock Improving federated learning face recognition via privacy-agnostic
  clusters.
\newblock In \emph{Submitted to Int. Conf. on Learning Representations
  ({ICLR})}, pp.\  1--21, 2022{\natexlab{e}}.
\newblock under review.

\bibitem[Anonymous(2022{\natexlab{f}})]{anonymous2022privacypreserving}
Anonymous.
\newblock Privacy-preserving task-agnostic vision transformer for image
  processing.
\newblock In \emph{Submitted to Int. Conf. on Learning Representations
  ({ICLR})}, pp.\  1--28, 2022{\natexlab{f}}.
\newblock under review.

\bibitem[Anonymous(2022{\natexlab{g}})]{anonymous2022recycling}
Anonymous.
\newblock Recycling model updates in federated learning: Are gradient subspaces
  low-rank?
\newblock In \emph{Submitted to Int. Conf. on Learning Representations
  ({ICLR})}, pp.\  1--71, 2022{\natexlab{g}}.
\newblock under review.

\bibitem[Anonymous(2022{\natexlab{h}})]{anonymous2022unsupervised}
Anonymous.
\newblock Unsupervised federated learning is possible.
\newblock In \emph{Submitted to Int. Conf. on Learning Representations
  ({ICLR})}, pp.\  1--22, 2022{\natexlab{h}}.
\newblock under review.

\bibitem[Balunovi{\'c} et~al.(2021)Balunovi{\'c}, Dimitrov, Staab, and
  Vechev]{balunovic2021bayesian}
Balunovi{\'c}, M., Dimitrov, D.~I., Staab, R., and Vechev, M.
\newblock Bayesian framework for gradient leakage.
\newblock \emph{arXiv preprint arXiv:2111.04706}, 2021.

\bibitem[Bao et~al.(2015)Bao, Liu, Xiao, Zhou, and Zhang]{BLXZ15}
Bao, X., Liu, L., Xiao, N., Zhou, Y., and Zhang, Q.
\newblock Policy-driven autonomic configuration management for nosql.
\newblock In \emph{Proceedings of the 2015 IEEE International Conference on
  Cloud Computing (CLOUD'15)}, pp.\  245--252, New York, NY, June 27-July 2
  2015.

\bibitem[Caldas et~al.(2018{\natexlab{a}})Caldas, Kone{\v{c}}n{\'y}, McMahan,
  and Talwalkar]{CKMT18}
Caldas, S., Kone{\v{c}}n{\'y}, J., McMahan, H.~B., and Talwalkar, A.
\newblock Expanding the reach of federated learning by reducing client resource
  requirements.
\newblock \emph{CoRR}, abs/1812.07210, 2018{\natexlab{a}}.

\bibitem[Caldas et~al.(2018{\natexlab{b}})Caldas, Wu, Li, Kone{\v{c}}n{\'y},
  McMahan, Smith, and Talwalkar]{CWLK18}
Caldas, S., Wu, P., Li, T., Kone{\v{c}}n{\'y}, J., McMahan, H.~B., Smith, V.,
  and Talwalkar, A.
\newblock {LEAF:} {A} benchmark for federated settings.
\newblock \emph{CoRR}, abs/1812.01097, 2018{\natexlab{b}}.

\bibitem[Chen \& Chao(2021)Chen and Chao]{chen2021bridging}
Chen, H.-Y. and Chao, W.-L.
\newblock On bridging generic and personalized federated learning.
\newblock \emph{arXiv preprint arXiv:2107.00778}, 2021.

\bibitem[Chen et~al.(2020)Chen, Poor, Saad, and Cui]{Chen2020Convergence}
Chen, M., Poor, H.~V., Saad, W., and Cui, S.
\newblock Convergence time minimization of federated learning over wireless
  networks.
\newblock In \emph{IEEE Int. Conf. on Communications ({ICC})}, pp.\  1--6,
  2020.

\bibitem[Dai et~al.(2020)Dai, Low, and Jaillet]{Dai2020Federated}
Dai, Z., Low, B. K.~H., and Jaillet, P.
\newblock Federated bayesian optimization via thompson sampling.
\newblock In \emph{Annual Conf. on Neural Information Processing Systems
  ({NeurIPS})}, pp.\  1--13, 2020.

\bibitem[Duchi et~al.(2011{\natexlab{a}})Duchi, Hazan, and
  Singer]{duchi2011adaptive}
Duchi, J., Hazan, E., and Singer, Y.
\newblock Adaptive subgradient methods for online learning and stochastic
  optimization.
\newblock \emph{Journal of machine learning research}, 12\penalty0 (7),
  2011{\natexlab{a}}.

\bibitem[Duchi et~al.(2011{\natexlab{b}})Duchi, Hazan, and Singer]{DHS11}
Duchi, J.~C., Hazan, E., and Singer, Y.
\newblock Adaptive subgradient methods for online learning and stochastic
  optimization.
\newblock \emph{J. Mach. Learn. Res.}, 12:\penalty0 2121--2159,
  2011{\natexlab{b}}.

\bibitem[Fowl et~al.(2021)Fowl, Geiping, Czaja, Goldblum, and
  Goldstein]{fowl2021robbing}
Fowl, L., Geiping, J., Czaja, W., Goldblum, M., and Goldstein, T.
\newblock Robbing the fed: Directly obtaining private data in federated
  learning with modified models.
\newblock \emph{arXiv preprint arXiv:2110.13057}, 2021.

\bibitem[Gao et~al.(2021)Gao, Xu, and Huang]{gao2021convergence}
Gao, H., Xu, A., and Huang, H.
\newblock On the convergence of communication-efficient local sgd for federated
  learning.
\newblock In \emph{AAAI Conf. on Artificial Intelligence}, volume~35, pp.\
  7510--7518, 2021.

\bibitem[Goswami et~al.(2020)Goswami, Pokhrel, Lee, Liu, Zhang, and
  Zhou]{GPLL20}
Goswami, S., Pokhrel, A., Lee, K., Liu, L., Zhang, Q., and Zhou, Y.
\newblock Graphmap: Scalable iterative graph processing using nosql.
\newblock \emph{The Journal of Supercomputing (TJSC)}, 76\penalty0
  (9):\penalty0 6619--6647, 2020.

\bibitem[Guimu~Guo \& Zhou(2022)Guimu~Guo and Zhou]{GYYK22}
Guimu~Guo, Da~Yan, L. Y. J. K. C. L. Z.~J. and Zhou, Y.
\newblock Maximal directed quasi-clique mining.
\newblock In \emph{Proceedings of the 38th IEEE International Conference on
  Data Engineering (ICDE'22)}, Kuala Lumpur, Malaysia, May 9-12 2022.

\bibitem[Hamer et~al.(2020{\natexlab{a}})Hamer, Mohri, and Suresh]{HMS20}
Hamer, J., Mohri, M., and Suresh, A.~T.
\newblock Fedboost: {A} communication-efficient algorithm for federated
  learning.
\newblock In \emph{Proceedings of the 37th International Conference on Machine
  Learning, {ICML} 2020, 13-18 July 2020, Virtual Event}, pp.\  3973--3983,
  2020{\natexlab{a}}.

\bibitem[Hamer et~al.(2020{\natexlab{b}})Hamer, Mohri, and
  Suresh]{Hamer2020FedBoost}
Hamer, J., Mohri, M., and Suresh, A.~T.
\newblock {F}ed{B}oost: A communication-efficient algorithm for federated
  learning.
\newblock In \emph{Int. Conf. on Machine Learning ({ICML})}, volume 119, pp.\
  3973--3983, 2020{\natexlab{b}}.

\bibitem[He et~al.(2020)He, Annavaram, and Avestimehr]{HAA20}
He, C., Annavaram, M., and Avestimehr, S.
\newblock Group knowledge transfer: Federated learning of large cnns at the
  edge.
\newblock In \emph{Advances in Neural Information Processing Systems 33: Annual
  Conference on Neural Information Processing Systems 2020, NeurIPS 2020,
  December 6-12, 2020, virtual}, 2020.

\bibitem[Hong et~al.(2021)Hong, Wang, Wang, and Zhou]{hong2021federated}
Hong, J., Wang, H., Wang, Z., and Zhou, J.
\newblock Federated robustness propagation: Sharing adversarial robustness in
  federated learning.
\newblock \emph{arXiv preprint arXiv:2106.10196}, 2021.

\bibitem[Hsieh et~al.(2020)Hsieh, Phanishayee, Mutlu, and Gibbons]{HPMG20}
Hsieh, K., Phanishayee, A., Mutlu, O., and Gibbons, P.~B.
\newblock The non-iid data quagmire of decentralized machine learning.
\newblock In \emph{Proceedings of the 37th International Conference on Machine
  Learning, {ICML} 2020, 13-18 July 2020, Virtual Event}, volume 119 of
  \emph{Proceedings of Machine Learning Research}, pp.\  4387--4398. {PMLR},
  2020.

\bibitem[Hyeon-Woo et~al.(2021)Hyeon-Woo, Ye-Bin, and Oh]{hyeon2021fedpara}
Hyeon-Woo, N., Ye-Bin, M., and Oh, T.-H.
\newblock Fedpara: Low-rank hadamard product for communication-efficient
  federated learning.
\newblock \emph{arXiv preprint arXiv:2108.06098}, 2021.

\bibitem[Ingerman \& Ostrowski(2019)Ingerman and Ostrowski]{InOs19}
Ingerman, A. and Ostrowski, K.
\newblock Introducing tensorflow federated.
\newblock \url{https://medium.com/tensorflow/introducing-tensorflow-federated},
  2019.

\bibitem[Jiang et~al.(2016)Jiang, Perng, Sailer, Silva-Lepe, Zhou, and
  Li]{JPSS16}
Jiang, Y., Perng, C.-S., Sailer, A., Silva-Lepe, I., Zhou, Y., and Li, T.
\newblock Csm: A cloud service marketplace for complex service acquisition.
\newblock \emph{ACM Transactions on Intelligent Systems and Technology (TIST)},
  8\penalty0 (1):\penalty0 1--25, 2016.

\bibitem[Kairouz et~al.(2019)Kairouz, McMahan, Avent, Bellet, Bennis, Bhagoji,
  Bonawitz, Charles, Cormode, Cummings, et~al.]{kairouz2019advances}
Kairouz, P., McMahan, H.~B., Avent, B., Bellet, A., Bennis, M., Bhagoji, A.~N.,
  Bonawitz, K., Charles, Z., Cormode, G., Cummings, R., et~al.
\newblock Advances and open problems in federated learning.
\newblock \emph{arXiv preprint arXiv:1912.04977}, 2019.

\bibitem[Kairouz et~al.(2021)Kairouz, McMahan, Avent, Bellet, Bennis, Bhagoji,
  Bonawitz, Charles, Cormode, Cummings, D'Oliveira, Eichner, Rouayheb, Evans,
  Gardner, Garrett, Gasc{\'{o}}n, Ghazi, Gibbons, Gruteser, Harchaoui, He, He,
  Huo, Hutchinson, Hsu, Jaggi, Javidi, Joshi, Khodak, Kone{\v{c}}n{\'y},
  Korolova, Koushanfar, Koyejo, Lepoint, Liu, Mittal, Mohri, Nock,
  {\"{O}}zg{\"{u}}r, Pagh, Qi, Ramage, Raskar, Raykova, Song, Song, Stich, Sun,
  Suresh, Tram{\`{e}}r, Vepakomma, Wang, Xiong, Xu, Yang, Yu, Yu, and
  Zhao]{KMAB21}
Kairouz, P., McMahan, H.~B., Avent, B., Bellet, A., Bennis, M., Bhagoji, A.~N.,
  Bonawitz, K.~A., Charles, Z., Cormode, G., Cummings, R., D'Oliveira, R.
  G.~L., Eichner, H., Rouayheb, S.~E., Evans, D., Gardner, J., Garrett, Z.,
  Gasc{\'{o}}n, A., Ghazi, B., Gibbons, P.~B., Gruteser, M., Harchaoui, Z., He,
  C., He, L., Huo, Z., Hutchinson, B., Hsu, J., Jaggi, M., Javidi, T., Joshi,
  G., Khodak, M., Kone{\v{c}}n{\'y}, J., Korolova, A., Koushanfar, F., Koyejo,
  S., Lepoint, T., Liu, Y., Mittal, P., Mohri, M., Nock, R., {\"{O}}zg{\"{u}}r,
  A., Pagh, R., Qi, H., Ramage, D., Raskar, R., Raykova, M., Song, D., Song,
  W., Stich, S.~U., Sun, Z., Suresh, A.~T., Tram{\`{e}}r, F., Vepakomma, P.,
  Wang, J., Xiong, L., Xu, Z., Yang, Q., Yu, F.~X., Yu, H., and Zhao, S.
\newblock Advances and open problems in federated learning.
\newblock \emph{Found. Trends Mach. Learn.}, 14\penalty0 (1-2):\penalty0
  1--210, 2021.

\bibitem[Karimireddy et~al.(2020{\natexlab{a}})Karimireddy, He, and
  Jaggi]{karimireddy2020byzantine}
Karimireddy, S.~P., He, L., and Jaggi, M.
\newblock Byzantine-robust learning on heterogeneous datasets via bucketing.
\newblock \emph{arXiv preprint arXiv:2006.09365}, 2020{\natexlab{a}}.

\bibitem[Karimireddy et~al.(2020{\natexlab{b}})Karimireddy, Jaggi, Kale, Mohri,
  Reddi, Stich, and Suresh]{karimireddy2020mime}
Karimireddy, S.~P., Jaggi, M., Kale, S., Mohri, M., Reddi, S.~J., Stich, S.~U.,
  and Suresh, A.~T.
\newblock Mime: Mimicking centralized stochastic algorithms in federated
  learning.
\newblock \emph{arXiv preprint arXiv:2008.03606}, 2020{\natexlab{b}}.

\bibitem[Karimireddy et~al.(2020{\natexlab{c}})Karimireddy, Kale, Mohri, Reddi,
  Stich, and Suresh]{karimireddy2020scaffold}
Karimireddy, S.~P., Kale, S., Mohri, M., Reddi, S., Stich, S., and Suresh,
  A.~T.
\newblock Scaffold: Stochastic controlled averaging for federated learning.
\newblock In \emph{Int. Conf. on Machine Learning ({ICML})}, pp.\
  5132--=--5143, 2020{\natexlab{c}}.

\bibitem[Karimireddy et~al.(2020{\natexlab{d}})Karimireddy, Kale, Mohri, Reddi,
  Stich, and Suresh]{KKMR20}
Karimireddy, S.~P., Kale, S., Mohri, M., Reddi, S.~J., Stich, S.~U., and
  Suresh, A.~T.
\newblock {SCAFFOLD:} stochastic controlled averaging for federated learning.
\newblock In \emph{Proceedings of the 37th International Conference on Machine
  Learning, {ICML} 2020, 13-18 July 2020, Virtual Event}, volume 119 of
  \emph{Proceedings of Machine Learning Research}, pp.\  5132--5143. {PMLR},
  2020{\natexlab{d}}.

\bibitem[Karimireddy et~al.(2021)Karimireddy, Jaggi, Kale, Mohri, Reddi, Stich,
  and Suresh]{KJKM21}
Karimireddy, S.~P., Jaggi, M., Kale, S., Mohri, M., Reddi, S.~J., Stich, S.~U.,
  and Suresh, A.~T.
\newblock Mime: Mimicking centralized stochastic algorithms in federated
  learning.
\newblock In \emph{9th International Conference on Learning Representations,
  {ICLR} 2021, Virtual Event, Austria, May 3-7, 2021}, 2021.

\bibitem[Khanduri et~al.(2021)Khanduri, Sharma, Yang, Hong, Liu, Rajawat, and
  Varshney]{khanduri2021stem}
Khanduri, P., Sharma, P., Yang, H., Hong, M., Liu, J., Rajawat, K., and
  Varshney, P.~K.
\newblock Stem: A stochastic two-sided momentum algorithm achieving
  near-optimal sample and communication complexities for federated learning.
\newblock In \emph{Annual Conf. on Neural Information Processing Systems
  ({NeurIPS})}, pp.\  1–12, 2021.

\bibitem[Kingma \& Ba(2014)Kingma and Ba]{kingma2014adam}
Kingma, D.~P. and Ba, J.
\newblock Adam: A method for stochastic optimization.
\newblock \emph{arXiv preprint arXiv:1412.6980}, 2014.

\bibitem[Kingma \& Ba(2015)Kingma and Ba]{KiBa14}
Kingma, D.~P. and Ba, J.
\newblock Adam: {A} method for stochastic optimization.
\newblock In Bengio, Y. and LeCun, Y. (eds.), \emph{3rd International
  Conference on Learning Representations, {ICLR} 2015, San Diego, CA, USA, May
  7-9, 2015, Conference Track Proceedings}, 2015.

\bibitem[Kone{\v{c}}n{\'y} et~al.(2016{\natexlab{a}})Kone{\v{c}}n{\'y},
  McMahan, Ramage, and Richt{\'{a}}rik]{KMRR16}
Kone{\v{c}}n{\'y}, J., McMahan, H.~B., Ramage, D., and Richt{\'{a}}rik, P.
\newblock Federated optimization: Distributed machine learning for on-device
  intelligence.
\newblock \emph{CoRR}, abs/1610.02527, 2016{\natexlab{a}}.

\bibitem[Kone{\v{c}}n{\'y} et~al.(2016{\natexlab{b}})Kone{\v{c}}n{\'y},
  McMahan, Yu, Richt{\'{a}}rik, Suresh, and Bacon]{KMYR16}
Kone{\v{c}}n{\'y}, J., McMahan, H.~B., Yu, F.~X., Richt{\'{a}}rik, P., Suresh,
  A.~T., and Bacon, D.
\newblock Federated learning: Strategies for improving communication
  efficiency.
\newblock In \emph{NIPS Workshop on Private Multi-Party Machine Learning},
  2016{\natexlab{b}}.

\bibitem[Krizhevsky(2009)]{Kriz09}
Krizhevsky, A.
\newblock Learning multiple layers of features from tiny images.
\newblock \emph{Technical Report}, 2009.

\bibitem[Lee et~al.(2013)Lee, Liu, Tang, Zhang, and Zhou]{LLTZ13}
Lee, K., Liu, L., Tang, Y., Zhang, Q., and Zhou, Y.
\newblock Efficient and customizable data partitioning framework for
  distributed big rdf data processing in the cloud.
\newblock In \emph{Proceedings of the 2013 IEEE International Conference on
  Cloud Computing (CLOUD'13)}, pp.\  327--334, Santa Clara, CA, June 27-July 2
  2013.

\bibitem[Lee et~al.(2015)Lee, Liu, Schwan, Pu, Zhang, Zhou, Yigitoglu, and
  Yuan]{LLSP15}
Lee, K., Liu, L., Schwan, K., Pu, C., Zhang, Q., Zhou, Y., Yigitoglu, E., and
  Yuan, P.
\newblock Scaling iterative graph computations with graphmap.
\newblock In \emph{Proceedings of the 27th IEEE international conference for
  High Performance Computing, Networking, Storage and Analysis (SC'15)}, pp.\
  57:1--57:12, Austin, TX, November 15-20 2015.

\bibitem[Lee et~al.(2019)Lee, Liu, Ganti, Srivatsa, Zhang, Zhou, and
  Wang]{LLGS19}
Lee, K., Liu, L., Ganti, R.~L., Srivatsa, M., Zhang, Q., Zhou, Y., and Wang, Q.
\newblock Lightwieight indexing and querying services for big spatial data.
\newblock \emph{IEEE Transactions on Services Computing (TSC)}, 12\penalty0
  (3):\penalty0 343--355, 2019.

\bibitem[Leroy et~al.(2019)Leroy, Coucke, Lavril, Gisselbrecht, and
  Dureau]{David2019Federated}
Leroy, D., Coucke, A., Lavril, T., Gisselbrecht, T., and Dureau, J.
\newblock Federated learning for keyword spotting.
\newblock In \emph{IEEE Int. Conf. on Acoustics, Speech and Signal Processing
  ({ICASSP})}, pp.\  6341--6345, 2019.

\bibitem[Li \& McCallum(2006)Li and McCallum]{LiMc06}
Li, W. and McCallum, A.
\newblock Pachinko allocation: Dag-structured mixture models of topic
  correlations.
\newblock In Cohen, W.~W. and Moore, A.~W. (eds.), \emph{Machine Learning,
  Proceedings of the Twenty-Third International Conference {(ICML} 2006),
  Pittsburgh, Pennsylvania, USA, June 25-29, 2006}, volume 148 of \emph{{ACM}
  International Conference Proceeding Series}, pp.\  577--584. {ACM}, 2006.

\bibitem[Li et~al.(2020)Li, Kovalev, Qian, and Richtarik]{Li2020Acceleration}
Li, Z., Kovalev, D., Qian, X., and Richtarik, P.
\newblock Acceleration for compressed gradient descent in distributed and
  federated optimization.
\newblock In \emph{Int. Conf. on Machine Learning ({ICML})}, volume 119, pp.\
  5895--5904, 2020.

\bibitem[Liu et~al.(2021)Liu, Huang, Zhou, Li, Ji, Xiong, and
  Dou]{liu2021distributed}
Liu, J., Huang, J., Zhou, Y., Li, X., Ji, S., Xiong, H., and Dou, D.
\newblock From distributed machine learning to federated learning: A survey.
\newblock \emph{arXiv preprint arXiv:2104.14362}, 2021.

\bibitem[Liu et~al.(2022)Liu, Huang, Zhou, Li, Ji, Xiong, and Dou]{LHZL22}
Liu, J., Huang, J., Zhou, Y., Li, X., Ji, S., Xiong, H., and Dou, D.
\newblock From distributed machine learning to federated learning: {A} survey.
\newblock \emph{Knowledge and Information Systems (KAIS)}, 64\penalty0
  (4):\penalty0 885--917, 2022.

\bibitem[Liu et~al.(2020{\natexlab{a}})Liu, Chen, Chen, and Zhang]{LCCZ20}
Liu, W., Chen, L., Chen, Y., and Zhang, W.
\newblock Accelerating federated learning via momentum gradient descent.
\newblock \emph{{IEEE} Trans. Parallel Distributed Syst.}, 31\penalty0
  (8):\penalty0 1754--1766, 2020{\natexlab{a}}.

\bibitem[Liu et~al.(2020{\natexlab{b}})Liu, Chen, Chen, and
  Zhang]{Liu2020Accelerating}
Liu, W., Chen, L., Chen, Y., and Zhang, W.
\newblock Accelerating federated learning via momentum gradient descent.
\newblock \emph{IEEE Transactions on Parallel and Distributed Systems
  ({TPDS})}, 31\penalty0 (8):\penalty0 1754--1766, 2020{\natexlab{b}}.

\bibitem[Liu et~al.(2019)Liu, Kang, Zhang, Li, Cheng, Chen, Hong, and
  Yang]{LKZL19}
Liu, Y., Kang, Y., Zhang, X., Li, L., Cheng, Y., Chen, T., Hong, M., and Yang,
  Q.
\newblock A communication efficient vertical federated learning framework.
\newblock \emph{CoRR}, abs/1912.11187, 2019.

\bibitem[McMahan et~al.(2017{\natexlab{a}})McMahan, Moore, Ramage, Hampson, and
  y~Arcas]{MMRH17}
McMahan, B., Moore, E., Ramage, D., Hampson, S., and y~Arcas, B.~A.
\newblock Communication-efficient learning of deep networks from decentralized
  data.
\newblock In Singh, A. and Zhu, X.~J. (eds.), \emph{Proceedings of the 20th
  International Conference on Artificial Intelligence and Statistics, {AISTATS}
  2017, 20-22 April 2017, Fort Lauderdale, FL, {USA}}, volume~54 of
  \emph{Proceedings of Machine Learning Research}, pp.\  1273--1282. {PMLR},
  2017{\natexlab{a}}.

\bibitem[McMahan et~al.(2017{\natexlab{b}})McMahan, Moore, Ramage, Hampson, and
  y~Arcas]{McMahan2017Communication}
McMahan, B., Moore, E., Ramage, D., Hampson, S., and y~Arcas, B.~A.
\newblock Communication-efficient learning of deep networks from decentralized
  data.
\newblock In \emph{Artificial Intelligence and Statistics}, pp.\  1273--1282,
  2017{\natexlab{b}}.

\bibitem[McMahan \& Streeter(2010)McMahan and Streeter]{McSt10}
McMahan, H.~B. and Streeter, M.~J.
\newblock Adaptive bound optimization for online convex optimization.
\newblock In Kalai, A.~T. and Mohri, M. (eds.), \emph{{COLT} 2010 - The 23rd
  Conference on Learning Theory, Haifa, Israel, June 27-29, 2010}, pp.\
  244--256. Omnipress, 2010.

\bibitem[McMahan et~al.(2016)McMahan, Moore, Ramage, and y~Arcas]{MMRA16}
McMahan, H.~B., Moore, E., Ramage, D., and y~Arcas, B.~A.
\newblock Federated learning of deep networks using model averaging.
\newblock \emph{CoRR}, abs/1602.05629, 2016.

\bibitem[Mills et~al.(2019)Mills, Hu, and Min]{mills2019communication}
Mills, J., Hu, J., and Min, G.
\newblock Communication-efficient federated learning for wireless edge
  intelligence in iot.
\newblock \emph{IEEE Internet of Things Journal}, 7\penalty0 (7):\penalty0
  5986--5994, 2019.

\bibitem[Mills et~al.(2021)Mills, Hu, Min, Jin, Zheng, and
  Wang]{mills2021accelerating}
Mills, J., Hu, J., Min, G., Jin, R., Zheng, S., and Wang, J.
\newblock Accelerating federated learning with a global biased optimiser.
\newblock \emph{arXiv preprint arXiv:2108.09134}, 2021.

\bibitem[Mitra et~al.(2021{\natexlab{a}})Mitra, Jaafar, Pappas, and
  Hassani]{MJPH21}
Mitra, A., Jaafar, R., Pappas, G., and Hassani, H.
\newblock Linear convergence in federated learning: Tackling client
  heterogeneity and sparse gradients.
\newblock In \emph{The 35th Conference on Neural Information Processing
  Systems, (NeurIPS'21)}, Online, December 6-14 2021{\natexlab{a}}.

\bibitem[Mitra et~al.(2021{\natexlab{b}})Mitra, Jaafar, Pappas, and
  Hassani]{mitra2021linear}
Mitra, A., Jaafar, R., Pappas, G.~J., and Hassani, H.
\newblock Linear convergence in federated learning: Tackling client
  heterogeneity and sparse gradients.
\newblock In \emph{Annual Conf. on Neural Information Processing Systems
  ({NeurIPS})}, pp.\  1--14, 2021{\natexlab{b}}.

\bibitem[Oh et~al.(2021)Oh, Kim, and Yun]{oh2021fedbabu}
Oh, J., Kim, S., and Yun, S.-Y.
\newblock Fedbabu: Towards enhanced representation for federated image
  classification.
\newblock \emph{arXiv preprint arXiv:2106.06042}, 2021.

\bibitem[Ozfatura et~al.(2021)Ozfatura, Ozfatura, and
  G{\"u}nd{\"u}z]{ozfatura2021fedadc}
Ozfatura, E., Ozfatura, K., and G{\"u}nd{\"u}z, D.
\newblock Fedadc: Accelerated federated learning with drift control.
\newblock In \emph{IEEE Int. Symposium on Information Theory ({ISIT})}, pp.\
  467--472, 2021.

\bibitem[Palanisamy et~al.(2014)Palanisamy, Liu, Lee, Meng, Tang, and
  Zhou]{PLLM14}
Palanisamy, B., Liu, L., Lee, K., Meng, S., Tang, Y., and Zhou, Y.
\newblock Anonymizing continuous queries with delay-tolerant mix-zones over
  road networks.
\newblock \emph{Distributed and Parallel Databases (DAPD)}, 32\penalty0
  (1):\penalty0 91--118, 2014.

\bibitem[Palanisamy et~al.(2018)Palanisamy, Liu, Zhou, and Wang]{PLZW18}
Palanisamy, B., Liu, L., Zhou, Y., and Wang, Q.
\newblock Privacy-preserving publishing of multilevel utility-controlled graph
  datasets.
\newblock \emph{ACM Transactions on Internet Technology (TOIT)}, 18\penalty0
  (2):\penalty0 24:1--24:21, 2018.

\bibitem[Pathak \& Wainwright(2020)Pathak and Wainwright]{Pathak2020FedSplit}
Pathak, R. and Wainwright, M.~J.
\newblock Fedsplit: an algorithmic framework for fast federated optimization.
\newblock In \emph{Annual Conf. on Neural Information Processing Systems
  ({NeurIPS})}, volume~33, pp.\  7057--7066, 2020.

\bibitem[Qian(1999)]{Qian99}
Qian, N.
\newblock On the momentum term in gradient descent learning algorithms.
\newblock \emph{Neural Networks}, 12\penalty0 (1):\penalty0 145--151, 1999.

\bibitem[Reddi et~al.(2018)Reddi, Zaheer, Sachan, Kale, and
  Kumar]{reddi2018adaptive}
Reddi, S., Zaheer, M., Sachan, D., Kale, S., and Kumar, S.
\newblock Adaptive methods for nonconvex optimization.
\newblock In \emph{Annual Conf. on Neural Information Processing Systems
  ({NeurIPS})}, pp.\  1--17, 2018.

\bibitem[Reddi et~al.(2021{\natexlab{a}})Reddi, Charles, Zaheer, Garrett, Rush,
  Kone{\v{c}}n{\'y}, Kumar, and McMahan]{RCZG21}
Reddi, S.~J., Charles, Z., Zaheer, M., Garrett, Z., Rush, K.,
  Kone{\v{c}}n{\'y}, J., Kumar, S., and McMahan, H.~B.
\newblock Adaptive federated optimization.
\newblock In \emph{9th International Conference on Learning Representations,
  {ICLR} 2021, Virtual Event, Austria, May 3-7, 2021}, 2021{\natexlab{a}}.

\bibitem[Reddi et~al.(2021{\natexlab{b}})Reddi, Charles, Zaheer, Garrett, Rush,
  Kone{\v{c}}n{\'y}, Kumar, and McMahan]{reddi2021adaptive}
Reddi, S.~J., Charles, Z., Zaheer, M., Garrett, Z., Rush, K.,
  Kone{\v{c}}n{\'y}, J., Kumar, S., and McMahan, H.~B.
\newblock Adaptive federated optimization.
\newblock In \emph{Int. Conf. on Learning Representations ({ICLR})},
  2021{\natexlab{b}}.

\bibitem[Rothchild et~al.(2020)Rothchild, Panda, Ullah, Ivkin, Stoica,
  Braverman, Gonzalez, and Arora]{rothchild2020fetchsgd}
Rothchild, D., Panda, A., Ullah, E., Ivkin, N., Stoica, I., Braverman, V.,
  Gonzalez, J., and Arora, R.
\newblock Fetchsgd: Communication-efficient federated learning with sketching.
\newblock In \emph{Int. Conf. on Machine Learning ({ICML})}, pp.\  8253--8265,
  2020.

\bibitem[Rumelhart et~al.(1986)Rumelhart, Hinton, and Williams]{RHW86}
Rumelhart, D.~E., Hinton, G.~E., and Williams, R.~J.
\newblock Learning internal representations by error propagation.
\newblock In Rumelhart, D.~E. and McClelland, J.~L. (eds.), \emph{Parallel
  Distributed Processing}. MIT Press, 1986.

\bibitem[Sharma et~al.(2021)Sharma, Chen, Zhao, Qiu, Chaterji, and
  Bagchi]{sharma2021tesseract}
Sharma, A., Chen, W., Zhao, J., Qiu, Q., Chaterji, S., and Bagchi, S.
\newblock Tesseract: Gradient flip score to secure federated learning against
  model poisoning attacks.
\newblock \emph{arXiv preprint arXiv:2110.10108}, 2021.

\bibitem[Su et~al.(2013)Su, Liu, Li, Fan, and Zhou]{SLLF13}
Su, Z., Liu, L., Li, M., Fan, X., and Zhou, Y.
\newblock Servicetrust: Trust management in service provision networks.
\newblock In \emph{Proceedings of the 10th IEEE International Conference on
  Services Computing (SCC'13)}, pp.\  272--279, Santa Clara, CA, June 27-July 2
  2013.

\bibitem[Su et~al.(2015)Su, Liu, Li, Fan, and Zhou]{SLLF15}
Su, Z., Liu, L., Li, M., Fan, X., and Zhou, Y.
\newblock Reliable and resilient trust management in distributed service
  provision networks.
\newblock \emph{ACM Transactions on the Web (TWEB)}, 9\penalty0 (3):\penalty0
  1--37, 2015.

\bibitem[Sutskever et~al.(2013)Sutskever, Martens, Dahl, and Hinton]{SMDH13}
Sutskever, I., Martens, J., Dahl, G.~E., and Hinton, G.~E.
\newblock On the importance of initialization and momentum in deep learning.
\newblock In \emph{Proceedings of the 30th International Conference on Machine
  Learning, {ICML} 2013, Atlanta, GA, USA, 16-21 June 2013}, volume~28 of
  \emph{{JMLR} Workshop and Conference Proceedings}, pp.\  1139--1147, 2013.

\bibitem[TensorFlow(2019)]{SO19}
TensorFlow.
\newblock Tensorflow federated stack overflow dataset.
\newblock
  \url{https://www.tensorflow.org/federated/api\_docs/python/tff/simulation/datasets/stackoverflow/load\_data},
  2019.

\bibitem[Triastcyn et~al.(2021)Triastcyn, Reisser, and
  Louizos]{triastcyn2021dp}
Triastcyn, A., Reisser, M., and Louizos, C.
\newblock Dp-rec: Private \& communication-efficient federated learning.
\newblock \emph{arXiv preprint arXiv:2111.05454}, 2021.

\bibitem[Wang et~al.(2021{\natexlab{a}})Wang, Stich, He, and
  Fritz]{wang2021progfed}
Wang, H.-P., Stich, S.~U., He, Y., and Fritz, M.
\newblock Progfed: Effective, communication, and computation efficient
  federated learning by progressive training.
\newblock \emph{arXiv preprint arXiv:2110.05323}, 2021{\natexlab{a}}.

\bibitem[Wang et~al.(2020)Wang, Tantia, Ballas, and Rabbat]{Wang2020SlowMo}
Wang, J., Tantia, V., Ballas, N., and Rabbat, M.
\newblock Slowmo: Improving communication-efficient distributed sgd with slow
  momentum.
\newblock In \emph{Int. Conf. on Learning Representations ({ICLR})}, pp.\
  1--27, 2020.

\bibitem[Wang et~al.(2021{\natexlab{b}})Wang, Xu, Garrett, Charles, Liu, and
  Joshi]{WXGC21}
Wang, J., Xu, Z., Garrett, Z., Charles, Z., Liu, L., and Joshi, G.
\newblock Local adaptivity in federated learning: Convergence and consistency.
\newblock In \emph{The International Workshop on Federated Learning for User
  Privacy and Data Confidentiality in Conjunction with ICML 2021,
  (FL-ICML'21)}, Online, December 6-14 2021{\natexlab{b}}.

\bibitem[Wang et~al.(2021{\natexlab{c}})Wang, Xu, Garrett, Charles, Liu, and
  Joshi]{wang2021local}
Wang, J., Xu, Z., Garrett, Z., Charles, Z., Liu, L., and Joshi, G.
\newblock Local adaptivity in federated learning: Convergence and consistency.
\newblock \emph{arXiv preprint arXiv:2106.02305}, 2021{\natexlab{c}}.

\bibitem[Woodworth et~al.(2020)Woodworth, Patel, and Srebro]{WPS20}
Woodworth, B.~E., Patel, K.~K., and Srebro, N.
\newblock Minibatch vs local {SGD} for heterogeneous distributed learning.
\newblock In Larochelle, H., Ranzato, M., Hadsell, R., Balcan, M., and Lin, H.
  (eds.), \emph{Advances in Neural Information Processing Systems 33: Annual
  Conference on Neural Information Processing Systems 2020, NeurIPS 2020,
  December 6-12, 2020, virtual}, 2020.

\bibitem[Wu \& Wang(2021)Wu and Wang]{wu2021fast}
Wu, H. and Wang, P.
\newblock Fast-convergent federated learning with adaptive weighting.
\newblock \emph{IEEE Transactions on Cognitive Communications and Networking},
  2021.

\bibitem[Wu et~al.(2020)Wu, He, Lin, and Mao]{wu2020accelerating}
Wu, W., He, L., Lin, W., and Mao, R.
\newblock Accelerating federated learning over reliability-agnostic clients in
  mobile edge computing systems.
\newblock \emph{IEEE Transactions on Parallel and Distributed Systems
  ({TPDS})}, 32\penalty0 (7):\penalty0 1539--1551, 2020.

\bibitem[Wu \& He(2020)Wu and He]{WuHe20}
Wu, Y. and He, K.
\newblock Group normalization.
\newblock \emph{Int. J. Comput. Vis.}, 128\penalty0 (3):\penalty0 742--755,
  2020.

\bibitem[Xia et~al.(2021)Xia, Zhu, Yang, Zhou, Shi, and Chen]{xia2021fast}
Xia, S., Zhu, J., Yang, Y., Zhou, Y., Shi, Y., and Chen, W.
\newblock Fast convergence algorithm for analog federated learning.
\newblock In \emph{IEEE Int. Conf. on Communications ({ICC})}, pp.\  1--6,
  2021.

\bibitem[Xie et~al.(2019)Xie, Koyejo, Gupta, and Lin]{XKGL19}
Xie, C., Koyejo, O., Gupta, I., and Lin, H.
\newblock Local adaalter: Communication-efficient stochastic gradient descent
  with adaptive learning rates.
\newblock \emph{CoRR}, abs/1911.09030, 2019.

\bibitem[Yan et~al.(2022{\natexlab{a}})Yan, Qu, Guo, Wang, and Zhou]{YQGW22}
Yan, D., Qu, W., Guo, G., Wang, X., and Zhou, Y.
\newblock Prefixfpm: A parallel framework for general-purpose mining of
  frequent and closed patterns.
\newblock \emph{The VLDB Journal (VLDBJ)}, 31\penalty0 (2):\penalty0 253--286,
  2022{\natexlab{a}}.

\bibitem[Yan et~al.(2022{\natexlab{b}})Yan, Zhou, Guo, and Liu]{YZGL22}
Yan, D., Zhou, Y., Guo, G., and Liu, H.
\newblock Parallel graph processing.
\newblock In Sherif~Sakr, A. Y.~Z. and Taheri, J. (eds.), \emph{Encyclopedia of
  Big Data Technologies}. Springer, 2022{\natexlab{b}}.

\bibitem[Yang(2021)]{Yang2021hfl}
Yang, H.
\newblock H-fl: A hierarchical communication-efficient and privacy-protected
  architecture for federated learning.
\newblock In \emph{Int. Joint Conf. on Artificial Intelligence ({IJCAI})}, pp.\
   479--485, 2021.

\bibitem[Yapp et~al.(2021)Yapp, Koh, Lai, Kang, Li, Ng, Jiang, Lim, Xiong, and
  Niyato1]{Yapp2021Communication}
Yapp, A. Z.~H., Koh, H. S.~N., Lai, Y.~T., Kang, J., Li, X., Ng, J.~S., Jiang,
  H., Lim, W. Y.~B., Xiong, Z., and Niyato1, D.
\newblock Communication-efficient and scalable decentralized federated edge
  learning.
\newblock In \emph{Int. Joint Conf. on Artificial Intelligence ({IJCAI})}, pp.\
   5032--5035, 2021.

\bibitem[Yuan et~al.(2021{\natexlab{a}})Yuan, Morningstar, Ning, and
  Singhal]{yuan2021we}
Yuan, H., Morningstar, W., Ning, L., and Singhal, K.
\newblock What do we mean by generalization in federated learning?
\newblock \emph{arXiv preprint arXiv:2110.14216}, 2021{\natexlab{a}}.

\bibitem[Yuan et~al.(2021{\natexlab{b}})Yuan, Zaheer, and
  Reddi]{Yuan2021Federated}
Yuan, H., Zaheer, M., and Reddi, S.
\newblock Federated composite optimization.
\newblock In \emph{Int. Conf. on Machine Learning ({ICML})}, volume 139, pp.\
  12253--12266, 2021{\natexlab{b}}.

\bibitem[Yun et~al.(2021)Yun, Rajput, and Sra]{yun2021minibatch}
Yun, C., Rajput, S., and Sra, S.
\newblock Minibatch vs local sgd with shuffling: Tight convergence bounds and
  beyond.
\newblock \emph{arXiv preprint arXiv:2110.10342}, 2021.

\bibitem[Zhang et~al.(2022)Zhang, Liu, Jia, Zhou, Dai, and Dou]{ZLJZ22b}
Zhang, H., Liu, J., Jia, J., Zhou, Y., Dai, H., and Dou, D.
\newblock Fedduap: Federated learning with dynamic update and adaptive pruning
  using shared data on the server.
\newblock In \emph{Proceedings of the 31st International Joint Conference on
  Artificial Intelligence (IJCAI'22)}, Messe Wien, Vienna, Austria, July 23-29
  2022.

\bibitem[Zhang et~al.(2019)Zhang, Karimireddy, Veit, Kim, Reddi, Kumar, and
  Sra]{ZKVK19}
Zhang, J., Karimireddy, S.~P., Veit, A., Kim, S., Reddi, S.~J., Kumar, S., and
  Sra, S.
\newblock Why {ADAM} beats {SGD} for attention models.
\newblock \emph{CoRR}, abs/1912.03194, 2019.

\bibitem[Zhang et~al.(2021)Zhang, Sapra, Fidler, Yeung, and
  Alvarez]{zhang2021personalized}
Zhang, M., Sapra, K., Fidler, S., Yeung, S., and Alvarez, J.~M.
\newblock Personalized federated learning with first order model optimization.
\newblock In \emph{Int. Conf. on Learning Representations ({ICLR})}, 2021.

\bibitem[Zhang et~al.(2013)Zhang, Liu, Ren, Lee, Tang, Zhao, and Zhou]{ZLRL13}
Zhang, Q., Liu, L., Ren, Y., Lee, K., Tang, Y., Zhao, X., and Zhou, Y.
\newblock Residency aware inter-vm communication in virtualized cloud:
  Performance measurement and analysis.
\newblock In \emph{Proceedings of the 2013 IEEE International Conference on
  Cloud Computing (CLOUD'13)}, pp.\  204--211, Santa Clara, CA, June 27-July 2
  2013.

\bibitem[Zhang et~al.(2014)Zhang, Liu, Lee, Zhou, Singh, Mandagere, Gopisetty,
  and Alatorre]{ZLLZ14}
Zhang, Q., Liu, L., Lee, K., Zhou, Y., Singh, A., Mandagere, N., Gopisetty, S.,
  and Alatorre, G.
\newblock Improving hadoop service provisioning in a geographically distributed
  cloud.
\newblock In \emph{Proceedings of the 2014 IEEE International Conference on
  Cloud Computing (CLOUD'14)}, pp.\  432--439, Anchorage, AK, June 27-July 2
  2014.

\bibitem[Zhang et~al.()Zhang, Jin, Zhang, Zhou, Zhao, Ren, Liu, Wu, Jin, and
  Dou]{ZJZZ21}
Zhang, Z., Jin, J., Zhang, Z., Zhou, Y., Zhao, X., Ren, J., Liu, J., Wu, L.,
  Jin, R., and Dou, D.
\newblock Validating the lottery ticket hypothesis with inertial manifold
  theory.
\newblock In \emph{Advances in Neural Information Processing Systems 34: Annual
  Conference on Neural Information Processing Systems 2021 (NeurIPS'21)},
  Virtual.

\bibitem[Zhou et~al.(2022{\natexlab{a}})Zhou, Liu, Jia, Zhou, Zhou, Dai, and
  Dou]{ZLJZ22a}
Zhou, C., Liu, J., Jia, J., Zhou, J., Zhou, Y., Dai, H., and Dou, D.
\newblock Efficient device scheduling with multi-job federated learning.
\newblock In \emph{Proceedings of the 36th AAAI Conference on Artificial
  Intelligence (AAAI'22)}, Vancouver, Canada, February 22-March 1
  2022{\natexlab{a}}.

\bibitem[Zhou et~al.(2022{\natexlab{b}})Zhou, Liu, Jia, Zhou, Zhou, Dai, and
  Dou]{zhou2021efficient}
Zhou, C., Liu, J., Jia, J., Zhou, J., Zhou, Y., Dai, H., and Dou, D.
\newblock Efficient device scheduling with multi-job federated learning.
\newblock In \emph{AAAI Conf. on Artificial Intelligence}, pp.\  1--14,
  2022{\natexlab{b}}.

\bibitem[Zhou(2017)]{Zhou17}
Zhou, Y.
\newblock \emph{Innovative Mining, Processing, and Application of Big Graphs}.
\newblock PhD thesis, Georgia Institute of Technology, Atlanta, GA, {USA},
  2017.

\bibitem[Zhou \& Liu(2019)Zhou and Liu]{ZhLi19}
Zhou, Y. and Liu, L.
\newblock Approximate deep network embedding for mining large-scale graphs.
\newblock In \emph{Proceedings of the 2019 IEEE International Conference on
  Cognitive Machine Intelligence (CogMI'19)}, pp.\  53--60, Los Angeles, CA,
  December 12-14 2019.

\bibitem[Zhou et~al.(2015{\natexlab{a}})Zhou, Liu, Lee, Pu, and Zhang]{ZLLP15}
Zhou, Y., Liu, L., Lee, K., Pu, C., and Zhang, Q.
\newblock Fast iterative graph computation with resource aware graph parallel
  abstractions.
\newblock In \emph{Proceedings of the 24th ACM Symposium on High-Performance
  Parallel and Distributed Computing (HPDC'15)}, pp.\  179--190, Portland, OR,
  June 15-19 2015{\natexlab{a}}.

\bibitem[Zhou et~al.(2015{\natexlab{b}})Zhou, Liu, Lee, and Zhang]{ZLLZ15}
Zhou, Y., Liu, L., Lee, K., and Zhang, Q.
\newblock Graphtwist: Fast iterative graph computation with two-tier
  optimizations.
\newblock \emph{Proceedings of the VLDB Endowment (PVLDB)}, 8\penalty0
  (11):\penalty0 1262--1273, 2015{\natexlab{b}}.

\bibitem[Zhou et~al.(2018{\natexlab{a}})Zhou, Amimeur, Jiang, Dou, Jin, and
  Wang]{ZAJD18}
Zhou, Y., Amimeur, A., Jiang, C., Dou, D., Jin, R., and Wang, P.
\newblock Density-aware local siamese autoencoder network embedding with
  autoencoder graph clustering.
\newblock In \emph{Proceedings of the 2018 IEEE International Conference on Big
  Data (BigData'18)}, pp.\  1162--1167, Seattle, WA, December 10-13
  2018{\natexlab{a}}.

\bibitem[Zhou et~al.(2018{\natexlab{b}})Zhou, Wu, Jiang, Zhang, Dou, Jin, and
  Wang]{ZWJZ18}
Zhou, Y., Wu, S., Jiang, C., Zhang, Z., Dou, D., Jin, R., and Wang, P.
\newblock Density-adaptive local edge representation learning with generative
  adversarial network multi-label edge classification.
\newblock In \emph{Proceedings of the 18th IEEE International Conference on
  Data Mining (ICDM'18)}, pp.\  1464--1469, Singapore, November 17-20
  2018{\natexlab{b}}.

\bibitem[Zhou et~al.(2022{\natexlab{c}})Zhou, Ye, and
  Lv]{Zhou2022Communication}
Zhou, Y., Ye, Q., and Lv, J.
\newblock Communication-efficient federated learning with compensated
  overlap-fedavg.
\newblock \emph{IEEE Transactions on Parallel and Distributed Systems
  ({TPDS})}, 33\penalty0 (1):\penalty0 192--205, 2022{\natexlab{c}}.

\end{thebibliography}

\newpage
\appendix
\onecolumn

\section{Appendix}\label{sec.appendix}
\subsection{Related Work}\label{sec.related}
Parallel, distributed, and federated learning have attracted active research in the last decade~\cite{ZLRL13,LLTZ13,SLLF13,PLLM14,ZLLZ14,SLLF15,ZLLP15,BLXZ15,ZLLZ15,LLSP15,JPSS16,Zhou17,ZWJZ18,ZAJD18,PLZW18,LLGS19,ZhLi19,GPLL20,ZJZZ21,ZLJZ22a,YZGL22,YQGW22,LHZL22,GYYK22,ZLJZ22b}.
While it achieves much advancement in federated learning, FedAvg \cite{McMahan2017Communication} does not consider the adaptive optimization when aggregating the weights or gradients from devices, which makes the training process inefficient and makes it difficult to tune or to achieve desired accuracy. SCAFFOLD \cite{karimireddy2020scaffold} exploits control variate to reduce the client drift problem, i.e., ``over-fitting'' to local device data, in order to reduce the influence of the heterogeneity of non-IID data and to improve the performance (accuracy). The control parameters may result in stateful devices \cite{karimireddy2020scaffold,acar2021federated}, which is not compatible with cross-device Federated Learning (FL) because of full participation of devices. The cross-device FL simultaneously deals with large amounts of devices while each round only samples a fraction of the available devices for the training process in order to mitigate the straggler problem \cite{kairouz2019advances,liu2021distributed}. Although the contribution of the devices can be exploited to dynamically update the global model \cite{wu2021fast}, the contribution-based dynamic update cannot well address the client-drift problem. The contribution can calculated based on the angle between the local gradient and the global gradient. 

Adaptive optimization can be applied either at the server side to improve the performance \cite{reddi2021adaptive} and to reduce the communication and computation costs \cite{mills2021accelerating}, using multiple adaptive optimization method, e.g., AdaGrad \cite{duchi2011adaptive}, Yogi \cite{reddi2018adaptive}, Adam \cite{kingma2014adam,mills2019communication} and momentum \cite{rothchild2020fetchsgd,mills2021accelerating}, or at the client side \cite{Yuan2021Federated} to reduce the number of rounds \cite{mills2019communication} utilizing momentum \cite{Liu2020Accelerating,gao2021convergence,Wang2020SlowMo}, Adam \cite{mills2019communication}. In contrast, the single side application of existing adaptive optimization, i.e., the server side or the device side, may lead to insufficient performance, due to incomplete adaptation of the adaptive optimization methods. While correction techniques can be utilized to mitigate the bias of the convergence point for the application of AdaGrad at the device side \cite{wang2021local}, the application at both the server side and the device should be addressed at the same time to achieve better performance. In addition, the application of momentum at the device side \cite{Liu2020Accelerating} may incur severe client-drift because of heterogeneity of non-IID data and law participation rate of devices. Moreover, the synchronization of the global momentum parameters may incur extra communication costs \cite{Wang2020SlowMo}. Momentum \cite{karimireddy2020mime,ozfatura2021fedadc,khanduri2021stem} and Adam \cite{David2019Federated} can be utilized on both the server and the devices to achieve fast training speed and high convergence accuracy, while the simple application may correspond to limited improvement.

The adaptive optimization methods can be combined with compression mechanism \cite{mills2019communication,gao2021convergence,rothchild2020fetchsgd,Li2020Acceleration} to further reduce the communication costs. In addition, device selection methods are utilized to achieve faster convergence \cite{xia2021fast} and higher accuracy \cite{Chen2020Convergence}, even with multiple jobs \cite{zhou2021efficient}. Furthermore, overlapping the local training process and the data transfer can improve the communication efficiency \cite{Zhou2022Communication}. Sparsification \cite{mitra2021linear} and quantization \cite{anonymous2022aquila} can be exploited in FL to improve the communication efficiency, wherein the update at the device side relies on the gradients of the last round of all the devices and more communication is required to configure the sparsification parameters \cite{mitra2021linear}. Moreover, the combination of gossip protocol-based decentralized model aggregation, which is performed at devices, and the centralized model aggregation, which is performed at the server, helps to speed up the training process of FL \cite{anonymous2022hybrid}. Leading principal components \cite{anonymous2022recycling}, low-rank hadamard product \cite{hyeon2021fedpara}, and progressive training \cite{wang2021progfed}, i.e., the model progressively grows along with the training process, can be exploited as well to reduce the communication costs and improve the performance. Some other methods, e.g., communication-efficient ensemble algorithms \cite{Hamer2020FedBoost}, hierarchical architecture \cite{Yang2021hfl,wu2020accelerating}, blockchain-based mechanism \cite{Yapp2021Communication}, federated optimization \cite{Pathak2020FedSplit,Dai2020Federated}, dual averaging procedure for non-smooth regularizer \cite{Yuan2021Federated}, agnostic constrained learning formulation for class imbalance problems \cite{anonymous2022an}, separation of unseen client data and unseen client distributions \cite{yuan2021we}, knowledge distillation for heterogeneous model structures \cite{afonin2021towards}, minibatch and local random reshuffling \cite{yun2021minibatch}, unlabeled data transformation for unsupervised LF \cite{anonymous2022unsupervised}, multi-task FL for image processing \cite{anonymous2022privacypreserving}, and personalization \cite{zhang2021personalized,oh2021fedbabu,hyeon2021fedpara} or the trade-off between personalization and generalization \cite{chen2021bridging}, are proposed to further improve the performance of FL. However, the above works are orthogonal to our approach and out of the scope of this paper. 

As private data may be recovered with small malicious modifications of the shared model \cite{fowl2021robbing} or sensitive information may still be leaked from the gradients \cite{balunovic2021bayesian}, differential privacy \cite{anonymous2022improving,triastcyn2021dp} and compression techniques \cite{triastcyn2021dp} are combined to improve the privacy and communication efficiency of FL, while the encoding of gradient knowledge with the regularization of locally trained parameters \cite{anonymous2022acceleration} helps improve the performance and robustness at the same time. While gradient flips are indicative of attacks, reputation-based weighted aggregation can help to improve the robustness of FL \cite{sharma2021tesseract}. Batch-normalization \cite{hong2021federated} and Bayes optimal method \cite{balunovic2021bayesian} can be exploited to deal with adversarial training attack, while bucketing methods can alleviate the impact of byzantine attacks \cite{karimireddy2020byzantine}. These methods can be combined with our approach to improve the robustness and the privacy of FL.

This work presents a theoretical principle on where to and how to design and utilize adaptive optimization methods in federated settings. We theoretically analyze the connection between federated optimization methods and decompositions of ODEs of corresponding centralized optimizers. We develop a momentum decoupling adaptive optimization method to make full use of the strength of fast convergence and high accuracy of the global momentum. We also utilize the full batch gradients to mimic centralized optimization for ensuring the convergence and overcoming the possible inconsistency caused by adaptive optimization methods.

\subsection{Momentum Decoupling Adaptive Optimization: FedDA+Adam and FedDA+AdaGrad}\label{sec.FedDA+Adam+AdaGrad}
Recall the centralized Adam optimizer.
\begin{equation}\label{ADAM}
\begin{split}
&m(t+1) = \beta_1*m(t) + (1-\beta_1)*g(W(t)),\\
& v(t+1) = \beta_2*v(t) + (1-\beta_2)*(g(W(t))^2,\\
& \hat m(t+1) = m(t+1)/(1-\beta_1^t),\\
& \hat v(t+1) = v(t+1)/(1-\beta_2^t),\\
& W(t+1) = W(t) -\hat m(t+1)/(\sqrt{\hat v(t+1)}+\epsilon)*\eta.
\end{split}
\end{equation}
where $g(W) = \sum_{i=1}^M \frac{N^i}{N} g^i(W)$.
The Adam optimizer is the numerical solution to the ODE system
\begin{equation}\label{ADAM-ODE}
\begin{split}
&\eta\frac{d}{d\tau}m(\tau) = -(1-\beta_1)m(\tau) + (1-\beta_1)*g(W(\tau)),\\
& \eta \frac{d}{d\tau} v(\tau) = -(1-\beta_2)v(\tau) + (1-\beta_2)*(g(W(\tau))^2,\\
& \hat m(\tau) = m(\tau)/(1-\beta_1^\tau),\\
& \hat v(\tau) = v(\tau)/(1-\beta_2^\tau),\\
& \frac{d}{d\tau}W(\tau) =  -\hat m(\tau)/(\sqrt{\hat v(\tau)}+\epsilon)*\eta.
\end{split}
\end{equation}
The equilibria of the above is system are $(m,v,W) = (0,0,W^\ast)$, where $W^\ast$ is local minimum point of the loss function, i.e., $g(W^\ast) = 0$. Therefore, the Adam optimizer trains the model to converge to local minimum point of the loss function. Applying our decoupling method, we first decouple the global momentum and velocity $m$, $v$ with local training, i.e.,
 \begin{equation}
\begin{split}
&\frac{d}{d\tau}W^i(\tau) = -g^i(W^i(\tau)),\\
&\eta\frac{d}{d\tau}m(\tau) = -(1-\beta_1)m(\tau) + (1-\beta_1)*g(W(\tau)),\\
&\eta \frac{d}{d\tau} v(\tau) = -(1-\beta_2)v(\tau) + (1-\beta_2)*(g(W(\tau))^2.
\end{split}
\end{equation}
Similar to the SGDM case, the equation of $m(\tau)$ is totally linear and can be decomposed precisely as 
\begin{equation}\eta\frac{d}{d\tau}m^i(\tau) = -(1-\beta_1)m^i(\tau) + (1-\beta_1)g^i(W^i(\tau)).\end{equation}
Unlike the equation of $m$ which is linear, the equation of $v$ is nonlinear due to the presence of the nonlinear term $(g(W))^2 = \big(\sum_{i=1}^M \frac{N^i}{N} g^i(W)\big)^2$. It is apparent that \begin{equation*}\big(\sum_{i=1}^M \frac{N^i}{N} g^i(W)\big)^2 \ne \sum_{i=1}^M \frac{N^i}{N} (g^i(W))^2.\end{equation*} 
Therefore, it is not possible to decompose the equation of $v$ precisely. Moreover, $ \sum_{i=1}^M \frac{N^i}{N} (g^i(W))^2$ is not even a close approximation of $\big(\sum_{i=1}^M \frac{N^i}{N} g^i(W)\big)^2$.  An immediate counter example is that $(1-1)^2 \neq 1^2+(-1)^2$. For centralized Adam, the expectation of each component of $\hat{m}/\sqrt{\hat v}$ is approximately $\pm 1$
, which is a crucial point of Adam optimizer. Thus, when attempting to generalize centralized Adam optimizer to FL, the key is that $m$, $v$ must be highly synchronized as in centralized Adam. There are two possible resolutions. First, as in previous arguments of precise decomposition, by reducing local iteration number to 1, one obtains a precise decomposition of Eq.\eqref{ADAM-ODE}. Consequently, the FL training matches centralized training. However, the communication cost is too expensive to be affordable for this approach. Another possible way is to utilize Adam optimizer on server, i.e., local models are trained with SGDM and then server aggregates total gradients uploaded by local devices and input the aggregation into a global Adam. This is the same as FedOpt method. However, on second thought, though $v$ cannot be precisely decomposed, the global momentum $m$ can be decomposed precisely as mentioned before and it still can be utilized fully. Therefore, we propose our FedDA+Adam method as follows. 
At round $E$, pick participating clients $S^1,\cdots, S^k$. For each $S^i$, $i = 1,\cdots, K$, initialize $P^i=0$, $W^i(0) = W(E)$, $m^i(0) = m(E)$. For $t = 0, T-1$,
\begin{equation}
\begin{split}
&W^i(t+1) = W^i(t) -  g^i(W^i(t))*\eta\\
&m^i(t+1) = \beta_1 m^i(t) + (1-\beta_1) g^i(W^i(t)) \\
& P^i = P^i + m^i(t+1)
\end{split}
\end{equation}
The global update rule is:
\begin{equation}
\begin{split}
&P = \text{aggregation of }\; P^i,\\
&m(E+1) = \text{aggregation of}\; m^i(T)\\
&\mathcal{G} =  \big(P - \beta_1*m(E)\big)/(1-\beta_1)\\
&\hat m(E+1) = \big(\beta_1*m(E) + (1-\beta_1)*\mathcal{G}\big)/(1-\beta_1^E)\\
&V(E+1) = \beta_2*V(E) +(1-\beta_2)*\mathcal{G}^2\\
&\hat V(E+1) = V(E)/(1-\beta_2^E)\\
&W (E+1)= W(E) - \hat m(E+1)/(\sqrt{\hat V(E+1)}+\epsilon)*\eta*\alpha
\end{split}
\end{equation}

The idea is to treat $P$ as the  momentum of the global Adam updated with a global gradient $\mathcal{G}$, i.e., $P = \beta_1 + (1-\beta_1)*\mathcal{G}$. Simple algebra gives the global gradient $\mathcal{G} = \big(P - \beta_1*m(E)\big)/(1-\beta_1)$. Then the global velocity $v$  is updated with $\mathcal{G}$ followed by the  update of $W$. 

Recall the AdaGrad optimizer 
\begin{equation}
\begin{split}
&V(t) = V(t-1) +  g(W(t))^2, \\
& W(t+1) = W(t) -  g(W(t))/\big({\sqrt{V(t)}+\epsilon}\big)*\eta.
\end{split}
\end{equation}
The corresponding ODE system is 
\begin{equation}
\begin{split}
&\eta*\frac{d}{d\tau}V(\tau) =  g(W(\tau))^2, \\
& \frac{d}{d\tau}W(\tau) =  -g(W(\tau))/\big({\sqrt{V(\tau)}+\epsilon}\big).
\end{split}
\end{equation}

For the same reason, i.e., the nonlinearity of $\big(\sum_{i=1}^M \frac{N^i}{N} g^i(W)\big)^2$, AdaGrad optimizer is not suitable for decomposing onto local devices as well. As in FedDA+Adam, we propose to utilize the decoupled global momentum to generate a global gradient $\mathcal{G}$ and update the global AdaGrad optimizer with input $\mathcal{G}$. 
\begin{equation}
\begin{split}
&W^i(t+1) = W^i(t) -  g^i(W^i(t))*\eta\\
&m^i(t+1) = \beta_1 m^i(t) + (1-\beta_1) g^i(W^i(t)) \\
& P^i = P^i + m^i(t+1)
\end{split}
\end{equation}
The global update rule is:
\begin{equation}
\begin{split}
&P = \text{aggregation of }\; P^i,\\
&m(E+1) = \text{aggregation of}\; m^i(T)\\
&\mathcal{G} =  \big(P - \beta_1*m(E)\big)/(1-\beta_1)\\
&V(E+1) = V(E) + \mathcal{G}^2\\
&W (E+1)= W(E) - \mathcal{G}/(\sqrt{\hat V(E+1)}+\epsilon)*\eta
\end{split}
\end{equation}

By assembling all the pieces in Sections~\ref{sec.FedDA+SGDM} and \ref{sec.FedDA+Adam+AdaGrad}, we provide the pseudo code of our \method\ algorithm with the adaptive optimization of SGDM, Adam, and AdaGrad in Algorithms~\ref{algo:algo1} and~\ref{algo:algo2} respectively.

\setlength{\textfloatsep}{0pt}
\begin{algorithm}[H]
{\small
\caption{\textbf{FedDA+SGDM}} \label{algo:algo1}
\begin{flushleft}
\textbf{Input:} $W_0$, $m_0$ \\
\end{flushleft}
}
\begin{algorithmic}[1]
{\small
\STATE \textbf{for}\; $r = 0,\cdots,R-1$ \textbf{do}
 \STATE \hspace{0.2cm} Sample a subset $S$ of clients 
\STATE \hspace{0.2cm}  \textbf{for} each client $S^i\in S$ \textbf{in parallel do}
\STATE \hspace{0.4 cm} $P^i_{r,0} = 0$, $W^i_{r,0} = W_r$, $m^i_{r,0} = m_r$
\STATE \hspace{0.4cm} \textbf{for} $t = 0,\cdots,T-1$ \textbf{do}
\STATE \hspace{0.6cm} $W^i_{r,t+1} = W^i_{r,t} - g^i_{r,t}(W^i_{r,t})*\eta$
\STATE \hspace{0.6cm} $m^i_{r,t+1} = \beta*m^i_{r,t} + (1-\beta)*g^i_{r,t}(W^i_{r,t})$
\STATE \hspace{0.6cm} $P^i_{r,t+1} =  P^i_{r,t} + m^i_{r,t+1}$
 \STATE \hspace{0.2cm} $P_r = \text{aggregation of }\; P_{r,T}^i$
 \STATE \hspace{0.2cm} $m_{r+1} = \text{aggregation of }\; m_{r,T}^i$
 \STATE \hspace{0.2cm} $W_{r+1} = W_r -P_r*\eta*\alpha$
}
\end{algorithmic}
\end{algorithm}

\setlength{\textfloatsep}{0pt}
\begin{algorithm}[H]
{\small
\caption{\textbf{FedDA+ADAM\& FedDA+AdaGrad}} \label{algo:algo2}
\begin{flushleft}
\textbf{Input:} $W_0$, $m_0$, $V_0$ \\
\end{flushleft}
}
\begin{algorithmic}[1]
{\small
\STATE \textbf{for}\; $r = 0,\cdots,R-1$ \textbf{do}
 \STATE \hspace{0.2cm} Sample a subset $S$ of clients 
\STATE \hspace{0.2cm}  \textbf{for} each client $S^i\in S$ \textbf{in parallel do}
\STATE \hspace{0.4 cm} $P^i_{r,0} = 0$, $W^i_{r,0} = W_r$, $m^i_{r,0} = m_r$
\STATE \hspace{0.4cm} \textbf{for} $t = 0,\cdots,T-1$ \textbf{do}
\STATE \hspace{0.6cm} $W^i_{r,t+1} = W^i_{r,t} - g^i_{r,t}(W^i_{r,t})*\eta$
\STATE \hspace{0.6cm} $m^i_{r,t+1} = \beta_1*m^i_{r,t} + (1-\beta_1)*g^i_{r,t}(W^i_{r,t})$
\STATE \hspace{0.6cm} $P^i_{r,t+1} =  P^i_{r,t} + m^i_{r,t+1}$
 \STATE \hspace{0.2cm} $P_r = \text{aggregation of }\; P_{r,T}^i$
 \STATE \hspace{0.2cm} $m_{r+1} = \text{aggregation of }\; m_{r,T}^i$
 \STATE \hspace{0.2cm} $\mathcal{G}_r = (P_r -\beta_1*m_r)/(1-\beta_1)$
 \STATE \hspace{0.2cm} \textbf{ADAM}
 \STATE \hspace{0.4cm} $\hat m_{r+1}= (\beta_1*m_r +(1-\beta_1)*\mathcal{G}_r)/(1-\beta_1^r)$
 \STATE \hspace{0.4cm} $ V_{r+1}= (\beta_2*V_r +(1-\beta_2)*\mathcal{G}_r^2)$
  \STATE \hspace{0.4cm} $\hat V_{r+1} = V_{r+1}/(1-\beta_2^r)$
   \STATE \hspace{0.4cm} $W_{r+1} = W_r - \hat m_{r+1}/(\sqrt{V_{r+1}}+\epsilon)*\eta*\alpha$
   \STATE \hspace{0.2cm} \textbf{AdaGrad}
 \STATE \hspace{0.4cm} $ V_{r+1}= V_r + \mathcal{G}_r^2$
    \STATE \hspace{0.4cm} $W_{r+1} = W_r - \mathcal{G}_r/(\sqrt{V_{r+1}}+\epsilon)*\eta*\alpha$
}
\end{algorithmic}
\end{algorithm}

\subsection{Advantage of Global Momentum in FedDA}\label{sec.Advantage}

Assume that \begin{enumerate}
\item (Lipschitz Gradient). There exists a constant $L_g$ such that $\|g^i(W_1) - g^i(W_2)\|\le L_g\|W_1-W_2\|$ for any $W_1,W_2$ and $i = 1,\cdots,m$.
\item (Bounded Gradient) $\|g^i\|_{L^\infty}<\infty$ for any $i=1,2,\cdots,m$.
\end{enumerate}
Under these assumptions, we theoretically demonstrate that (1) local momentum deviates from the centralized one at exponential rate $O(e^{\lambda t})$; and (2) global momentum in FedDA deviates from the centralized one at algebraic rate $O(t^2)$.

{\bf Local momentum deviates from the centralized one at exponential rate.}
Let $(m^i(\tau),W^i(\tau))$ be the solution to the  decomposed system 
\begin{equation}
\begin{split}
&\eta\frac{d}{d\tau}m^i(\tau) = -(1-\beta)m^i(\tau) + (1-\beta)g^i(W^i(\tau)), \\
&\frac{d}{d\tau}W^i(\tau)= - m^i(\tau).
\end{split}
\end{equation}
Let $(m(\tau),W(\tau))$ be the solution to Eq.\eqref{SGDm-ODE} with $(m(0),W(0)) = (m^i(0), W^i(0))$, i.e., centralized SGDm optimizer with the same initialization as the decomposed optimizer. By direct calculations, we have
\begin{equation}\label{diffmim}
\begin{split}
&\eta\frac{d}{d\tau}(m^i(\tau) - m(\tau)) = (1-\beta)\Big(-(m^i(\tau)-m(\tau)) + \big(g^i(W^i(\tau)) - g^i(W(\tau))\big)+R(\tau)\Big), \\
&\frac{d}{d\tau}(W^i(\tau) - W(\tau)) = - (m^i(\tau) - m(\tau)),
\end{split}
\end{equation}
where $R(\tau) = \sum_{j\ne i}\frac{N_j}{N}\big(g^j(W(\tau)) - g^i(W(\tau))\big)$. One can check that $\nabla\|x\| = \frac{x}{\|x\|}$ for nay $x\in \mathbb{R}^n$. Therefore, it holds that $\|\nabla \|x\|\|\le 1$ and $x\cdot \nabla \|x\| = \|x\|$, where $\cdot$ means dot product. Taking the inner product of the first equation in Eq.\eqref{diffmim} with $\nabla\|m^i(\tau) - m(\tau)\|$, we obtain
\begin{equation}\label{diffm}
\begin{split}
\eta\frac{d}{d\tau}\|m^i(\tau) - m(\tau)\| \le (1-\beta)\Big(&-\|m^i(\tau)-m(\tau)\| + L_g\|W^i(\tau)-W(\tau)\|+\|R(\tau)\Big).
\end{split}
\end{equation}

Similarly, take the inner produce of the second equation in Eq.\eqref{diffmim} with $\nabla\|W^i - W\|$, we obtain
\begin{equation}\label{diffW}
\frac{d}{d\tau}\|W^i(\tau) - W(\tau)\|\le \|m^i(\tau) - m(\tau)\|
\end{equation}
The system of Eq.\eqref{diffm} and Eq.\eqref{diffW} can be written as the following matrix form. 
\begin{equation}\label{diffmatrix}
\frac{d}{d\tau}\begin{pmatrix}\|m^i(\tau) - m(\tau)\| \\ \|W^i(\tau) - W(\tau)\| \end{pmatrix}\le A\begin{pmatrix}\|m^i(\tau) - m(\tau)\| \\ \|W^i(\tau) - W(\tau)\| \end{pmatrix} +(1-\beta)\begin{pmatrix}\|R(\tau)\| /\eta\\ 0 \end{pmatrix}
\end{equation}
where $ A = \begin{pmatrix} -(1-\beta)/\eta&(1-\beta)L_g/\eta \\ 1& 0 \end{pmatrix}$. 
By direct computations, the eigenvalue of the matrix $A$ is $\lambda^\pm = \frac{-(1-\beta)\pm \sqrt{(1-\beta)^2+4(1-\beta)L_g}}{2\eta}$. Therefore $\|e^{At}\|\le C_Ae^{\lambda^+ t}$ for any $t\ge 0$ for some constant $C_A$. Note that $\lambda^+>0$.   Applying variation of constants formula to Eq.\eqref{diffmatrix}, one has
\begin{equation}
\begin{pmatrix}\|m^i(t\eta) - m(t\eta)\| \\ \|W^i(t\eta) - W(t\eta)\| \end{pmatrix} \le e^{At\eta}\begin{pmatrix}\|m^i(0) - m(0)\| \\ \|W^i(0) - W(0)\| \end{pmatrix} + (1-\beta)\int_0^{t\eta} e^{A(t\eta-\tau)}\begin{pmatrix}\|R(\tau)\| /\eta\\ 0 \end{pmatrix}d\tau.
\end{equation}
Note that $m^i(0) - m(0)=0$ and $W^i(0) - W(0)$. Therefore, for any $t$, it holds
\begin{equation}
\begin{pmatrix}\|m^i(t\eta) - m(t\eta)\| \\ \|W^i(t\eta) - W(t\eta)\| \end{pmatrix} \le (1-\beta)\int_0^{t\eta} e^{A(t\eta-\tau)}\begin{pmatrix}\|R(\tau)\| /\eta\\ 0 \end{pmatrix}d\tau.
\end{equation}
Utilizing  $\|e^{At}\|\le C_Ae^{\lambda^+ t}$ for $t\ge 0$ in the above inequality , we obtain
\begin{equation}
\begin{split}
&\|m^i(t\eta) - m(t\eta)\| + \|W^i(t\eta) - W(t\eta)\|\\
 \le &\int_0^{t\eta} e^{\lambda^+(t\eta-\tau)}\|R(\tau)\| /\eta d\tau = e^{\lambda^+t\eta}(1-\beta)\int_0^{t\eta} e^{-\lambda^+ \tau}\begin{pmatrix}\|R(\tau)\| /\eta\end{pmatrix}d\tau,
\end{split}
\end{equation}
which implies that $\|m^i(t\eta) - m(t\eta)\|$ is exponentially growing.   Essentially, the term $\big(g^i(W^i(\tau)) - g^i(W(\tau))$ in Eq. \eqref{diffmim} causes such exponential growth. More precisely,  since $g^i(W^i) - g^i(W) = \nabla g^i(W)(W^i-W) + O(|W^i-W|^2)$, it contributes a linear term $ \nabla g^i(W)(W^i-W)$ to Eq.\eqref{diffmim}. Consequently, the matrix $A$ has a positive eigenvalue, which implies that $\|e^{At}\|$ is exponentially growing.  Thus, the difference of momentum $\|m^i(t\eta) - m(t\eta)\|$ is exponentially growing. However, in our FedDA method, since momentum is decoupled with local training, there is no such a term as $g^i(W^i) - g^i(W)$. Therefore, in our FedDA framework, the matrix analogous to matrix $A$ is  $ \begin{pmatrix} -(1-\beta)/\eta&0\\ 1& 0 \end{pmatrix}$, whose eigenvalues are $0,-(1-\beta)$. Therefore, the momentum deviation in our FedDA method is only algebraic growing. 

{\bf Global momentum in FedDA deviates from the centralized one at algebraic rate.}
Let $(\bar{W}^i(\tau), \bar m(\tau), \bar W(\tau))$ be the solution to Eq.\eqref{SGDm-Decouple} with $\alpha =1$. Let $(m(\tau),W(\tau))$ be the solution to Eq.\eqref{SGDm-ODE} with $(m(0),W(0)) = (\bar m(0), \bar W(0))$. It is straightforward to compute 
\begin{equation}\label{diffmim-2}
\begin{split}
&\eta\frac{d}{d\tau}(\bar m(\tau) - m(\tau)) = -(1-\beta)\Big((\bar m(\tau)-m(\tau)) + \sum_i^M\frac{N_i}{N}\big(g^i(\bar W^i(\tau)) - g^i(W(\tau))\big)\Big) \\
&\frac{d}{d\tau}(\bar W(\tau) - W(\tau)) = - (\bar m(\tau) - m(\tau)),
\end{split}
\end{equation}
By calculations similar to the above ones, we first have 
\begin{equation}\label{diffmatrix-2}
\frac{d}{d\tau}\begin{pmatrix}\|\bar m(\tau) - m(\tau)\| \\ \|\bar W(\tau) - W(\tau)\| \end{pmatrix}\le \bar A\begin{pmatrix}\|\bar m(\tau) - m(\tau)\| \\ \|\bar W(\tau) - W(\tau)\| \end{pmatrix} +(1-\beta)\begin{pmatrix}\|\sum_i^M\frac{N_i}{N}\big(g^i(\bar W^i(\tau)) - g^i(W(\tau))\big)\| /\eta\\ 0 \end{pmatrix}
\end{equation}
where $ \bar A = \begin{pmatrix} -(1-\beta)/\eta&0 \\ 1& 0 \end{pmatrix}$. One can check that the eigenvalues of $\bar A$ are $0,-(1-\beta)/\eta$, which implies that there exists a constant $C_{\bar A }$ such that $\|e^{\bar A t}\|\le C_{\bar A}$ for any $t\ge 0$, i.e., $\|e^{\bar A t}\|$ is uniformly bounded without growth. Clearly, 
\begin{equation}\label{z1}
\|g^i(\bar W^i(\tau)) - g^i(W(\tau))\|\le L_g \|\bar W^i(\tau) - W(\tau)\|.
\end{equation}
Moreover, utilizing $\bar W^i(0) = W(0)$, we have 
\begin{equation} 
\bar W^i(\tau) - W(\tau) = \int_0^\tau -g^i(\bar W^i(s)) + m(s)ds.
\end{equation}
It follows that 
\begin{equation} 
\|\bar W^i(\tau) - W(\tau)\| \le \big(\|g^i\|_{L^\infty} + \underset{s\le \tau}{\sup} \|m(s)\|\big)\tau.
\end{equation}
By variation of constants formula, we have 
\begin{equation}
\|m(\tau)\| \le e^{-(1-\beta)\tau/\eta}\|m(0)\| +\frac{1}{\eta} \int_0^\tau e^{-(1-\beta)(\tau-s)/\eta}(1-\beta)\|g(W(s))\|ds,
\end{equation}
which implies that $|m(\tau)|\le |m(0)| + \|g\|_{L^\infty}$ for any $\tau\ge 0$.
Therefore, we have 
\begin{equation} \label{z2}
\|\bar W^i(\tau) - W(\tau)\| = \big(\|g^i\|_{L^\infty} + \|m(0)\| + \|g\|_{L^\infty})\tau.
\end{equation}
Combining Eq.\eqref{z1} and Eq.\eqref{z2}, we have 
\begin{equation}\label{z3}
\|g^i(\bar W^i(\tau)) - g^i(W(\tau))\|\le \|\nabla g^i\|_{L^\infty}  \big(\|g^i\|_{L^\infty} + \|m(0)\| + \|g\|_{L^\infty})\tau.
\end{equation}
Applying the variation of constants formula to Eq.\eqref{diffmatrix-2}, we have 
\begin{equation}
\begin{pmatrix}\|\bar m(t\eta) - m(t\eta)\| \\ \|\bar W(t\eta) - W(t\eta)\| \end{pmatrix}\le \int_0^{t\eta} e^{\bar A(t\eta-\tau)}\begin{pmatrix}\|\sum_i^M\frac{N_i}{N}\big(g^i(\bar W^i(\tau)) - g^i(W(\tau))\big)\| /\eta\\ 0 \end{pmatrix}d\tau.
\end{equation}
Utilizing $\|e^{\bar A t}\|\le C_{\bar A}$  and Eq.\eqref{z3}, it holds 
\begin{equation}
\|\bar m(t\eta) - m(t\eta)| + \|\bar W(t\eta) - W(t\eta)\|\le \int_0^{t\eta} C^\ast \tau/\eta d\tau = \frac{C^\ast\eta}{2}t^2,
\end{equation}
where $C^\ast = \underset{i}{\sup}\|\nabla g^i\|_{L^\infty}  \big(\|g^i\|_{L^\infty} + \|m(0)\| + \|g\|_{L^\infty}) $.
Therefore, the momentum in our FedDA method deviates from the centralized one in algebraic rate $O(t^2)$, which is much slower than that of local momentum.

\subsection{Additional Experiments}\label{sec.AdditionalExperiments}

In this section, we conduct more experiments to validate the accuracy and convergence of our proposed \method method and evaluate the sensitivity of client and server learning rates in our momentum decoupling adaptive optimization method for the FL  task.

\begin{figure}[H]
\mbox{
\subfigure[SGDM]{\epsfig{figure=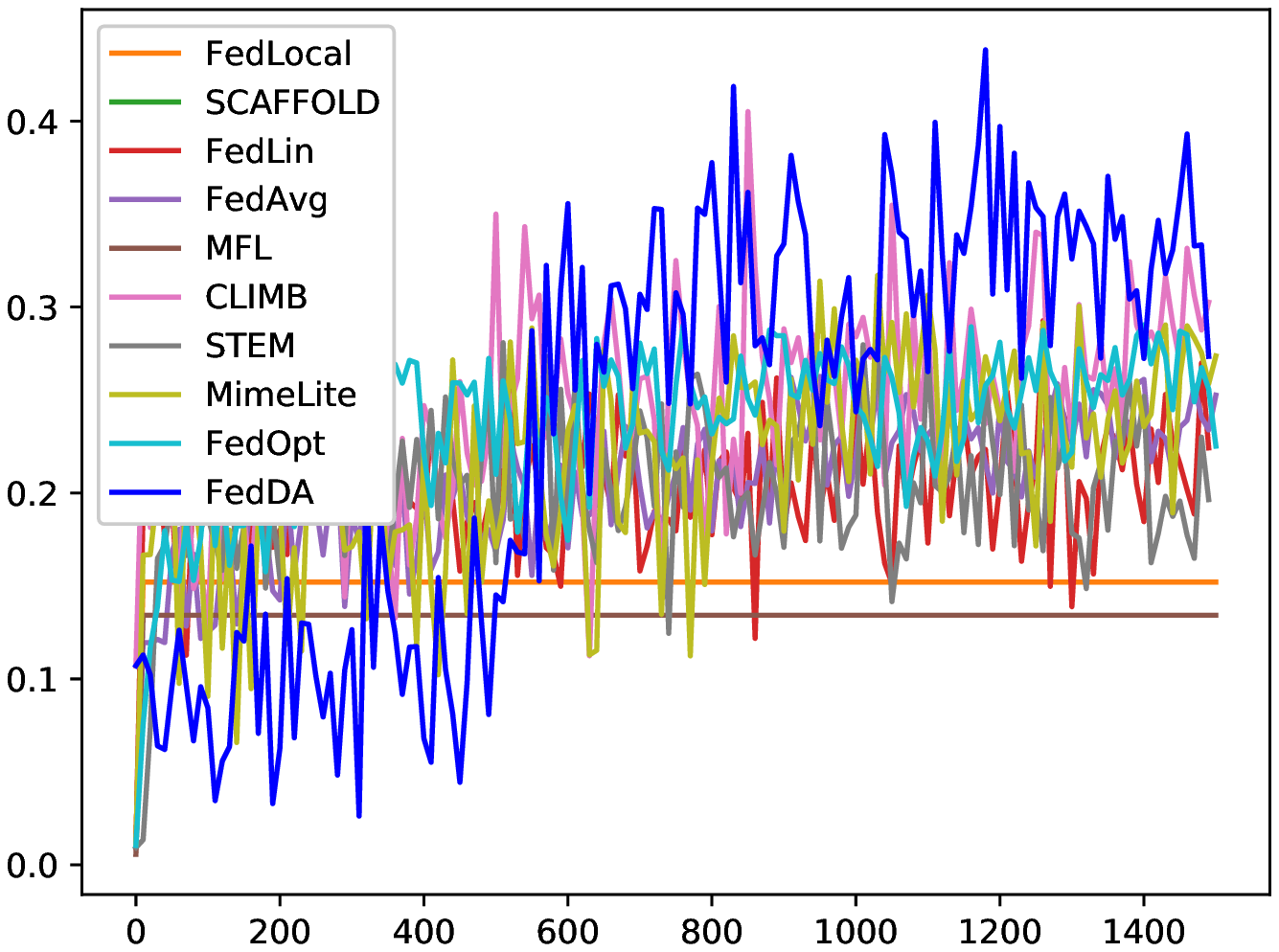, height=2in, width=0.35\linewidth}} \hspace{-0.5cm}
\subfigure[Adam]{\epsfig{figure=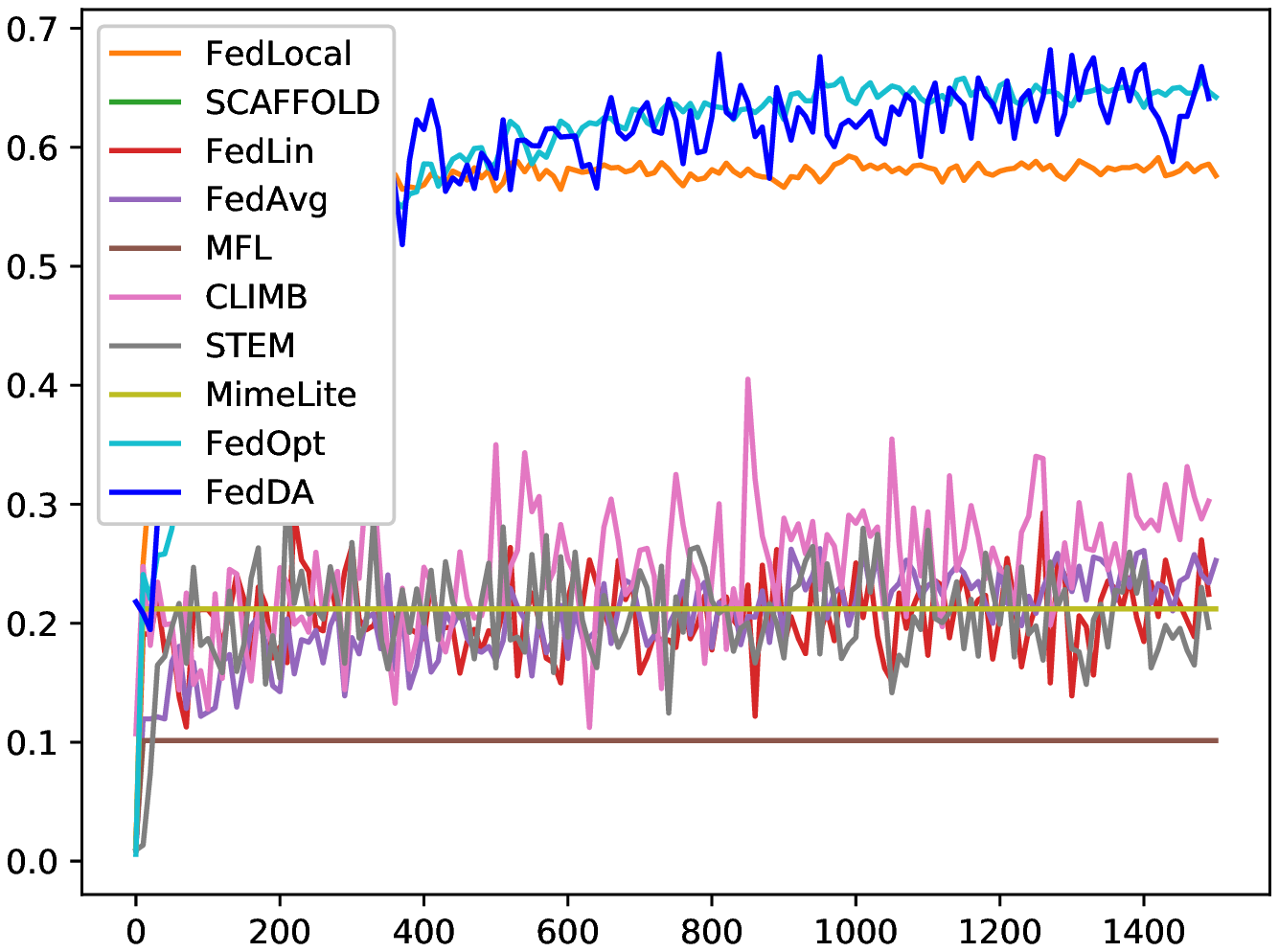, height=2in, width=0.35\linewidth}} \hspace{-0.5cm}
\subfigure[AdaGrad]{\epsfig{figure=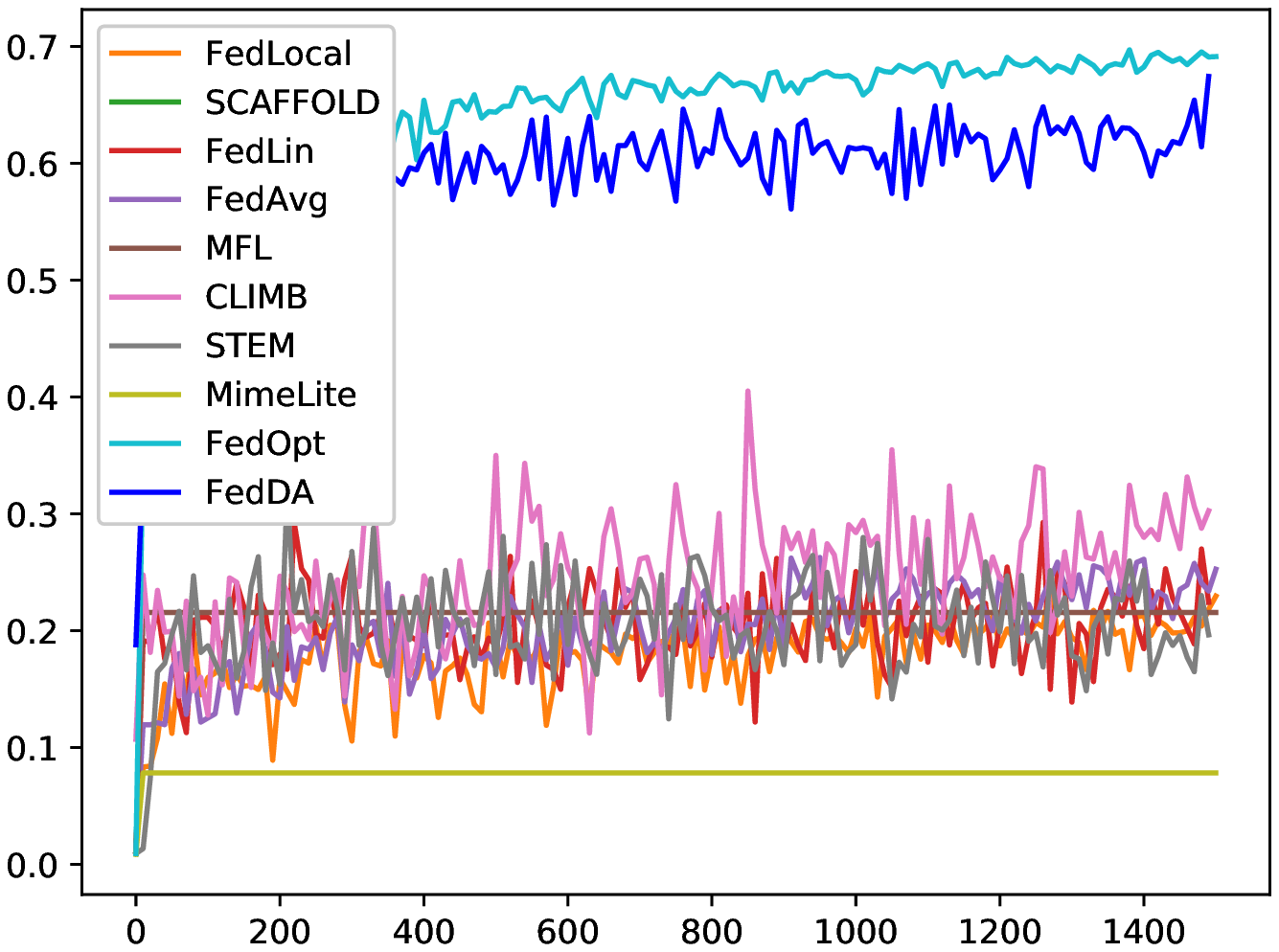, height=2in, width=0.35\linewidth}}}
\caption{Convergence on Stack Overflow with Three Optimizers}
\label{fig.StackOverflowConvergence}
\end{figure}

\begin{figure}[H]
\mbox{
\subfigure[SGDM]{\epsfig{figure=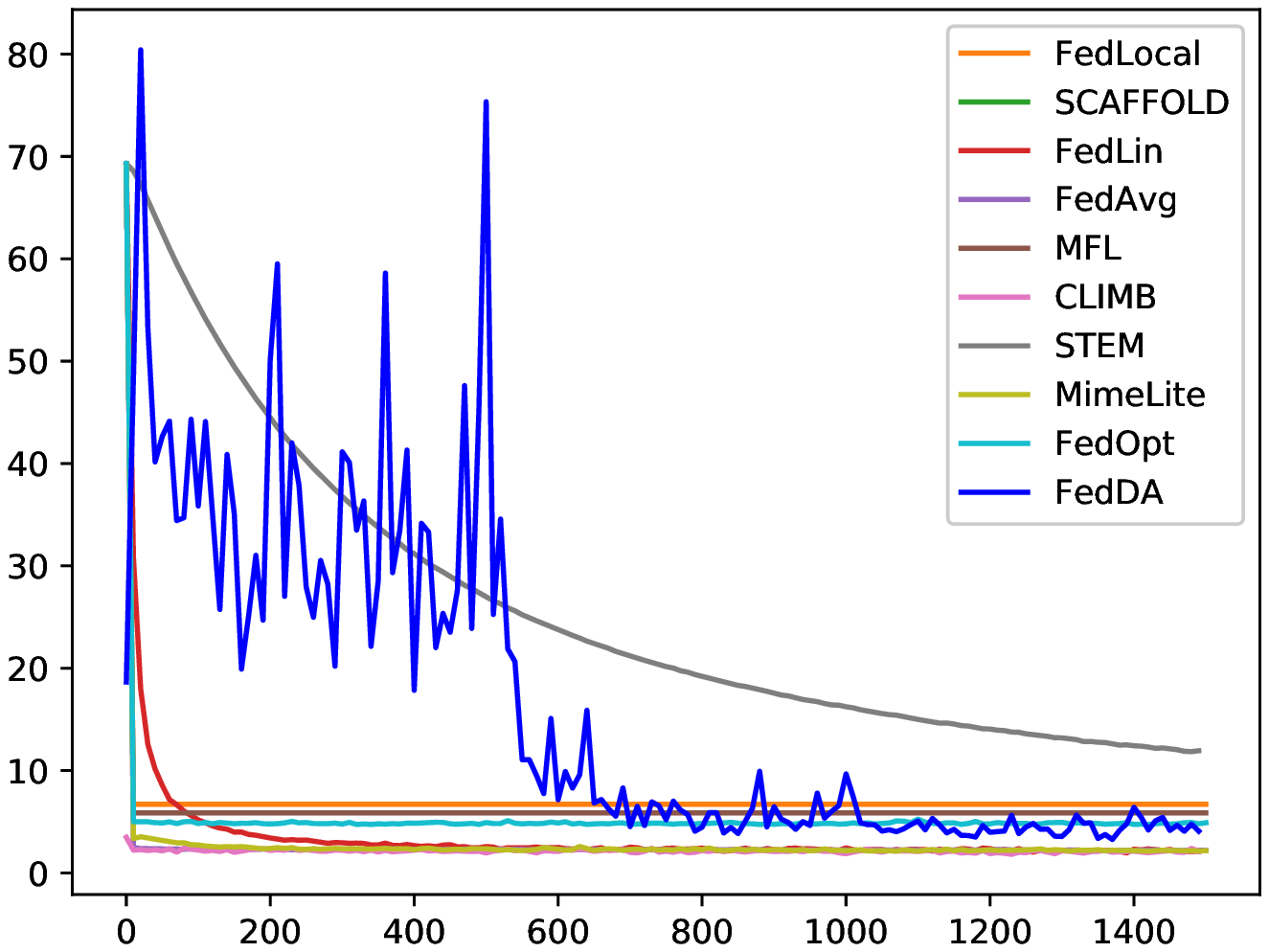, height=2in, width=0.35\linewidth}} \hspace{-0.5cm}
\subfigure[Adam]{\epsfig{figure=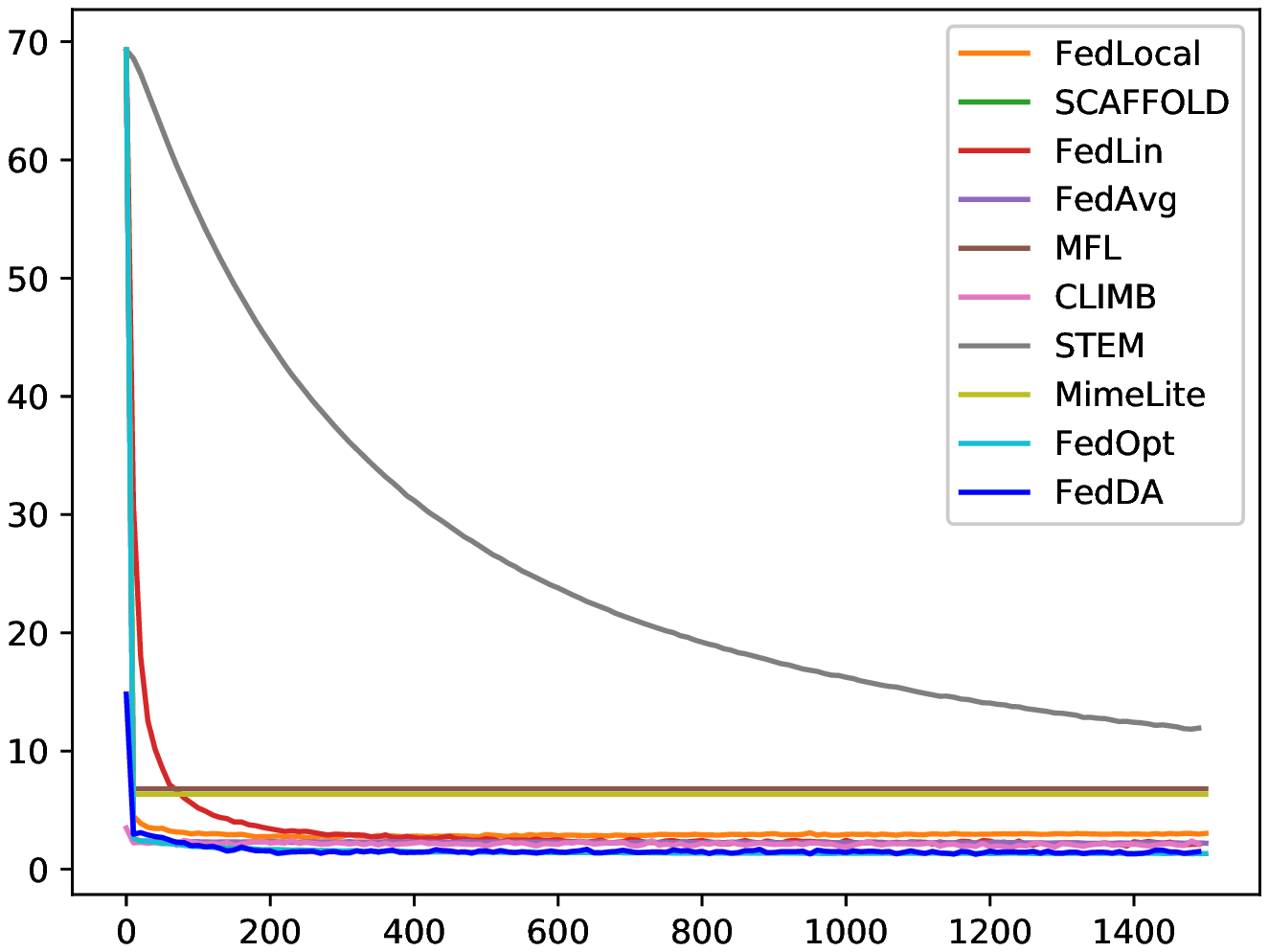, height=2in, width=0.35\linewidth}} \hspace{-0.5cm}
\subfigure[AdaGrad]{\epsfig{figure=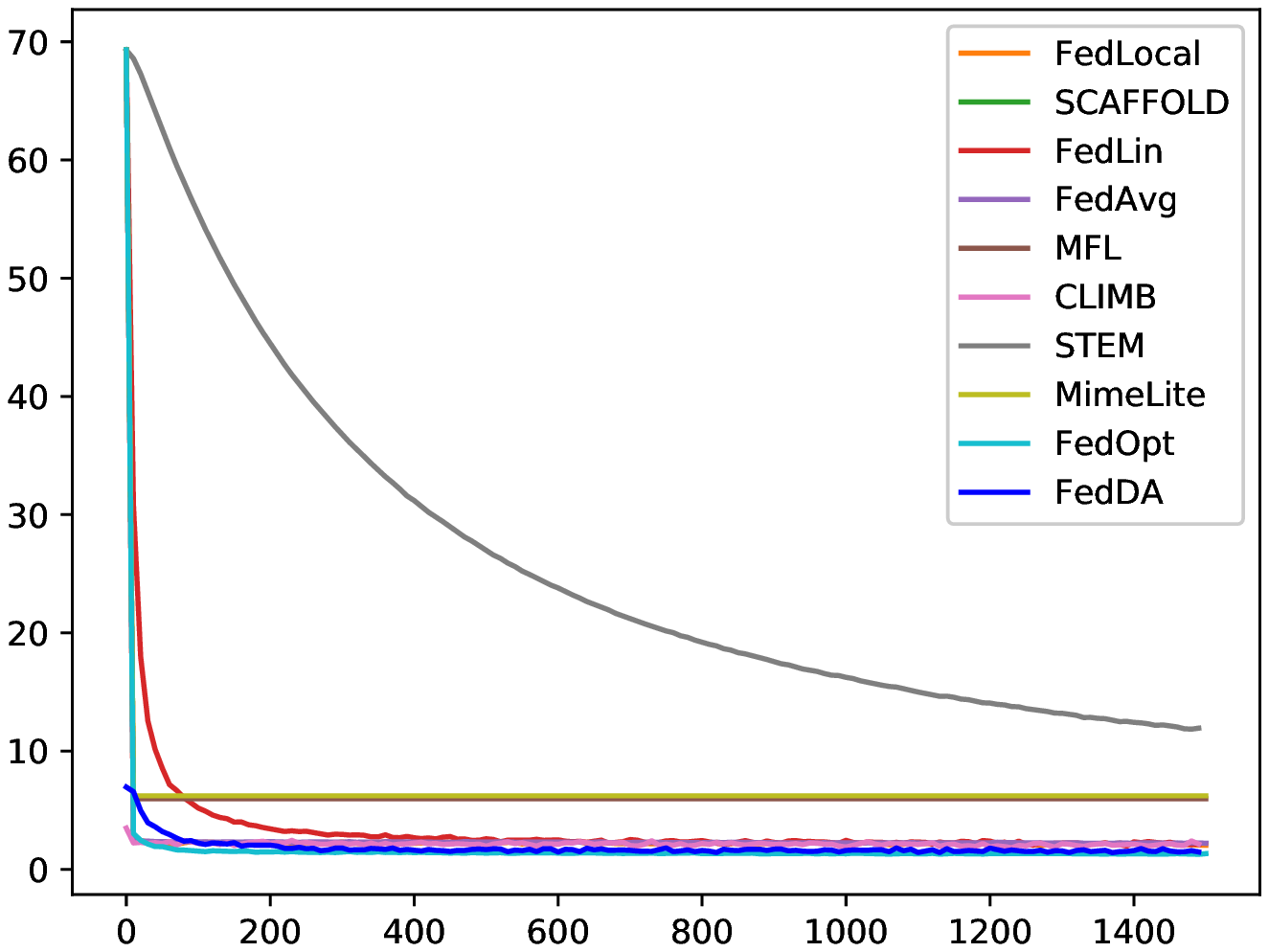, height=2in, width=0.35\linewidth}}}
\caption{Loss on Stack Overflow with Three Optimizers}
\label{fig.StackOverflowLoss}
\end{figure}

{\bf Convergence and loss over Stack Overflow.} Figures \ref{fig.StackOverflowConvergence}-\ref{fig.StackOverflowLoss} present the $Accuracy$ and $Loss$ curves of ten federated learning algorithms on Stack Overflow.
Similar trends are observed for the performance comparison: \method\ achieves the 75.4\% convergence improvement, which are obviously better than other methods in most experiments. This demonstrates that the full batch gradient techniques that mimic centralized optimization in the end of the training process are able to ensure the convergence and overcome the possible inconsistency caused by adaptive optimization methods.

{\bf Final $Accuracy$ over Stack Overflow.} Table \ref{tbl.FinalAccuracyStackOverflow} presents the final $Accuracy$ scores of ten federated learning algorithms on Stack Overflow. We have observed similar trends: the accuracy achieved by our \method\ is the highest in most tests. Especially, as shown the experiment with SGDM as the optimizer, compared to the best competitors among ten federated learning algorithms, the final $Accuracy$ scores achieved by \method\ averagely achieves 22.3\% improvement. A rational guess is that the global momentum in our \method\ method makes the best effort to mimic the role of momentum in centralized training, which can accelerate the convergence of FL training. 

\begin{table}[H]
\caption{Final Accuracy on Stack Overflow}
\begin{center}
\begin{tabular}{l|ccc}
\hline {\bf Optimizer} & {\bf SGDM} & {\bf Adam} & {\bf AdaGrad} \\
\hline FedLocal & 0.152  & 0.576  & 0.229 \\
SCAFFOLD & 0.250  & 0.250  & 0.250 \\
FedLin & 0.224  & 0.224  & 0.224 \\
FedAvg & 0.252  & 0.252  & 0.252 \\
MFL & 0.134  & 0.101  & 0.215 \\
CLIMB & 0.302  & 0.302  & 0.302 \\
STEM & 0.196  & 0.196  & 0.196 \\
MimeLite & 0.271  & 0.211  & 0.078 \\
FedOpt & 0.225  & {\bf 0.642}  & {\bf 0.691} \\
\hline FedDA & {\bf 0.273} & {\bf 0.642}  & 0.674 \\
\hline
\end{tabular}
\label{tbl.FinalAccuracyStackOverflow}
\end{center}
\end{table}

{\bf Impact of client and server learning rates.} 
Tables~\ref{tbl.SGDMVaryingClient}-\ref{tbl.AdaGradVaryingServer} report how the test $Accuracy$ changes with server and client learning rates on three datasets by fixing the server learning rates and changing the client learning rates, or by utilizing the reverse settings. We have observed that the $Accuracy$ scores oscillate within the range of 0.002 and 0.868 when changing the client learning rates, while the $Accuracy$ values fluctuate between 0.010 and 0.861. This demonstrates that it is crucial to choose the optimal learning rates for the training on the clients and server to achieve the competitive performance. Please refer to Table \ref{tbl.HyperparameterSettings} for the implementation details of the server and client learning rates used in our current experiments.

\begin{table}[H]
\caption{Final Accuracy with SGDM Optimizer and Varying Client Learning Rate}
\begin{center}
\begin{tabular}{l|c|cccc}
\hline {\bf Dataset} & {\bf Server Learning Rate} & \multicolumn{4}{|c} {\bf Accuracy / Client Learning Rate} \\
\hline CIFAR-100 & 0.1 & 0.521 / 1 & 0.448 / 3.3 & 0.236 / 10 & 0.099 / 33 \\
EMNIST & 0.1 & 0.778 / 1 & 0.819 / 3.3 & 0.783 / 10 & 0.052 / 33 \\
Stack Overflow & 0.1 & 0.174 / 1 & 0.262 / 10 & 0.250 / 100 & 0.002 / 1,000 \\
\hline
\end{tabular}
\label{tbl.SGDMVaryingClient}
\end{center}
\end{table}

\begin{table}[H]
\caption{Final Accuracy with Adam Optimizer and Varying Client Learning Rate}
\begin{center}
\begin{tabular}{l|c|cccc}
\hline {\bf Dataset} & {\bf Server Learning Rate} & \multicolumn{4}{|c} {\bf Accuracy / Client Learning Rate} \\
\hline CIFAR-100 & 0.1 & 0.179 / 1 & 0.317 / 3.3 & 0.269 / 10 & 0.179 / 33 \\
EMNIST & 0.03 & 0.488 / 0.01 & 0.836 / 0.1 & 0.842 / 0.3 & 0.804 / 1 \\
Stack Overflow & 0.3 & 0.409 / 0.1 & 0.492 / 1 & 0.473 / 10 & 0.459 / 100 \\
\hline
\end{tabular}
\label{tbl.AdamVaryingClient}
\end{center}
\end{table}

\begin{table}[H]
\caption{Final Accuracy with AdaGrad Optimizer and Varying Client Learning Rate}
\begin{center}
\begin{tabular}{l|c|cccc}
\hline {\bf Dataset} & {\bf Server Learning Rate} & \multicolumn{4}{|c} {\bf Accuracy / Client Learning Rate} \\
\hline CIFAR-100 & 0.1 & 0.462 / 0.03 & 0.488 / 0.1 & 0.410 / 0.3 & 0.259 / 1 \\
EMNIST & 0.1 & 0.055 / 0.00001 & 0.055 / 0.001 & 0.055 / 0.01 & 0.868 / 0.1 \\
Stack Overflow & 10 & 0.576 /1 & 0.651 / 10 & 	0.629 / 30 & 0.632 / 100 \\
\hline
\end{tabular}
\label{tbl.AdaGradVaryingClient}
\end{center}
\end{table}

\begin{table}[H]
\caption{Final Accuracy with SGDM Optimizer and Varying Server Learning Rate}
\begin{center}
\begin{tabular}{l|c|cccc}
\hline {\bf Dataset} & {\bf Client Learning Rate} & \multicolumn{4}{|c} {\bf Accuracy / Server Learning Rate} \\
\hline CIFAR-100 & 0.03 & 0.374 / 1 & 0.522 / 3.3 & 0.356 / 10 & 0.273 / 33 \\
EMNIST & 0.1 & 0.859 / 0.3 & 0.861 / 1 & 0.778 / 3.3 & 0.783 / 10 \\
Stack Overflow & 100 & 0.191 / 0.001 & 0.256 / 0.003 & 0.204 / 0.01 & 0.292 / 0.1 \\
\hline
\end{tabular}
\label{tbl.SGDMVaryingServer}
\end{center}
\end{table}

\begin{table}[H]
\caption{Final Accuracy with Adam Optimizer and Varying Server Learning Rate}
\begin{center}
\begin{tabular}{l|c|cccc}
\hline {\bf Dataset} & {\bf Client Learning Rate} & \multicolumn{4}{|c} {\bf Accuracy / Server Learning Rate} \\
\hline CIFAR-100 & 0.03 & 0.510 / 0.33 & 0.183 / 3.3 & 0.010 / 10 & 0.010 / 33 \\
EMNIST & 0.03 & 0.546 / 0.1 & 0.803 / 1 & 0.051 / 3.3 & 0.051 / 10 \\
Stack Overflow & 100 & 0.425 / 0.1 & 0.522 / 0.3 & 0.641 / 1 & 0.633 / 10 \\
\hline
\end{tabular}
\label{tbl.AdamVaryingServer}
\end{center}
\end{table}

\begin{table}[H]
\caption{Final Accuracy with AdaGrad Optimizer and Varying Server Learning Rate}
\begin{center}
\begin{tabular}{l|c|cccc}
\hline {\bf Dataset} & {\bf Client Learning Rate} & \multicolumn{4}{|c} {\bf Accuracy / Server Learning Rate} \\
\hline CIFAR-100 & 0.03 & 0.352 / 0.03 & 0.462 / 0.1 & 0.466 / 0.3 & 0.312 / 1 \\
EMNIST & 0.03 & 0.806 / 0.1 & 0.055 / 3.3 & 0.055 / 10 & 0.055 / 33 \\
Stack Overflow & 100 & 0.227 / 0.1 & 0.306 / 0.3 & 0.397 / 1 & 0.632 / 10 \\
\hline
\end{tabular}
\label{tbl.AdaGradVaryingServer}
\end{center}
\end{table}

\subsection{Experimental Details}\label{sec.ExperimentDetails}

{\bf Environment.} Our experiments were conducted on a compute server running on Red Hat Enterprise Linux 7.2 with 2 CPUs of Intel Xeon E5-2650 v4 (at 2.66 GHz) and 8 GPUs of NVIDIA GeForce GTX 2080 Ti (with 11GB of GDDR6 on a 352-bit memory bus and memory bandwidth in the neighborhood of 620GB/s), 256GB of RAM, and 1TB of HDD. Overall, our experiments took about 2 days in a shared resource setting. We expect that a consumer-grade single-GPU machine (e.g., with a 1080 Ti GPU) could complete our full set of experiments in around tens of hours, if its full resources were dedicated.

All the codes were implemented based on the Tensorflow Federated (TFF) package~\cite{InOs19}.
Clients are sampled uniformly at random, without replacement in a given round, but with replacement across rounds. Our implementation follows the same settings in the approaches of FedAvg~\cite{MMRH17} and FedOpt~\cite{RCZG21}. First, instead of doing $K$ training steps per client, we do $E$ epochs of training over each client’s dataset. Second, to account for varying numbers of gradient steps per client, we weight the average of the client outputs by each client’s number of training samples.
Since the datasets used are all public datasets and the hyperparameter settings are explicitly described, our experiments can be easily reproduced on top of a GPU server. We promise to release our open-source codes on GitHub and maintain a project website with detailed documentation for long-term access by other researchers and end-users after the paper is accepted.

{\bf Datasets.} We following the same strategy in FedOpt~\cite{RCZG21} to create a federated version of CIFAR-100 by randomly partitioning the training data among 500 clients, with each client receiving 100 examples. We randomly partition the data to reflect the coarse and fine label structure of CIFAR-100 by using the Pachinko Allocation Method~\cite{LiMc06}. In the derived client datasets, the label distributions better resemble practical heterogeneous client datasets. We train a modified ResNet-18, where the batch normalization layers are replaced by group normalization layers~\cite{WuHe20}. We use two groups in each group normalization layer. Group normalization can lead to significant gains in accuracy over batch normalization in federated settings~\cite{HPMG20}.

The federated version of EMNIST partitions the digits by their author~\cite{CWLK18}. The dataset has natural heterogeneity stemming from the writing style of each person. We use a convolutional network for character recognition. The network has two convolutional layers with $3 \times 3$ kernels, max pooling, and dropout, followed by a 128 unit dense layer.

Stack Overflow is a language modeling dataset consisting of question and answers from the question and answer site~\cite{SO19}. The questions and answers also have associated metadata, including tags. The dataset contains 342,477 unique users which we use as clients. We choose the 10,000 most frequently used words, the 500 most frequent tags and adopt a one-versus-rest classification strategy, where each question/answer is represented as a bag-of-words vector.

{\bf Implementation.}
For four regular federated learning models of
FedAvg~\footnote{https://github.com/google-research/federated/tree/780767fdf68f2f11814d41bbbfe708274eb6d8b3/optimization},
SCAFFOLD~\footnote{https://github.com/google-research/public-data-in-dpfl}, 
STEM~\footnote{https://papers.neurips.cc/paper/2021/hash/3016a447172f3045b65f5fc83e04b554-Abstract.html}, and 
CLIMB~\footnote{https://openreview.net/forum?id=Xo0lbDt975}, we used the open-source implementation and default parameter settings by the original authors or the Google Research for our experiments.
For three federated optimization approaches of
MimeLite~\footnote{https://github.com/google-research/public-data-in-dpfl},
FedOpt~\footnote{https://github.com/google-research/federated/tree/master/optimization}, and 
FedLocal~\footnote{https://github.com/google-research/federated/tree/master/local\_adaptivity}, 
we also utilized the same model architecture as the official implementation provided by the Google Research and used the same datasets to validate the performance of these federated optimization models in all experiments. 
For other regular federated learning or federated optimization approaches, including MFL and FedLin, to our best knowledge, there are no publicly available open-source implementations on the Internet. All hyperparameters are standard values from the reference works. The above open-source codes from the GitHub are licensed under the MIT License, which only requires preservation of copyright and license notices and includes the permissions of commercial use, modification, distribution, and private use.

For our proposed decoupled adaptive optimization algorithm, we performed hyperparameter selection by performing a parameter sweep on training rounds $\in \{1,500, 2,000, 2,500, 3,000, 3,500, 4,000\}$, momentum parameter $\beta_1 \in \{0.84, 0.86, 0.88, 0.9, 0.92, 0.94\}$, second moment parameter $\beta_2 \in \{0.984, 0.986, 0.988, 0.99, 0.992, 0.994\}$, fuzz factor $\epsilon \in \{0.00001, 0.0001, 0.001, 0.01, 0.1\}$, local iteration number $\in \{1, 2, 5, 10, 20\}$, and learning rate $\eta \in \{0.001, 0.003, 0.01, 0.03, 0.1, 0.3, 1, 3.3, 10, 33, 100, 333, 1,000\}$.
The above search process is often done using validation data in centralized settings. However, such data is often inaccessible in federated settings, especially cross-device settings. Therefore, we tune by selecting the parameters that minimize the average training loss over the last 100 rounds of training. We run 1,500 rounds of training on the EMNIST and Stack Overflow, and 4,000 rounds over the CIFAR-100.

{\bf Hyperparameter settings.}

Unless otherwise explicitly stated, we used the following default parameter settings in the experiments, as shown in Table \ref{tbl.HyperparameterSettings}.

\begin{table}[H]\addtolength{\tabcolsep}{-4pt}
\caption{Hyperparameter Settings}
\begin{center}
\begin{tabular}{|c|c|}
\hline \textbf{Parameter} & \textbf{Value} \\
\hline Training rounds for EMNIST and Stack Overflow & 1,500 \\
\hline Training rounds for CIFAR-100 & 4,000 \\
\hline Momentum parameter $\beta_1$ & 0.9 \\
\hline Second moment parameter $\beta_2$ & 0.99 \\
\hline Client learning rate with SGDM on CIFAR-100 & 0.03 \\
\hline Client learning rate with Adam on CIFAR-100 & 0.03 \\
\hline Client learning rate with AdaGrad on CIFAR-100 & 0.1 \\
\hline Server learning rate with SGDM on CIFAR-100 & 3.3 \\
\hline Server learning rate with Adam on CIFAR-100 & 0.3 \\
\hline Server learning rate with AdaGrad on CIFAR-100 & 0.1 \\
\hline Local iteration number with SGDM on CIFAR-100 & 5 \\
\hline Local iteration number with Adam on CIFAR-100 & 5 \\
\hline Local iteration number with AdaGrad on CIFAR-100 & 5 \\
\hline Fuzz factor with Adam on CIFAR-100 & 0.1 \\
\hline Fuzz factor with AdaGrad on CIFAR-100 & 0.1 \\
\hline Client learning rate with SGDM on EMNIST & 0.1 \\
\hline Client learning rate with Adam on EMNIST & 0.1 \\
\hline Client learning rate with AdaGrad on EMNIST & 0.1 \\
\hline Server learning rate with SGDM on EMNIST & 1 \\
\hline Server learning rate with Adam on EMNIST & 0.1 \\
\hline Server learning rate with AdaGrad on EMNIST & 0.1 \\
\hline Local iteration number with SGDM on EMNIST & 10 \\
\hline Local iteration number with Adam on EMNIST & 10 \\
\hline Local iteration number with AdaGrad on EMNIST & 5 \\
\hline Fuzz factor with Adam on EMNIST & 0.1 \\
\hline Fuzz factor with AdaGrad on EMNIST & 0.1 \\
\hline Client learning rate with SGDM on Stack Overflow & 100 \\
\hline Client learning rate with Adam on Stack Overflow & 100 \\
\hline Client learning rate with AdaGrad on Stack Overflow & 100 \\
\hline Server learning rate with SGDM on Stack Overflow & 10 \\
\hline Server learning rate with Adam on Stack Overflow & 1 \\
\hline Server learning rate with AdaGrad on Stack Overflow & 10 \\
\hline Local iteration number with SGDM on Stack Overflow & 5 \\
\hline Local iteration number with Adam on Stack Overflow & 1 \\
\hline Local iteration number with AdaGrad on Stack Overflow & 5 \\
\hline Fuzz factor with Adam on Stack Overflow & 0.00001 \\
\hline Fuzz factor with AdaGrad on Stack Overflow & 0.00001 \\
\hline
\end{tabular}
\label{tbl.HyperparameterSettings}
\end{center}
\end{table}

\end{document}